\documentclass{article}

\usepackage{PRIMEarxiv}

\usepackage[utf8]{inputenc}
\usepackage[T1]{fontenc}
\usepackage{hyperref}
\usepackage{url}
\usepackage{booktabs}
\usepackage{longtable}
\usepackage{amsfonts}
\usepackage{amsmath}
\usepackage{nicefrac}
\usepackage{microtype}
\usepackage[numbers,sort&compress]{natbib}
\usepackage{fancyhdr}
\usepackage{graphicx}
\usepackage{siunitx}
\graphicspath{{media/}}

\title{Where LLM Graders Succeed and Break: \\ Evidence from Two Computer-Science Exams}

\author{%
  Ali Habibullah\thanks{Equal contribution.}\hspace{0.35em}\textsuperscript{,}\thanks{Corresponding author.} \ , Yazan Alshoibi\footnotemark[1] \ \& Mohammad Alshiekh\footnotemark[1] \\
  KAUST Academy \\
  Computer, Electrical \& Mathematical Sciences \& Engineering (CEMSE) \\
  King Abdullah University of Science and Technology (KAUST) \\
  Thuwal, Saudi Arabia \\
  \texttt{\{ali.habibullah, yazen.shaebi, mohammad.shiekh\}@kaust.edu.sa} \\
  \AND
  Salman Khan \\
  Visual Artificial Intelligence Laboratory \\
  Oxford Brookes University \\
  \texttt{salmankhan@brookes.ac.uk} \\
  \And
  Naeemullah Khan \\
  KAUST Academy \& CEMSE, KAUST \\
  Lady Margaret Hall, University of Oxford \\
  \texttt{naeemullah.khan@kaust.edu.sa}
}

\newcommand{\repourl}{\url{https://github.com/KAUST-Academy/where-llm-graders-succeed-and-break}}
\newcommand{\artifactsshipin}{are released at \repourl}
\newcommand{\artifactsin}{in the accompanying repository}
\newcommand{\artifactsnote}{\footnote{Code, data and results: \repourl.}}

\begin{document}
\maketitle
\begin{abstract}
One long-form exam in a large course costs hundreds of grader-hours, and qualified graders are scarce; LLM graders are a tempting alternative. To show its pitfalls we grade a practical Computer Vision exam ($570$ dual-graded students) under $171$ configurations spanning closed and open-weights models; the best reaches mean absolute error $1.64/35$, below the $2.61/35$ two human graders achieve against each other. The catch is the prompt: a short ``strict grader'' preamble drives $14$ of $17$ open-weights models out of the graded band ($\text{MAE} \ge 8$), three stopping grading altogether. The damage traces to the preamble's two credit-withholding sentences, not to tone or model scale; one of them, ``never give partial credit'', alone makes two of three probed models stop grading. The closed flagships of three vendors shift calibration under it but stay in the band. In $162$ further configurations on a second, independent Machine Learning exam from another course ($1{,}038$ dual-graded students), the preamble worsens ten models, moving three out of the band into collapse and one into refusal, yet improves seven whose neutral prompts over-mark: the vulnerability replicates, but its direction is exam-specific. Light LoRA fine-tuning repairs it: one adapter on the two exams' pooled $\sim 3{,}900$ graded examples brings five small open models to parity or better with a human grader in agreement with the grader pair, and sensitivity to the three harsh personas nearly vanishes ($\le 0.32$ MAE). We release the anonymised dataset, full ablation grid, and grading, fine-tuning and analysis pipelines\artifactsnote.

\end{abstract}


\section{Introduction}
\label{sec:intro}

Large language models (LLMs) are being used in computer-science education~\citep{prather2023robotsherenavigatinggenerative}, grading included~\citep{phung2023generativeaiprogrammingeducation}: hundreds of grader-hours per cycle and scarce qualified graders make the economic case. The methodological case is murkier: the usual metrics hide which configuration of model, prompt and persona grades usably, and how it survives an instructor's small prompt edits. On a practical Computer Vision (CV) exam, dual-graded by fixed grader pairs, we run $171$ configurations with bootstrap CIs: closed and open-weight models ($7$B--$480$B), prompt components, personas, few-shot demonstrations, temperatures. Six full-cohort configurations, four Gemini and two GPT-5.5, match or beat the exam's inter-grader floor of $2.61/35$.

The central finding is an asymmetry: for closed models  the logical prompt (reference solution, grading guidelines, rubric breakdown, neutral persona) is already at a joint optimum (Section~\ref{sec:closed}), while open models are one sentence away from failure. A ``strict grader'' preamble knocks $14$ of $17$ models out of the graded band in all six tested open-weight families, three stopping grading entirely, the damage is unordered by scale and attributed to its two policy sentences, not its tone; across three closed vendors only Gemini's cheapest tier leaves the band, and only on the second exam (Sections~\ref{sec:brittle} and~\ref{sec:closed:vendors}). Failure takes three forms MAE alone cannot separate (zeroed submissions, blanket mark-downs, one structured field dismantled while the rest grades on); a conflicting-instruction account fits only the last, and only as a hypothesis (Section~\ref{sec:brittle:mech}).

The collapse is not an artifact  of one exam: under the same preamble, a $162$-configuration replication on another Machine Learning (ML) exam from a different course worsens $10$ of $17$ models, three leaving the graded band and one refusing outright, while the best closed and open configurations again land below its floor. The direction does not carry over: seven models \emph{improve}, the ML exam's over-marking cancelling against the persona, calibration masquerading as robustness (Section~\ref{sec:brittle:replication}).

The collapse trains away. One LoRA adapter on the two exams' pooled $\sim 3{,}900$ examples, supervised by per-question grader marks, takes five open models ($4$B--$30$B; Qwen, Llama, Gemma) from significantly worse than one human grader to parity or better under a paired third-grader test on \emph{both} exams; an adapter trained on one exam already improves the unseen other. The persona sweep costing base models up to $20$ MAE points moves the pooled adapters by at most $0.32$ under the three harsh personas, and by at most $0.39$ under \emph{lenient} bar Qwen3-Coder-$30$B-A$3$B (up to $1.81$; Section~\ref{sec:finetune}).

\section{Related Work}
\label{sec:related}

\paragraph{LLMs as judges and graders.}
LLM judging is standard since \citet{zheng2023judgingllmasajudgemtbenchchatbot} showed GPT-4 matches expert labellers on MT-Bench and \citet{liu2023gevalnlgevaluationusing} formalised prompt-plus-chain-of-thought scoring. Programming auto-grading mostly unit-tests functional correctness~\citep{messer2024automatedgradingfeedbacktools}, without rubric-aligned partial credit, or correlates LLM with human raters on $\sim 100$-submission datasets; the nearest benchmark, \citet{phung2023generativeaiprogrammingeducation}, compares ChatGPT and GPT-4 with human tutors. We apply the same scaffolding to \emph{student code grading} on two real, released exams ($n = 570$ and $1{,}038$) whose ground truth is two independent human graders, not a curated gold reference. Our few-shot arm supplies two worked examples per question labelled with D$01$'s scores and rationales, not the graders' marks --- in-context distillation of the best closed grader in the format of \citet{brown2020languagemodelsfewshotlearners} (Appendix~\ref{app:open}).

\paragraph{Prompt brittleness and personas.}
\citet{sclar2024quantifyinglanguagemodelssensitivity} show that semantically meaningless formatting changes shift benchmark performance by tens of percentage points and, with \citet{mizrahi2024stateartmultipromptllm}, urge reporting distributions over prompt variants; \citet{deshpande2023toxicitychatgptanalyzingpersonaassigned} report persona-dependent toxicity rises up to $6\times$ in ChatGPT. Our perturbation instead carries meaning, and we replay it on a second exam, where the vulnerability replicates but its direction and failure mode do not (Section~\ref{sec:brittle:replication}).

\paragraph{Instruction hierarchies and conflicts.}
Section~\ref{sec:brittle:mech}'s failure mode --- abandoning one of two incompatible instructions rather than negotiating them --- connects to instruction arbitration. Models treat instructions as equally privileged unless trained on an explicit hierarchy~\citep{wallace2024instructionhierarchy}, fail a non-trivial fraction of simple verifiable instructions~\citep{zhou2023instructionfollowingevaluationlarge}, break stated rules even on straightforward tests~\citep{mu2023llmsfollowsimplerules}, degrade sharply when instructions conflict~\citep{zhang2025ihevalevaluatinglanguagemodels}, and rarely flag the conflict~\citep{he2025coninstructevaluatinglargelanguage}. Here the conflict arises from an ordinary instructor edit, not an attack, with a clean behavioural signature: the structured score channel surrenders while the prose channel keeps following the rubric.

\section{Experimental Setup}
\label{sec:setup}

\paragraph{Exams, graders and grid.}
$570$ students took the CV exam's four code questions (\textbf{Q1} transfer learning, \textbf{Q2} CNN-from-scratch classification, \textbf{Q3} semantic segmentation, \textbf{Q4} bonus colorization); \textbf{Q1}--\textbf{Q3} form the $35$-point base scale (weights $12/11/12$) used throughout; bonuses graded but excluded. Two of $20$ graders in $10$ fixed pairs ($\approx 57$ students each) grade every submission independently; the grader-average total is the ground truth (Section~\ref{sec:floor}). The ML exam --- another course, $1{,}038$ students, three questions, $\approx 65$-point scale, dual-graded by $49$ graders in non-fixed pairs --- carries a $162$-configuration replication (Section~\ref{sec:brittle:replication}, Appendix~\ref{app:replication}). We evaluate $4$ Gemini models~\citep{geminiteam2025geminifamilyhighlycapable,comanici2025gemini25pushingfrontier} via the Vertex API, GPT-5.5, GPT-5.4 and Claude Opus 5 via their batch APIs (Section~\ref{sec:closed:vendors}), and $17$ open-weights variants from six vendors, from $7$B to $480$B, dense and mixture-of-experts, served locally with vLLM~\citep{kwon2023efficientmemorymanagementlarge} on A100 nodes. The $171$ CV configurations ($33$ closed, $138$ open-weights) span per-model baselines, prompt-component removals, thinking, five-persona sweeps, few-shot prompting, temperature and mechanism probes (Section~\ref{sec:brittle:attrib}); Appendix~\ref{app:details} documents roster, serving stack, reproducibility, thinking configuration and exclusions; Appendices~\ref{app:runs} and~\ref{app:iaruns} tabulate every run.

\paragraph{The prompt.}
The default prompt supplies the question, its rubric, the reference solution, the student notebook, grading guidelines and a breakdown of each scorable item; the model returns a score, any bonus and a rationale. Each question is graded in its own call (three or four per student) with only its own notebook, serialised cell by cell; prompts run to $\approx 22$k characters on the CV exam, $\approx 16$k on the ML exam, half to two thirds shared reference material. Personas are short preambles; \emph{strict} reads ``\emph{You are a HARSH teaching assistant. Award the MINIMUM defensible score for any task that is incomplete, buggy, or deviates from the rubric. Never give partial credit if the task does not run correctly.}'', the \emph{neutral} default ``\emph{You are a strict but fair teaching assistant.}'', with \emph{lenient}, \emph{rigorous} and \emph{exacting} analogous.

\paragraph{Metrics.}
The headline metric is MAE of the AI total against the grader-average total, with $95\%$ percentile bootstrap Confidence Intervals (CIs) (\num{2000} resamples)~\citep{10.1214/aos/1176344552,efron1994introductionbootstrap}, alongside the \emph{mean signed error} (bias); the human floor is computed identically, $\text{MAE} = \mathbb{E}[\,|\text{G}_1 - \text{G}_2|\,]$ over the grader pair, and both are recomputed unchanged on the ML exam's own scale. Because a total can hide per-question errors of opposite sign, Appendix~\ref{app:perq} decomposes the bias per question for the headline runs and Table~\ref{tab:itemcorr} (Appendix~\ref{app:details}) gives the item-level Spearman $\rho$ per question alongside the total. Students whose grading call permanently failed are excluded from that run's metrics --- none in most runs, at most $3$ elsewhere, exceptions in Appendix~\ref{app:details}.

\section{The Human-Grader Floor}
\label{sec:floor}

Across all $570$ dual-graded submissions of the CV exam the two graders' totals agree to $\text{MAE} = 2.61 / 35$ ($95\%$ bootstrap CI $[2.37, 2.85]$; Pearson $r = 0.868$). We treat $2.61$ as the human-grader floor: an AI configuration whose MAE sits at or below it makes, in aggregate, no more error against the grader average than the two graders make against each other. The floor is comparative only, and leans in the AI's favour because averaging two graders cancels part of their noise. Beating it yields a parity claim of the kind Section~\ref{sec:finetune:parity} retests with a paired third-grader test. The floor pools across pairs whose internal disagreement varies by $4.2\times$: per-pair MAE runs $0.85$ to $3.55$ over the $10$ fixed pairs, so which pair a student draws is a source of variability (Appendix~\ref{sec:groundtruth}).

\section{The Strict-Persona Collapse}
\label{sec:brittle}

\subsection{The collapse, quantified}
\label{sec:brittle:numbers}

Table~\ref{tab:brittle} (Appendix~\ref{app:brittle}) pairs each model's \emph{strict} and \emph{neutral} runs, identical bar the persona sentence; Figure~\ref{fig:brittle} renders all $51$ strict-flavoured runs. MAE alone cannot separate the failures (zeroing every student scores $\text{MAE} \approx 26$, the distance from the grader mean), so we add a \emph{behaviour} class: a \textbf{refusal} zeroes at least $90\%$ of students with awarded-total standard deviation below $0.5$, a \emph{near-refusal} keeps some variation at that zero-rate, a \textbf{collapse} reaches $3.07\times$ its own exam's inter-grader floor while still discriminating between students, and anything else is \emph{graded}. The CV exam's $\text{MAE} \ge 8$ fixes that multiple ($3.07 \times 2.61$) and gives $\text{MAE} \ge 15.7$ on the ML exam ($3.07 \times 5.13$), the same severity on both scales.

\paragraph{Every open-weight family breaks.}
Prepending a short ``strict grader'' preamble to the otherwise-best prompt takes $14$ of the $17$ config-matched pairs out of the graded band, all six families represented, MAE rising $\times 1.69$ to $\times 7.11$. Three models stop grading outright: Llama-3.1-8B and Mistral-Small-24B award $0.00/35$ on average, GLM-4-9B $0.56$; Qwen2.5-Coder-$32$B still marks students ($\text{MAE} = 20.29$, bias $-20.28$, mean awarded $5.75/35$). No Gemini configuration collapses under any wording here (Appendix~\ref{app:brittle}): \texttt{gemini-3.1-pro-preview} moves $1.86 \to 2.75$ under \emph{strict} (F$03$, still at the floor), Flash-Lite's strict-flavoured runs sit at $3.73$--$5.75$ --- an imperfect calibration knob, not a failure mode --- and GPT-5.5, GPT-5.4 and Claude Opus 5 stay in band (Section~\ref{sec:closed:vendors}).

\paragraph{Scale does not order the damage.}
Parameter count and the strict/neutral MAE ratio correlate at $\rho = -0.21$ ($p = 0.42$) across the $17$ sized models. The two largest ratios belong to a $24$B model that stops grading and a $14$B that zeroes $88\%$ of submissions, while $106$B GLM-4.5-Air collapses to $21.02$; GLM \emph{improves} from $9$B to $32$B before worsening at $106$B while Gemma \emph{worsens} from $12$B to $27$B --- opposite directions over matched rungs. Robustness does appear at the top, Qwen3-$235$B-A$22$B and Qwen3-Coder-$480$B staying in band, but both are Qwen, the only family with rungs above $106$B, confounding family with scale exactly where it matters; the third in-band model, Qwen2.5-Coder-$7$B, clears the threshold by $0.08$ from an already-weak baseline.

\paragraph{The damage is specific to one preset.}
Two of $13$ models leave the band under \emph{rigorous} and two under \emph{exacting}, against $14$ of $17$ under \emph{strict} (Table~\ref{tab:wording}, Appendix~\ref{app:brittle}); only \emph{strict} carries the two policy sentences, so the gap is about policy content, not vocabulary.

\subsection{Attribution: the policy sentences, not the adjective}
\label{sec:brittle:attrib}

The \emph{strict} preset is three sentences: a frame (``You are a HARSH teaching assistant.'') and two policy sentences --- \textbf{S1}, ``Award the MINIMUM defensible score for any task that is incomplete, buggy, or deviates from the rubric.'', and \textbf{S2}, ``Never give partial credit if the task does not run correctly.'' Varying the adjective alone --- STRICT, RIGOROUS, FAIR --- over verbatim S1 and S2 moves MAE by $0.94$ to $2.15$ across five models, in no consistent direction (Table~\ref{tab:attrib}, Appendix~\ref{app:brittle}); \textbf{FAIR} with the harsh policy leaves Mistral-Small-24B at $25.05$, still not grading.

A $2 \times 2$ crosses S1 with S2 on three models, frame fixed: the frame alone moves each less than two points off neutral, both sentences take them to $20.26$--$26.04$, S2 alone to outright refusal on Llama-3.1-8B and Mistral-Small-24B. S1 is not benign --- alone it takes Llama-3.1-8B to $23.45$ and outdoes S2 on Qwen2.5-Coder-32B ($15.55$ against $12.92$) --- so dominance varies by family: S1 for the $32$B, S2 for the other two, on both exams, where S2 again alone stops graders grading (Appendix~\ref{app:replication}). S2 contradicts the partial-credit scale the rubric scaffold mandates --- hence the safer wordings, and advice about one instruction, not a tone.

\subsection{Mechanism: three failure modes, one MAE range}
\label{sec:brittle:mech}

Three explanations fit the headline numbers: refusal, calibration drift, conflicting instructions (``be strict'' vs.\ ``award partial credit per rubric''). All three occur, on different models, indistinguishably by MAE: Gemma-3-27B reaches $\text{MAE} = 11.55$ having zeroed Q$1$ on \emph{zero} of $570$ students, GLM-4-32B a lower $9.66$ by zeroing $273$. One question separates them --- of students scored $0$ on Q$1$, what fraction received a non-zero Q$2$? --- and it splits the $19$ matched strict runs into \textbf{blanket zeroing} (five), \textbf{selective field collapse} (five) and \textbf{uniform severity} (nine), both thresholds in empty bands (Table~\ref{tab:mechanism}, Appendix~\ref{app:brittle}).

In the selective group the prompt's two output channels disagree: on the two hand-annotated runs, $20\%$ and $43\%$ of zeroed submissions carry a rationale itemising partial credit while the score field reads $0$ (Appendix~\ref{app:brittle}) --- one channel obeys S2, the other the rubric breakdown. But this covers neither other group, and its clearest prediction fails: withdrawing the breakdown \emph{worsens} the collapse in five of five families against a clean neutral control (Appendix~\ref{app:brittle}). An alternative fits that failure --- the breakdown may anchor the score field to non-zero sub-totals, so withdrawing it lets the credit-withholding sentence run unopposed --- and our grid cannot separate the two; we treat the mechanism as classified rather than explained (Section~\ref{sec:limitations}).

\subsection{The collapse replicates on the ML exam; its direction does not}
\label{sec:brittle:replication}

We replicate it on the ML exam (Section~\ref{sec:setup}; floor $5.13$ computed identically) over $162$ configurations: neutral and \emph{strict} runs for all $17$ open-weights models, five-persona sweeps, attribution probes, a closed-model arm (Appendix~\ref{app:replication}). Under \emph{strict} four models cross from the graded band to above that exam's ($\text{MAE} \ge 15.7$); Llama-3.1-$8$B again refuses. A fixed $\text{MAE} \ge 8$, only $1.56\times$ this floor, would instead put ten strict runs out of band and eleven \emph{neutral} baselines with them, against one here (DeepSeek-Coder-V2-Lite, $16.41$). The collapse is thus no artifact of one rubric, cohort or scale --- nor of open weights alone: Flash-Lite worsens under \emph{strict} ($7.53 \to 9.02$) and leaves the band under \emph{lenient} ($17.86$), while \texttt{gemini-3.1-pro-preview} stays below the floor ($3.40 \to 3.67$, IG$23$).

\begin{figure}[t]
\centering
\includegraphics[width=\linewidth]{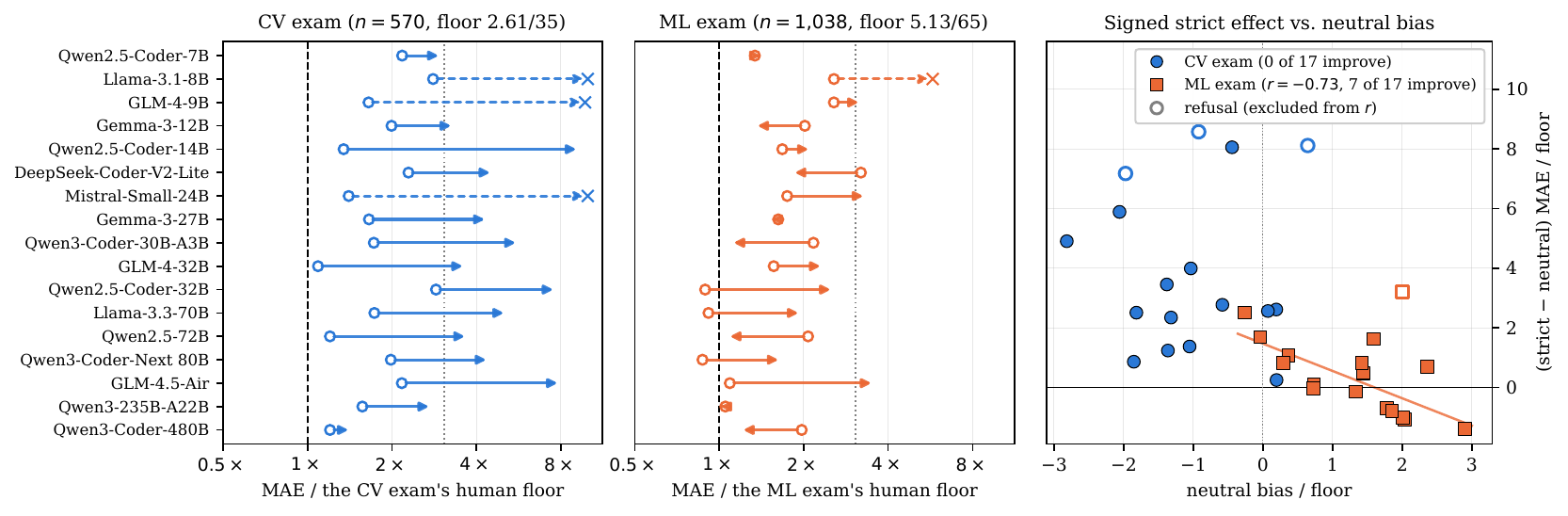}
\caption{The strict persona on the $17$ models run on both exams, floor-relative. \emph{Left, centre:} neutral (circle) $\to$ strict MAE; dashed = floor, dotted = each exam's band ($3.07\times$ floor: $\text{MAE} \ge 8$ on the CV exam, $\ge 15.7$ on the ML exam), dashed arrows = refusals (ceiling MAE). \emph{Right:} the signed strict effect $(\text{MAE}_{\text{strict}} - \text{MAE}_{\text{neutral}}) / \text{floor}$ against neutral bias, negative exactly where the persona cancels over-marking.}
\label{fig:crossexam}
\end{figure}

The \emph{direction} does not transfer (Figure~\ref{fig:crossexam}): seven of $17$ improve under the same sentence --- calibration, not robustness. The ML exam's neutral prompt over-marks for $15$ models, and across the $16$ non-refusal pairs neutral bias predicts the strict effect at $r = -0.73$ (permutation $p = 0.002$); on the CV exam, whose neutral biases straddle zero (none above $+1.68$), zero of $17$ improve. The decomposition also splits the two big Qwens' apparent robustness: the $480$B over-marks by $9.52$ points at neutral ($\text{MAE} = 10.10$, twice its floor) and is pulled back by the sentence; the $235$B is the one model above $106$B in band under both personas on both exams. Parity replicates on both sides (\texttt{gemini-3.1-pro-preview} at $0.66\times$ the ML exam's floor, Qwen3-Coder-Next at $4.47$), and the failure mode wanders: Qwen2.5-Coder-$32$B, selective here, zeroes whole submissions there.

\paragraph{One preamble, two remedies.}
One preamble separates competitive open-weights graders (Section~\ref{sec:open}) from $14$ out of band, a gap single-prompt evaluation cannot see~\citep{sclar2024quantifyinglanguagemodelssensitivity,mizrahi2024stateartmultipromptllm}, with two remedies: never forbid partial credit, or fine-tune lightly on in-house labels (Section~\ref{sec:finetune}).

\section{Closed-Model Configurations}
\label{sec:closed}

On the closed side the baseline recipe sits at the joint optimum: no prompt-component removal reliably improves on it.

\paragraph{Six full-cohort configurations at or below the floor.}
Six full-cohort configurations meet or undercut the human floor of $2.61$ with CI upper bounds below it (Table~\ref{tab:headline}, Appendix~\ref{app:closed}): D01 (\texttt{gemini-3-flash-preview}, neutral, $t = 0$) at $\text{MAE} = 1.64$ (bound $1.79$); F02 and D02 (the $3.1$-pro under \emph{lenient} and \emph{neutral}); B03 (Flash-Lite with thinking), an order of magnitude cheaper; and \texttt{gpt-5.5} under \emph{lenient} and \emph{neutral} (O03 $2.25$, O01 $2.43$; Section~\ref{sec:closed:vendors}). The floor comparison is not like-for-like (Section~\ref{sec:floor}), so we retest all six with the paired third-grader statistic of Eq.~\eqref{eq:paired} (Table~\ref{tab:zeroshotpaired}): three are better than a human grader --- D01 $\overline{d} = -0.54$ $[-0.71, -0.36]$, F02 $-0.37$ $[-0.56, -0.18]$, D02 $-0.37$ $[-0.55, -0.18]$ --- while B03 ($-0.19$ $[-0.39, +0.01]$) and both \texttt{gpt-5.5} runs ($-0.01$ $[-0.22, +0.21]$ lenient, $+0.19$ $[-0.02, +0.41]$ neutral) are at parity. The Flash-Lite baseline A01 is significantly worse ($+1.03$), as are \texttt{claude-opus-5} ($+1.13$) and all $17$ open-weights baselines' closest analogues (Section~\ref{sec:open}). Within Gemini the newer Flash beats the older Pro (D03, $4.10$); D01's scatter (Figure~\ref{fig:d01scatter}) is essentially unbiased, its residual error concentrated in the low-score band where the graders also disagree more.

\paragraph{On Flash-Lite, thinking helps and dropping the solution hurts.}
The B-series varies one prompt component at a time on Flash-Lite (Table~\ref{tab:bseries}). Dropping the reference solution flips a small under-grading bias ($-1.78$) into over-grading ($+3.30$) costing $\approx 0.8$ MAE; dropping the guidelines or breakdown moves MAE by at most $0.12$; thinking (B03) is the one large move, closing about three-quarters of the gap to D01 on a far cheaper model. Two combination runs cross a persona with another component: F01 (Flash-Lite, \emph{strict} $+$ thinking) at $3.73$ recovers most of the \emph{strict} penalty ($5.75$); F02 ($3.1$ Pro $+$ \emph{lenient}) reaches $1.79$ against $1.86$ neutral (Table~\ref{tab:allruns}). The G-series ablations (\texttt{gemini-3-flash-preview}, first $100$ students; Appendix~\ref{sec:closed:gseries}) beat D01 only as a sample-size artifact; the $t = 0$ variance probes are in Appendix~\ref{app:closed}.

\subsection{The asymmetry holds across vendors}
\label{sec:closed:vendors}

Gemini is one vendor, so we replayed the persona sweep on two more via batch APIs --- same prompt bytes, thinking off, temperature $0$ where exposed (Appendix~\ref{app:closed}, Table~\ref{tab:closedvendors}): OpenAI's \texttt{gpt-5.5} and \texttt{gpt-5.4} under \emph{neutral}, \emph{strict} and \emph{lenient} on both exams, Anthropic's \texttt{claude-opus-5} under \emph{neutral} on both and \emph{strict} on the CV exam. None leaves the graded band under \emph{strict} --- \texttt{gpt-5.5} $2.43 \to 4.43$ (CV exam) and $3.54 \to 3.70$ (ML exam), \texttt{claude-opus-5} $3.54 \to 5.17$, \texttt{gpt-5.4} $4.69 \to 6.90$ and $4.30 \to 5.11$, zero-total rates at most $0.7\%$, no refusal --- the calibration shift the Gemini models show ($1.86 \to 2.75$, $3.34 \to 5.75$), against $14$ of $17$ open-weights models leaving it under the identical sentence. The cheaper tiers are the fragile ones: \texttt{gpt-5.4} is worst under \emph{lenient} on the ML exam ($4.30 \to 8.30$, bias $+7.94$) but stays in band, where Flash-Lite reaches $17.86$ and leaves it; on the CV exam \emph{lenient} nearly halves its error ($4.69 \to 2.69$) by cancelling a $-4.45$ under-marking --- calibration masquerading as robustness (Section~\ref{sec:brittle:replication}), now on a closed model. At neutral \texttt{gpt-5.5} is below the floor on both exams, \texttt{claude-opus-5} on the ML exam only ($4.37$).

\section{Open-Weights Configurations Under Neutral Prompts}
\label{sec:open}

Under neutral prompts nothing stands out: the $17$ baselines span $2.85$ to $7.49$ MAE, no family dominating or collapsing (Table~\ref{tab:openbaselines}, Appendix~\ref{app:open}); none of it predicts which stop grading under a persona sentence. Scaling is not monotone: Qwen2.5-Coder improves from $7$B to $14$B ($5.68\to3.51$) then regresses at $32$B ($7.49$, bias $-7.35$); dropping the reference solution recovers $3.12$, consistent with over-anchoring on it, though exam-locally: the same removal moves it $+0.16$ on the ML exam. Within Qwen3-Coder the $30$B-A$3$B MoE beats the $80$B Coder-Next ($4.50$ vs.\ $5.17$), reversing at $480$B ($3.14$).

\paragraph{GLM-4-32B takes four of the five best open-weights slots.}
Its best (rubric breakdown removed, neutral; L-BD$07$), $2.68$, is the grid's strongest open-weights grader (Table~\ref{tab:allruns}); L-X$04$ (\emph{rigorous}) reaches $2.74$, essentially unbiased; the fifth slot, Qwen3-Coder-Next $+$ \emph{lenient} ($2.81$) is a bias-correction artifact, not capability. Two gaps to D01 ($1.64$): just over one point on best achievable, $1.42$ on best surviving a persona sentence (Qwen3-Coder-$480$B under \emph{rigorous}, $3.06$; L-R$04$), since GLM-4-32B collapses to $9.66$ under \emph{strict}.

\paragraph{Model choice beats size.}
The $30$B-A$3$B beats the $80$B, the $14$B the $32$B, the $72$B dense the $235$B MoE; the $480$B's edge is persona robustness: at neutral it only ties the $72$B. \emph{Which} model is a property of the exam, not the model: across the $17$ run on both, neutral MAE is uncorrelated ($\rho=-0.07$) and Qwen2.5-Coder-$32$B, weakest here, ties statistically for best there. Open-weights graders are practically viable, but only zero-shot on a prompt guaranteed never to carry a strict-flavoured persona (Section~\ref{sec:brittle}), which Section~\ref{sec:finetune} removes with in-house labels.

\section{Fine-Tuning Repairs the Strict-Persona Collapse}
\label{sec:finetune}

  So far we have changed only the prompt. We have better material than prompts: two
  humans graded every submission in both exams. In this section we use their
  grades to fine-tune the models with LoRA~\citep{hu2021loralowrankadaptationlarge}.
  We fine-tune five small open models ($4$B to $30$B). Each model gets three
  adapters: one per exam, and one trained on both exams pooled (about
  $3{,}900$ examples). Three things happen. With the pooled
  adapter, every model grades as well as a human or better, on both exams.
  The skill transfers: an adapter trained on one exam also gets better at the
  other, without ever seeing it. And the persona problem is gone. The prompt
  that adds as much as $20$ MAE points to a base model barely moves the fine-tuned
  one.

\subsection{Setup}
\label{sec:finetune:setup}

  We fine-tune five open models across three families and sizes from $4$B to
  $30$B parameters with LoRA: Qwen2.5-Coder-$7$B and
  $14$B~\citep{hui2024qwen25codertechnicalreport},
  Qwen3-Coder-$30$B-A$3$B~\citep{yang2025qwen3technicalreport},
  Llama-3.1-$8$B~\citep{grattafiori2024llama3herdmodels}, and Gemma-4-E$4$B~\citep{gemmateam2026gemma4technicalreport}.
  Training hyperparameters are in Appendix~\ref{app:ft}. Each exam uses a seed-fixed $80/20$ student split: $456/114$ on the
  CV exam, giving $1{,}494$ training
  examples, and $830/208$ on the ML exam, giving $2{,}413$.
  The ML exam records one grade per question with the bonus
  folded in, so we split each grade into a score up to the base cap and the
  remainder as bonus. We train three adapters per model: one per exam and
  one pooled. The pooled training set is the union of the two train splits,
  so no adapter ever sees a held-out student. We evaluate all three, plus the
  untuned base, on both held-out sets. On the held-out students, the two human graders disagree with each other by
  $2.83$ marks on average on the CV exam and $5.19$ on the
  ML exam. These are the floors: a grader that disagrees with
  the humans by less than this agrees with them better than they agree with
  each other. We use two training recipes with identical prompts. The \emph{marks}
  recipe averages the two graders' scores into a single target, one example
  per student and question, with a \{score, bonus\} output format. The
  \emph{bd} (breakdown) recipe additionally distils a per-task breakdown from
  Gemini~3 Flash Preview (run D$01$), without its reasoning text.
\subsection{One adapter reaches human parity or better on both exams}
\label{sec:finetune:parity}

  Every untuned base grades worse than a human on both exams and
  mis-calibrates in opposite directions: Llama-3.1-$8$B reads $19.8$ on the
  CV exam where the grader average is $25.3$, yet $40.5$ on the
  ML exam where it is $30.9$. With a score standard deviation
  of $2.1$, it barely distinguishes the CV exam's students at
  all. The pooled fine-tune produces better results on both exams: it lands
  at $1.75$--$2.01$ MAE on the CV exam and $3.29$--$3.53$ on the
  ML exam (marks recipe; Table~\ref{tab:ftmain}), revives the collapsed
  score spread on the CV exam ($\sigma$ from $2.1 \to 7.3$--$7.5$,
  against $8.0$ for the grader average), and replaces the bases' bias of
  $-5.9$ to $+9.5$ marks with under one mark of generosity ($+0.2$ to $+0.9$)
  on both exams.

Because MAE against a two-grader \emph{average} is easier than any single grader faces (Section~\ref{sec:floor}), the verdicts come from a paired third-grader test:
\begin{equation}
d_i \;=\; \tfrac{1}{2}\big(|\mathrm{AI}_i-\mathrm{G}_{1,i}| + |\mathrm{AI}_i-\mathrm{G}_{2,i}|\big) \;-\; |\mathrm{G}_{1,i}-\mathrm{G}_{2,i}|.
\label{eq:paired}
\end{equation}
Pairing removes per-student exam difficulty; averaging $d$ over the held-out students with a $95\%$ CI, an interval containing $0$ means the AI is statistically indistinguishable from a human grader, entirely below $0$ that it disagrees with the humans \emph{less than they disagree with each other}, entirely above, worse. Below zero is possible because $|\mathrm{G}_1-\mathrm{G}_2|$ carries \emph{two} graders' idiosyncratic noise while $|\mathrm{AI}-\mathrm{G}_i|$ carries one human's plus the model's; the AI beats the floor exactly when its own noise is smaller than a single human's. With no ground truth beyond the graders, \emph{better}
  here means more reliable, not more correct: added to the grading pool, the
  model would introduce less disagreement than another human.

All base configurations are significantly worse than a human grader on both exams. The pooled marks adapter is significantly \emph{better} in $9$ of $10$ (model $\times$ exam) cells: all five models on the ML exam ($\overline{d}$
  from $-0.94$ to $-1.13$, every CI below zero) and four of five on the CV. The pooled bd adapter reaches at least parity in all $10$ cells and is better in two. No cell on either exam reads worse (Table~\ref{tab:ftpaired},
  Appendix~\ref{app:ft}). The ML exam sets a lower bar because its two graders disagree more (floor $5.19$ vs.\ $2.83$), but every point estimate there is negative regardless.

\subsection{Grading skill transfers across exams}
\label{sec:finetune:transfer}

An adapter fine-tuned on one exam improves grading on the other, which it never saw: mean MAE on the ML exam falls from $7.95$ (base) to $5.36$ under the CV-exam adapter, and on the CV exam from $5.54$ to $3.57$ under the ML-exam adapter. Every model improves in both directions, with one exception discussed in Section~\ref{sec:finetune:choices}. Transfer stops short of in-domain performance ($3.54$ and $1.90$), but pooling closes the gap for free: the pooled adapter matches or beats each specialist \emph{on the specialist's own exam} in all eight (exam $\times$ recipe) columns (Table~\ref{tab:ftmain}).

\begin{table}[t]
\centering
\caption{Held-out MAE vs.\ the grader average by training set (marks recipe; bd in Table~\ref{tab:ftmeans}, Appendix~\ref{app:ft}). The cross columns are transfer; pooled matches or beats each specialist on its own exam. $\ddagger$: the one negative transfer cell (Section~\ref{sec:finetune:choices}).}
\label{tab:ftmain}
\small
\setlength{\tabcolsep}{4.5pt}
\begin{tabular}{lrrrrrrrr}
\toprule
& \multicolumn{4}{c}{CV exam ($n{=}114$)} & \multicolumn{4}{c}{ML exam ($n{=}208$)} \\
\cmidrule(lr){2-5}\cmidrule(lr){6-9}
Model & base & CV & ML & pooled & base & CV & ML & pooled \\
\midrule
Qwen2.5-Coder-7B    & 6.60 & 1.93 & 3.22 & \textbf{1.84} & 6.48 & 6.37 & 3.67 & \textbf{3.40} \\
Qwen2.5-Coder-14B   & 4.06 & 1.89 & 3.29 & \textbf{1.89} & 7.70 & 5.41 & 3.35 & \textbf{3.29} \\
Qwen3-Coder-30B-A3B & 4.76 & 1.80 & 2.59 & \textbf{1.75} & 8.60 & 5.22 & 3.67 & \textbf{3.45} \\
Gemma-4-E4B         & 4.29 & 1.88 & 3.79 & \textbf{1.88} & 4.66 & 4.79$^\ddagger$ & 3.51 & 3.53 \\
Llama-3.1-8B        & 7.99 & 1.99 & 4.95 & \textbf{2.01} & 12.30 & 5.02 & 3.48 & \textbf{3.37} \\
\midrule
mean                & 5.54 & 1.90 & 3.57 & \textbf{1.87} & 7.95 & 5.36 & 3.54 & \textbf{3.41} \\
\bottomrule
\end{tabular}
\end{table}

\subsection{Fine-tuning immunises against the strict-persona collapse}
\label{sec:finetune:personas}

  We re-run the persona sweep for base and pooled models on both exams (Tables~\ref{tab:ftpersona} and~\ref{tab:ftpersonafull}, Appendix~\ref{app:ft}). The bases reproduce the collapse on both exams: under
  \emph{strict}, Qwen2.5-Coder-$14$B goes $4.06 \to 23.81$ on the CV exam, and Llama-3.1-$8$B reaches MAE $30.9$ on the ML with bias $-30.9$: it scores nearly every student zero. The pooled adapters are flat everywhere: at most
  $+0.32$ (marks) and $+0.62$ (bd) across three personas, two exams and five models, and the catastrophic tail the collapse creates is gone: under a neutral prompt the worst base mis-grades $43$ of $114$ CV-exam students by more than
  $10$ marks, while every pooled model is at $0$--$1$. On the CV exam the \emph{strict} preset is the most destructive on every base, as Section~\ref{sec:brittle:attrib} predicts: it alone carries the two policy sentences. The per-exam adapters had one immunity failure, and pooling fixes it: Llama-bd under \emph{strict} sat at $4.08$ when tuned on the CV exam alone and drops to $2.67$ with pooled training.

  On the ML exam \emph{strict} sometimes improves a base model (Qwen3-Coder-$30$B: $8.60 \to 6.34$). This is not the persona working. The bases over-grade the ML exam, and the induced harshness happens to cancel part of the bias, the
  accidental calibration of Section~\ref{sec:brittle:replication} again. So the same wording that destroys a model on one exam helps it on the other. An effect that flips sign between exams is not a calibration knob.

\subsection{What matters, what does not}
\label{sec:finetune:choices}

  Size and family remain nearly irrelevant. The five pooled marks fine-tunes converge to $1.75$--$2.01$ MAE on the CV exam and $3.29$--$3.53$ on the ML, and starting quality does not predict the tuned result: Llama-3.1-$8$B starts
  worst on both exams ($7.99$, $12.30$) and finishes with everyone else ($2.01$, $3.37$).   Human labels alone suffice: pooled marks matches or beats pooled bd on every (model, exam) cell, decisively on the ML exam ($3.41$ vs.\ $4.15$ mean). Nine of the eleven better verdicts in the paired test are marks verdicts, and
  marks needs no closed-model teacher. Which exam the labels come from matters less than having labels at all, with one exception. Gemma-4-E$4$B is the
  strongest base on the ML exam ($4.66$) and the only model that cross-exam transfer fails to help there ($4.79^\ddagger$), and in-domain labels move it only to $3.51$: fine-tuning buys the most where the base grades worst, and
  Gemma had the least room to improve. The residual errors concentrate on the students the two human graders also disagree on (correlation $+0.35$ to $+0.48$ for every pooled model, against no consistent relationship for the bases; Appendix~\ref{app:ft}): the tuned models are unsure exactly where the rubric is ambiguous.

\section{Discussion}
\label{sec:discussion}

\paragraph{What the collapse is, and is not.}
Of the three failure modes only selective field collapse looks like a failure to negotiate conflicting instructions, and only as a hypothesis (Section~\ref{sec:brittle:mech}); the others are calibration drift and refusal, which never appears on a Qwen model, so a single-family roster misses that mode. Spanning all six families unordered by scale, the failure belongs to open-weights instruction following: no closed model of the three vendors leaves the band on either exam, and robustness above $106$B is half illusory once the ML exam decomposes it. Fine-tuning shows the failure is zero-shot, not small-model. 

\paragraph{Why single-prompt evaluation misses it.}
One well-behaved prompt leaves the best open-weights configuration just over one MAE point behind Gemini's best ($2.68$ vs.\ $1.64$), one instructor-style persona preamble nearly fourteen behind ($15.58$): single-prompt evaluation would have told a sunnier story~\citep{sclar2024quantifyinglanguagemodelssensitivity,mizrahi2024stateartmultipromptllm}.

\paragraph{Implications for deployment.}

\textbf{(1)} \textbf{Never instruct a grader to withhold credit}: both policy sentences collapse graders on their own (Section~\ref{sec:brittle:attrib}), unordered by parameter count, and under-grading harms students who attempted the work in good faith. Verify the model against the prompt \emph{and the exam}: on the ML exam the same preamble helps seven of $17$ models purely by cancelling their over-marking, yet nearly doubles the error of the ML exam's best. \textbf{(2)} \textbf{With in-house dual-graded examples, fine-tune rather than prompt-engineer --- and pool the courses.} One adapter on the two exams' pooled $\sim 3{,}900$ examples brings every $4$B--$30$B model to parity or better with a human grader on both exams and removes the collapse on both (persona drift $\le 0.32$ MAE), with no closed-model teacher and no cost to either course. \textbf{(3)} \textbf{Evaluate several prompt variants}, including instructor-written ones. \textbf{(4)} \textbf{Report bias alongside MAE}: Flash-Lite's lenient and strict have comparable MAE with opposite-signed bias ($+4.01$ vs.\ $-5.44$): MAE alone does not show which way the harm runs.

\section{Limitations}
\label{sec:limitations}

\paragraph{Coverage.}
Both exams are from one university: replication spans courses, cohorts, rubrics and scales, not institutions. Fine-tuning's two transfer directions are not size-matched ($456$ vs.\ $830$ training students), confounding their comparison; Gemma-4's ML-exam baseline, unlike the other four, lacks independent corroboration. We varied three strict-flavoured wordings and two preset policy sentences, not every credit-withholding instruction. Both models above $106$B are Qwen, entangling scale with family where it matters (Section~\ref{sec:brittle:replication} separates their robustness, not that confound). The \texttt{gemini-3-flash-preview} ablations used the first $100$ students (matched-subset baselines, wider CIs). Closed coverage: four Gemini models; three personas on two OpenAI tiers and one Anthropic model (Section~\ref{sec:closed:vendors}; Anthropic \emph{strict} on the CV exam only), no \emph{rigorous} or \emph{exacting} outside Gemini. Grading is static: nothing is executed (Appendix~\ref{app:details}; no test suites, minutes-to-hours GPU cost, no stored outputs), so ``does not run'' verdicts are unaudited and their error rate on valid but unconventional code unmeasured. We know of no public corpus of long-form notebooks with dual human rubric grades for external benchmarking; automated-grading corpora are unit-test based~\citep{messer2024automatedgradingfeedbacktools} or score short per-task programs~\citep{phung2023generativeaiprogrammingeducation}.

\paragraph{ML-exam ground truth.}
Nine of the ML exam's $1{,}038$ rows record $0.0$ for one grader against real marks from the partner; they are retained in the headline floor but excluded from the maximum-disagreement figure; excluding them everywhere moves no behaviour class, no table ordering and no held-out verdict (Appendix~\ref{app:replication}). Its pairing is hybrid rather than fixed, so per-pair statistics are scoped to the $18$-pair backbone covering $79\%$ of the cohort.

\paragraph{Few-shot demonstrations.}
The K-series demonstrations carry D$01$'s labels, and their five students remain in the evaluated cohort ($5/570$), graded with their own worked answer in the prompt. The ML exam's demonstrations likewise carry IG$08$'s labels, with three students remaining; its Flash-Lite arm suffers biased dropout from prompt-length failures, so only its matched subsample is comparable (Appendix~\ref{app:replication}).

\paragraph{Claims.}
Section~\ref{sec:brittle:mech}'s three-way split classifies rather than explains: withdrawing the rubric breakdown contradicts the conflicting-instruction reading, leaving the causal story open. The few-shot uplift is capability, not held-out generalisation: the demonstrations carry D$01$'s own labels. One seed-fixed split and one run per configuration make the paired third-grader test load-bearing; its normal-approximation CIs leave three CV-exam \emph{better-than-grader} verdicts marginal: upper bounds $-0.044$ (14B marks), $-0.020$ (Gemma marks) and $-0.001$ (7B bd), all strictly below zero under a $2000$-resample percentile bootstrap, though the weakest would contain zero under a $t$-interval ($p = 0.052$ paired, $0.069$ Wilcoxon); the ML exam's are not (upper CIs $\le -0.30$). Appendix~\ref{app:limitations} details the remaining caveats.

\section{Conclusion}
\label{sec:Conclusion}
For a dual-graded exam the recipe is short. The stronger closed models (Gemini 3 Flash and 3.1 Pro, GPT-5.5) grade at or below the human floor untuned, from a neutral prompt; an open model gets there only with light fine-tuning on the exam's own grades, and only then survives the instructor edits that break it zero-shot. Between exams the vulnerability and its repair transfer; the model ranking, the persona's direction and $t = 0$ determinism do not. So verify a grader on the exam it will grade, under an instructor's own prompts.

\section*{Acknowledgments}
We thank Prof. Sultan Albarakati (KAUST) for his support, which made this paper possible. We thank all graders whose grading produced the human ground truth, and the course instructors for permission to use the anonymised exam data.  For computer time, this research used Ibex managed by the Supercomputing Core Laboratory at King Abdullah University of Science \& Technology (KAUST) in Thuwal, Saudi Arabia.

\section*{AI Use Statement}

We used generative AI tools (Opus 4.8, Opus 5.0 and Fable 5) as coding and writing assistants throughout this work. Specifically: for implementing and refactoring the data pipeline, the grading harness, the fine-tuning scripts, and the analysis code; for drafting and editing prose in this paper; and for exploratory literature search. We did not use generative AI to generate research ideas, to produce experimental results, or to write any of the analysis numbers reported here --- every number in this paper is computed by the released scripts from the released data, and the ledgers those scripts emit are the authority for the manuscript. All AI-assisted code was reviewed and executed by the authors, and the statistical results were independently re-derived before being reported. We take responsibility for the final content of this work, including all text, claims, and artifacts.

Note that generative models are also the \emph{object} of study here rather than a tool: the LLM graders evaluated in Sections~\ref{sec:closed}--\ref{sec:finetune} produced the scores we analyse, and those outputs are released in full.

\section*{Reproducibility Statement}

Every number in the paper is regenerated from released artifacts by released code. The two anonymised exam datasets, the rubrics and reference solutions, one result workbook per configuration, and the analysis, grading, and fine-tuning pipelines \artifactsshipin. Section~\ref{sec:setup} and Appendix~\ref{app:details} specify the prompt, the serving stack, the decoding parameters, and the exclusion rules; Appendix~\ref{app:runs} tabulates all $171$ CV-exam configurations with their bootstrap confidence intervals; Appendix~\ref{app:ft} gives the fine-tuning hyperparameters and the seed-fixed split. Training uses HuggingFace with PEFT and every evaluated number is served by vLLM, Gemma-4's adapter included. One caveat is stated where it arises: evaluation is not bitwise reproducible because vLLM batching reorders reductions ($\sim 0.01$ MAE); Gemma-4's vLLM serving path was cross-validated against HuggingFace generation to within $\sim 0.1$ MAE (Appendix~\ref{app:ft}).

\section*{Ethics Statement}

The data are student exam submissions and grader marks from two university courses, released with the course instructors' permission. Both datasets are anonymised before release: student identifiers, names, and email addresses are removed from notebook code, markdown, outputs, and file metadata, and graders appear only as opaque identifiers. No demographic attributes are collected or released, and no student is identifiable from the released artifacts.

The application this paper studies carries real risk to the people being graded, which is why we report the failure mode rather than only the headline accuracy. An under-grading collapse of the kind documented in Section~\ref{sec:brittle} harms students who attempted the work in good faith, and it is invisible to single-prompt evaluation. We therefore recommend against deploying an LLM grader on the basis of one prompt's measured accuracy, and we report bias alongside MAE throughout so that the direction of harm is visible. We regard human oversight of consequential grading decisions as a requirement, not an option, and none of the results here should be read as licensing unsupervised automated assessment.

\bibliographystyle{unsrtnat}
\bibliography{references}

\clearpage
\appendix

\section{Experimental Details}
\label{app:details}

\paragraph{The CV exam in full.}
The CV exam's $570$ submissions answer four code questions: \textbf{Q1} transfer learning with EfficientNetV2, \textbf{Q2} a CNN from scratch, \textbf{Q3} semantic segmentation with a pretrained encoder, \textbf{Q4} a colorization-dataset bonus. Beyond the $35$ base points (weights $12$ / $11$ / $12$) sit up to $13$ bonus points (Q1, Q2 and an all-or-nothing $5$-point Q4), graded but excluded from every reported metric.

\paragraph{Model roster and serving.}
On the closed-model side we use Google's \texttt{gemini-2.5-pro}~\citep{comanici2025gemini25pushingfrontier} and \texttt{gemini-flash-lite}, \texttt{gemini-3-flash-preview}, and \texttt{gemini-3.1-pro-preview}, for which no technical report was available at the time of writing, via the Vertex API~\citep{geminiteam2025geminifamilyhighlycapable}. Two further vendors enter only in the cross-vendor replication of Section~\ref{sec:closed:vendors}: OpenAI's \texttt{gpt-5.5} and \texttt{gpt-5.4} and Anthropic's \texttt{claude-opus-5} (Appendix~\ref{app:closed}, Table~\ref{tab:closedvendors}). On the open-weights side we use $17$ variants from six vendors. From Qwen: Qwen2.5-Coder at $7$B / $14$B / $32$B~\citep{hui2024qwen25codertechnicalreport} and the Qwen2.5 $72$B dense model~\citep{qwen2025qwen25technicalreport}, and from the Qwen3 family the Qwen3-Coder $30$B-A$3$B MoE, the $235$B-A$22$B MoE, and the $480$B Qwen3-Coder MoE in FP8~\citep{yang2025qwen3technicalreport}, and the $80$B Qwen3-Coder-Next MoE~\citep{qwen2026qwen3codernext}. From Meta: Llama-3.1-$8$B~\citep{grattafiori2024llama3herdmodels} and Llama-3.3-$70$B~\citep{meta2024llama33modelcard}. Z.ai: GLM-4-$9$B and GLM-4-$32$B, the $0414$ releases~\citep{glm2024chatglmfamilylargelanguage,zai2025glm49b0414,zai2025glm432b0414}, and the $106$B GLM-4.5-Air MoE~\citep{glm45team2025glm45agenticreasoningcoding}. Google: Gemma-3-$12$B and Gemma-3-$27$B~\citep{gemmateam2025gemma3technicalreport}. Mistral: Mistral-Small-$24$B~\citep{mistral2025small24binstruct2501}. DeepSeek: the $16$B DeepSeek-Coder-V2-Lite MoE ($2.4$B active)~\citep{deepseekai2024deepseekcoderv2breakingbarrierclosedsource}. Serving is vLLM~\citep{kwon2023efficientmemorymanagementlarge} on one $4 \times$~A100 node, or $8 \times$ for Qwen2.5-$72$B, $235$B-A$22$B and the $480$B FP8 model.

Open-weights generation is \emph{unconstrained}: the grader requests JSON in the prompt and parses the reply, with no grammar or schema enforcement at decode time (Vertex calls do pass a JSON response schema). A controlled A/B confirms that grammar-constrained decoding changes no score and costs $6.9\times$ in latency. Every open-weights grid run uses vLLM \texttt{0.19.1} or later, greedy decoding at temperature $0$ --- the $t = 0.5$ probes L-E$01$/L-E$02$ (Appendix~\ref{sec:closed:variance}) excepted --- and one request in flight. \paragraph{Notebook serialisation, parsing and retries.}
Every call receives one question's notebook as plain text --- a \texttt{--- Cell $k$ [code|markdown] ---} header then the cell's source --- with every output dropped: training logs, printed metrics and rendered plots or masks (base64 media) never reach the model. Nothing is truncated in the grid; the fine-tuning harness alone head-and-tail-truncates to $24{,}000$ characters. Replies are parsed as JSON; the open-weights path falls back from strict parsing to a lenient parse and then to \texttt{json\_repair}, the Vertex path parses strictly against its response schema. A Vertex call is retried up to three times with a $2s \times$ attempt back-off, except that a \texttt{MAX\_TOKENS} finish is not retried because it is deterministic at $t = 0$; the open-weights path does not retry a truncated or unparseable reply at $t = 0$ for the same reason. A question with no score after the run's retries and resumes is a permanent failure, and the student is excluded from that run's metrics as described under \emph{Exclusions}.

\paragraph{Ablation-grid legend.}
Appendix~\ref{app:runs} tabulates the grid's $33$ closed-model ($25$ Gemini, $6$ OpenAI, $2$ Anthropic) and $138$ open-weights runs; the machine-readable master comparison ships \artifactsin. Gemini series: \textbf{A} = baseline ($n{=}1$); \textbf{B} = prompt-component ablations on Flash-Lite ($n{=}4$: B$01$ drop solution, B$02$ drop guidelines, B$03$ thinking on, B$04$ drop breakdown); \textbf{C} = strictness persona on Flash-Lite ($n{=}2$); \textbf{D} = model swap ($n{=}3$); \textbf{E} = temperature variance ($n{=}3$); \textbf{F} = combos and the 3.1 Pro strict run ($n{=}3$); \textbf{G}, \textbf{M} = prompt ablations and persona on \texttt{gemini-3-flash-preview} ($n{=}4$, $n{=}2$); \textbf{K} = few-shot ($n{=}1$); \textbf{P} = alternate-wording personas on Flash-Lite ($n{=}2$). Cross-vendor closed series (batch APIs; Section~\ref{sec:closed:vendors}): \textbf{O} = OpenAI, \texttt{gpt-5.5} (O$01$--O$03$) and \texttt{gpt-5.4} (O$11$--O$13$) under \emph{neutral}, \emph{strict} and \emph{lenient} ($n{=}6$); \textbf{N} = Anthropic \texttt{claude-opus-5} under \emph{neutral} and \emph{strict} ($n{=}2$).

Open-weights series (prefixed \textbf{L-}) fall in three groups. \emph{Qwen ablations:} \textbf{L-A}/\textbf{L-AN} = per-model baselines ($n{=}5$); on Qwen2.5-Coder-32B, \textbf{L-B} = prompt-component ablations ($n{=}4$), \textbf{L-C} = persona ($n{=}2$) and \textbf{L-P} = alternate wordings ($n{=}2$); \textbf{L-D}/\textbf{L-DN} = $+$thinking on Qwen3 MoE ($n{=}2$); \textbf{L-E} = temperature variance ($n{=}2$); \textbf{L-G}, \textbf{L-H} = prompt ablations replayed on the $30$B and $7$B ($n{=}6$); \textbf{L-K} = few-shot ($n{=}1$); \textbf{L-M} = persona on additional Qwen models ($n{=}7$); \textbf{L-N}, \textbf{L-Q}, \textbf{L-R} = full five-persona sweeps on Qwen2.5-$72$B, Qwen3-$235$B-A$22$B and Qwen3-Coder-$480$B FP8 ($n{=}15$). \emph{Cross-family sweeps}, each five personas on all $570$ students: \textbf{L-S} = DeepSeek-Coder-V2-Lite; \textbf{L-U} = Llama-3.1-$8$B; \textbf{L-V} = Llama-3.3-$70$B; \textbf{L-W} = GLM-4-$9$B; \textbf{L-X} = GLM-4-$32$B; \textbf{L-J} = GLM-4.5-Air; \textbf{L-F} = Gemma-3-$12$B; \textbf{L-Y} = Gemma-3-$27$B; \textbf{L-Z} = Mistral-Small-$24$B ($n{=}45$). \emph{Mechanism probes:} \textbf{L-MP} = minimal pairs isolating the persona's headword from its policy sentences, on five models ($n{=}25$); \textbf{L-PS} = the $2\times2$ over the two policy sentences, on three models ($n{=}12$); \textbf{L-BD} = rubric-breakdown removal crossed with persona, on five models ($n{=}10$).

G- and M-series Gemini configurations run on the first $100$ students, \texttt{gemini-3-flash-preview} throughput being the binding constraint; comparisons against full-$n$ baselines restrict the baseline to the matching subset.

\paragraph{Gemini thinking configuration.}
The Vertex API exposes an internal reasoning budget, \texttt{thinking\_budget}. We run $21$ of the $25$ Gemini ablations with thinking \emph{disabled} (\texttt{thinking\_budget} $= 0$) and three (B$03$, F$01$, G$03$) \emph{enabled} at the dynamic budget ($-1$; the model picks the per-call budget). The exception is D$03$: \texttt{gemini-2.5-pro} rejects \texttt{thinking\_budget} $= 0$, so it ran under that model's default dynamic policy and reads as thinking-enabled (Section~\ref{sec:limitations}).

\paragraph{Exclusions.}
A student with any of Q$1$--Q$3$ ungraded after retries is excluded from that run's metrics rather than scored with a partial total: no students in most runs, at most $3$ elsewhere, the exceptions being P$02$ ($13$) and the temperature-variance E-series, where a student is kept only if all of Q$1$--Q$3$ were graded in at least one rerun ($21$ in E$01$, $7$ each in E$02$/E$03$); per-run $n$ accompanies the affected tables. P$02$'s $13$ are genuine permanent failures, one question each ($6$ on Q$1$, $7$ on Q$2$); E$01$'s $21$ are $14$ such failures plus the $7$ students who submitted no notebook for at least one of Q$1$--Q$3$, and E$02$/E$03$'s $7$ are those non-submitters alone --- neither run suffered a grading failure. Only the E-series rerun rule drops a non-submitter: every single-call run keeps them and scores the missing question $0$, and restoring them that way moves the E-series MAE by $-0.03$ ($3.38 \to 3.36$, $3.51 \to 3.48$, $3.60 \to 3.57$). No log survives for these four runs, so we cannot quote a \texttt{finish\_reason}; where logs do survive (the ML exam's Gemini grid and F$03$), $705$ of $709$ permanently failed cells are \texttt{MAX\_TOKENS} truncations and the other $4$ JSON parse errors, with $429$, $503$ and dropped connections retryable and no safety block at all. The artifacts agree: a failure burns $149$--$316$ s (P$02$, one call) or $2{,}095$--$2{,}529$ s (E$01$, five reruns) against a $5$--$6$ s normal call; E$01$'s $14$ cells fail $5/5$ at $t = 0$ yet grade $5/5$ at $t = 0.5$ and $0.7$; and the longest prompt, $39{,}418$ characters ($\approx 10{,}000$ tokens) against a million-token window, leaves only the $65{,}535$-token output cap to bind. Unlike the ML exam's few-shot Flash-Lite run (Appendix~\ref{app:replication}), the dropout is neither length- nor ability-biased: excluded notebooks are no longer than survivors' (P$02$ $29{,}433$ vs.\ $31{,}625$ characters, Welch $p = 0.35$; E$01$'s $14$, $31{,}657$ vs.\ $31{,}772$, $p = 0.90$) and their grader-average totals match ($26.9$ vs.\ $26.0$ out of $35$, $p = 0.69$; $26.0$ vs.\ $26.2$, $p = 0.91$). E-series per-question scores are the mean over all recorded reruns (five per cell by design; some Gemini cells logged more) and the total is the Q$1$--Q$3$ base sum as elsewhere.

\paragraph{Item-level agreement.}
Table~\ref{tab:itemcorr} gives per-question and total-level Spearman $\rho$ between the AI mark and the grader-average mark for the headline configurations of both exams.

\begin{table}[t]
\centering
\caption{Item-level agreement for both exams' headline configurations: Spearman $\rho$ between the AI mark and the grader-average mark, per question and at total level, over the same students each run's MAE uses. The total is the optimistic number: across the $40$ CV and $38$ ML runs measured here (the rows below plus each exam's $17$ matched neutral/strict pairs; Table~\ref{tab:brittle}, Section~\ref{sec:brittle:replication}), it exceeds the weakest question's $\rho$ in all $37$ CV and $37$ ML runs where every correlation is defined --- median gap $0.20$ (CV), $0.12$ (ML), and at least $0.15$ in $30$ CV and $12$ ML runs. The other three CV and one ML runs give every student the same mark, leaving their correlations undefined rather than zero. The weak item is the same within an exam: Q$2$ on CV ($30$ of $37$), Q$1$ on ML ($27$ of $37$). Pearson $r$ differs from $\rho$ by a median of $0.02$ on both exams. Generated by \texttt{analysis/checks/per\_question\_rank\_correlation.py}.}
\label{tab:itemcorr}
\small
\begin{tabular}{lllrrrrr}
\toprule
Run & Model & Persona & $n$ & Q$1$ $\rho$ & Q$2$ $\rho$ & Q$3$ $\rho$ & Total $\rho$ \\
\midrule
\multicolumn{8}{l}{\emph{CV exam}} \\
D01 & gemini-3-flash-preview & \emph{neutral} & 570 & 0.92 & 0.90 & 0.91 & 0.95 \\
D02 & gemini-3.1-pro-preview & \emph{neutral} & 570 & 0.92 & 0.90 & 0.90 & 0.94 \\
F02 & gemini-3.1-pro-preview & \emph{lenient} & 570 & 0.92 & 0.90 & 0.91 & 0.94 \\
B03 & gemini-flash-lite + thinking & \emph{neutral} & 570 & 0.89 & 0.88 & 0.87 & 0.92 \\
A01 & gemini-flash-lite & \emph{neutral} & 570 & 0.77 & 0.65 & 0.83 & 0.84 \\
C01 & gemini-flash-lite & \emph{strict} & 570 & 0.78 & 0.59 & 0.81 & 0.83 \\
\midrule
\multicolumn{8}{l}{\emph{ML exam}} \\
IG08 & gemini-3.1-pro-preview & \emph{neutral} & 1038 & 0.88 & 0.91 & 0.93 & 0.93 \\
IG07 & gemini-3-flash-preview & \emph{neutral} & 1038 & 0.86 & 0.91 & 0.94 & 0.93 \\
IG01 & gemini-flash-lite & \emph{neutral} & 1013 & 0.77 & 0.85 & 0.85 & 0.88 \\
IG05 & gemini-flash-lite & \emph{strict} & 1008 & 0.78 & 0.81 & 0.86 & 0.88 \\
\bottomrule
\end{tabular}
\end{table}

\clearpage
\section{Strict-Persona Collapse: Additional Tables}
\label{app:brittle}

This appendix carries the full exhibits behind Section~\ref{sec:brittle}.

\begin{table}[t]
\centering
\caption{The \emph{strict} persona against each model's own \emph{neutral} baseline at an identical prompt configuration. MAE in points out of $35$; ratio is strict $\div$ neutral; behaviour classes as in Section~\ref{sec:brittle:numbers}; a refusal's MAE is a ceiling artifact and must not be read as severity. Every row is $n = 570$. Generated by \texttt{analysis/computer\_vision\_make\_paper\_tables.py}.}
\label{tab:brittle}
\small
\begin{tabular}{llllrrrl}
\toprule
Run & Model & Family & Params & Neutral & \textbf{Strict} & Ratio & Behaviour \\
\midrule
\multicolumn{8}{l}{\emph{Open weights --- ordered by total parameters}} \\
L-M05 & Qwen2.5-Coder-7B & Qwen & 7B & 5.68 & \textbf{7.92} & $\times 1.39$ & graded \\
L-U02 & Llama-3.1-8B & Llama & 8B & 7.31 & \textbf{26.04} & $\times 3.56$ & \textbf{refusal} \\
L-W02 & GLM-4-9B & GLM & 9B & 4.31 & \textbf{25.48} & $\times 5.91$ & near-refusal \\
L-F02 & Gemma-3-12B & Gemma & 12B & 5.20 & \textbf{8.77} & $\times 1.69$ & collapse \\
L-M07 & Qwen2.5-Coder-14B & Qwen & 14B & 3.51 & \textbf{24.52} & $\times 6.99$ & collapse \\
L-S02 & DeepSeek-Coder-V2-Lite & DeepSeek & 16B & 5.98 & \textbf{12.09} & $\times 2.02$ & collapse \\
L-Z02 & Mistral-Small-24B & Mistral & 24B & 3.66 & \textbf{26.03} & $\times 7.11$ & \textbf{refusal} \\
L-Y02 & Gemma-3-27B & Gemma & 27B & 4.32 & \textbf{11.55} & $\times 2.67$ & collapse \\
L-M01 & Qwen3-Coder-30B-A3B & Qwen & 30B & 4.50 & \textbf{14.91} & $\times 3.31$ & collapse \\
L-X02 & GLM-4-32B & GLM & 32B & 2.85 & \textbf{9.66} & $\times 3.39$ & collapse \\
L-C01 & Qwen2.5-Coder-32B & Qwen & 32B & 7.49 & \textbf{20.29} & $\times 2.71$ & collapse \\
L-V02 & Llama-3.3-70B & Llama & 70B & 4.52 & \textbf{13.52} & $\times 2.99$ & collapse \\
L-N02 & Qwen2.5-72B & Qwen & 72B & 3.14 & \textbf{9.82} & $\times 3.13$ & collapse \\
L-M03 & Qwen3-Coder-Next 80B & Qwen & 80B & 5.17 & \textbf{11.70} & $\times 2.26$ & collapse \\
L-J02 & GLM-4.5-Air & GLM & 106B & 5.66 & \textbf{21.02} & $\times 3.71$ & collapse \\
L-Q02 & Qwen3-235B-A22B & Qwen & 235B & 4.09 & \textbf{7.32} & $\times 1.79$ & graded \\
L-R02 & Qwen3-Coder-480B & Qwen & 480B & 3.14 & \textbf{3.78} & $\times 1.20$ & graded \\
\midrule
\multicolumn{8}{l}{\emph{Closed --- reference}} \\
C01 & Flash-Lite & Gemini & --- & 3.34 & 5.75 & $\times 1.72$ & graded \\
F03 & 3.1 Pro Preview & Gemini & --- & 1.86 & 2.75 & $\times 1.48$ & graded \\
\bottomrule
\end{tabular}
\end{table}

\begin{figure}[t]
\centering
\includegraphics[width=\linewidth]{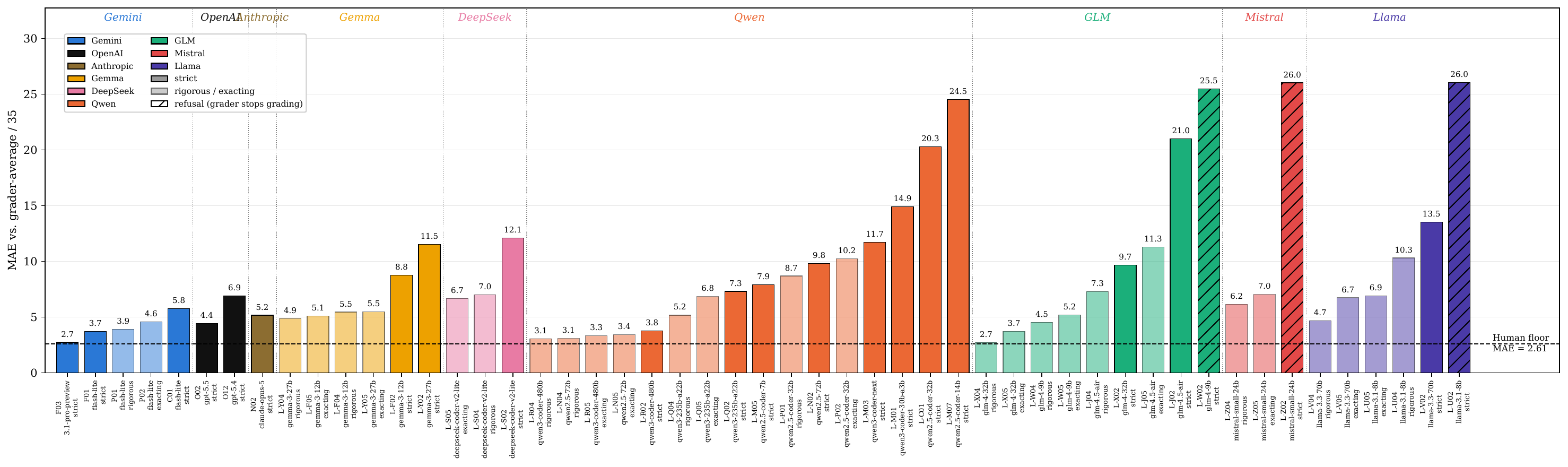}
\caption{Every strict-flavoured persona run on the full cohort ($51$ runs; $50$ at the default prompt configuration, plus F$01$, which additionally enables thinking), coloured by family and by wording (\emph{strict} darker; \emph{rigorous}/\emph{exacting} lighter). The dashed line is the human inter-grader floor ($\text{MAE} = 2.61$). Runs classified as refusals are marked: their bar height is the distance to the grader mean and is not a severity measurement (Section~\ref{sec:brittle:numbers}). The two default-configuration strict runs not shown are M$01$ ($n = 99$, below the full-cohort cutoff) and F$03$ (\texttt{gemini-3.1-pro-preview}, $2.75$; Table~\ref{tab:brittle}).}
\label{fig:brittle}
\end{figure}

\begin{table}[t]
\centering
\caption{The three strict-flavoured presets on the models that received all of them. Bold marks a run outside the graded band. These presets are \textbf{not} minimal pairs: only \emph{strict} carries the two policy sentences (Section~\ref{sec:brittle:attrib}); \emph{rigorous} and \emph{exacting} ask for rubric precision with neither. Qwen2.5-Coder-$7$B, Qwen3-Coder-$30$B-A$3$B and Qwen3-Coder-Next received only \emph{strict} in the full-cohort grid; Section~\ref{sec:finetune:personas} covers the first two.}
\label{tab:wording}
\small
\begin{tabular}{lllrrrr}
\toprule
Model & Family & Params & Neutral & \emph{strict} & \emph{rigorous} & \emph{exacting} \\
\midrule
Flash-Lite & Gemini & --- & 3.34 & 5.75 & 3.92 & 4.58 \\
\midrule
Llama-3.1-8B & Llama & 8B & 7.31 & \textbf{26.04} & \textbf{10.31} & 6.89 \\
GLM-4-9B & GLM & 9B & 4.31 & \textbf{25.48} & 4.53 & 5.18 \\
Gemma-3-12B & Gemma & 12B & 5.20 & \textbf{8.77} & 5.45 & 5.10 \\
DeepSeek-Coder-V2-Lite & DeepSeek & 16B & 5.98 & \textbf{12.09} & 7.01 & 6.69 \\
Mistral-Small-24B & Mistral & 24B & 3.66 & \textbf{26.03} & 6.15 & 7.04 \\
Gemma-3-27B & Gemma & 27B & 4.32 & \textbf{11.55} & 4.88 & 5.47 \\
GLM-4-32B & GLM & 32B & 2.85 & \textbf{9.66} & 2.74 & 3.72 \\
Qwen2.5-Coder-32B & Qwen & 32B & 7.49 & \textbf{20.29} & \textbf{8.70} & \textbf{10.23} \\
Llama-3.3-70B & Llama & 70B & 4.52 & \textbf{13.52} & 4.69 & 6.74 \\
Qwen2.5-72B & Qwen & 72B & 3.14 & \textbf{9.82} & 3.09 & 3.43 \\
GLM-4.5-Air & GLM & 106B & 5.66 & \textbf{21.02} & 7.29 & \textbf{11.29} \\
Qwen3-235B-A22B & Qwen & 235B & 4.09 & 7.32 & 5.18 & 6.85 \\
Qwen3-Coder-480B & Qwen & 480B & 3.14 & 3.78 & 3.06 & 3.33 \\
\bottomrule
\end{tabular}
\end{table}

M$01$ ($n = 99$, the only strict run on \texttt{gemini-3-flash-preview}) has a CI of $[3.60, 4.88]$ spanning most of the Gemini cluster of Section~\ref{sec:brittle:numbers}, so we do not rank it within that cluster.

\begin{table}[h]
\centering
\caption{Attribution of the collapse to the policy sentences rather than the headword. \emph{Left:} S1 and S2 held verbatim, only the adjective varies; the \emph{nharsh} control is the neutral frame with its one adjective swapped to ``harsh''. \emph{Right:} the HARSH frame held fixed, crossing S1 with S2 (both quoted in Section~\ref{sec:brittle:attrib}). All rows $n \ge 569$, default prompt configuration, $t = 0$. Spreads are max $-$ min of the unrounded MAEs, so they can differ from the difference of the printed cells by $0.01$. Bold marks cells whose damage matches the full preset's.}
\label{tab:attrib}
\small
\resizebox{\linewidth}{!}{%
\begin{tabular}{lrrrrr@{\hskip 2em}rrrr}
\toprule
& \multicolumn{5}{c}{headword varied, policy fixed} & \multicolumn{4}{c}{policy varied, frame fixed} \\
\cmidrule(r){2-6}\cmidrule(l){7-10}
Model & STRICT & RIGOROUS & FAIR & spread & \emph{nharsh} & frame & $+$S1 & $+$S2 & both \\
\midrule
Qwen2.5-Coder-32B  & 16.41 & 17.74 & 18.55 & 2.15 & 8.25 & 9.37 & 15.55 & 12.92 & \textbf{20.26} \\
Llama-3.3-70B      & 12.67 & 11.37 & 11.41 & 1.30 & 4.60 & ---  & ---   & ---   & --- \\
Gemma-3-27B        & 10.49 &  9.23 &  8.60 & 1.89 & 4.56 & ---  & ---   & ---   & --- \\
GLM-4-32B          &  7.02 &  7.43 &  6.41 & 1.02 & 2.88 & ---  & ---   & ---   & --- \\
Mistral-Small-24B  & 25.99 & 25.87 & 25.05 & 0.94 & 4.23 & 5.62 & 10.33 & \textbf{26.04} & \textbf{26.03} \\
Llama-3.1-8B       & ---   & ---   & ---   & ---  & ---  & 8.46 & 23.45 & \textbf{26.04} & \textbf{26.04} \\
\bottomrule
\end{tabular}}
\end{table}

\begin{table}[h]
\centering
\caption{How each matched \emph{strict} run fails. Q$1 = 0$ is the count of students awarded zero on Q$1$; the next column is how many of \emph{those} still received a non-zero Q$2$. Generated by \texttt{analysis/computer\_vision\_make\_paper\_tables.py}.}
\label{tab:mechanism}
\small
\begin{tabular}{lllrrr}
\toprule
Run & Model & Family & MAE & Q$1 = 0$ & \ldots of which Q$2 > 0$ \\
\midrule
\multicolumn{6}{l}{\emph{blanket zeroing --- consistent with refusal}} \\
L-U02 & Llama-3.1-8B & Llama & 26.04 & 570 / 570 (100\%) & 0 (0\%) \\
L-Z02 & Mistral-Small-24B & Mistral & 26.03 & 570 / 570 (100\%) & 0 (0\%) \\
L-W02 & GLM-4-9B & GLM & 25.48 & 570 / 570 (100\%) & 29 (5\%) \\
L-M07 & Qwen2.5-Coder-14B & Qwen & 24.52 & 533 / 570 (94\%) & 31 (6\%) \\
L-J02 & GLM-4.5-Air & GLM & 21.02 & 361 / 570 (63\%) & 55 (15\%) \\
\midrule
\multicolumn{6}{l}{\emph{selective field collapse --- consistent with conflicting instructions (a hypothesis; Section~\ref{sec:brittle:mech})}} \\
L-C01 & Qwen2.5-Coder-32B & Qwen & 20.29 & 437 / 570 (77\%) & 284 (65\%) \\
L-M01 & Qwen3-Coder-30B-A3B & Qwen & 14.91 & 357 / 570 (63\%) & 274 (77\%) \\
L-V02 & Llama-3.3-70B & Llama & 13.52 & 291 / 570 (51\%) & 183 (63\%) \\
L-N02 & Qwen2.5-72B & Qwen & 9.82 & 136 / 570 (24\%) & 97 (71\%) \\
L-X02 & GLM-4-32B & GLM & 9.66 & 273 / 570 (48\%) & 139 (51\%) \\
\midrule
\multicolumn{6}{l}{\emph{uniform severity --- consistent with calibration drift}} \\
L-S02 & DeepSeek-Coder-V2-Lite & DeepSeek & 12.09 & 1 / 570 (0\%) & 1 (100\%) \\
L-M03 & Qwen3-Coder-Next 80B & Qwen & 11.70 & 19 / 570 (3\%) & 8 (42\%) \\
L-Y02 & Gemma-3-27B & Gemma & 11.55 & 0 / 570 (0\%) & n/a \\
L-F02 & Gemma-3-12B & Gemma & 8.77 & 1 / 570 (0\%) & 0 (0\%) \\
L-M05 & Qwen2.5-Coder-7B & Qwen & 7.92 & 8 / 570 (1\%) & 6 (75\%) \\
L-Q02 & Qwen3-235B-A22B & Qwen & 7.32 & 5 / 570 (1\%) & 1 (20\%) \\
C01 & Flash-Lite & Gemini & 5.75 & 31 / 570 (5\%) & 21 (68\%) \\
L-R02 & Qwen3-Coder-480B & Qwen & 3.78 & 1 / 570 (0\%) & 1 (100\%) \\
F03 & 3.1 Pro Preview & Gemini & 2.75 & 8 / 570 (1\%) & 5 (62\%) \\
\bottomrule
\end{tabular}
\end{table}

The three near-total zeroers have awarded-total standard deviations of $0.00$, $0.09$ and $2.42$ and AI--grader correlations undefined, $-0.01$ and $0.19$. GLM-4.5-Air blanket-zeroes $63\%$ of students and still grades the rest ($r = 0.53$), which is why Table~\ref{tab:brittle} classes it a collapse rather than a refusal. Qwen2.5-Coder-$14$B (L-M$07$) is the second intermediate case: it zeroes $88\%$ of whole submissions (mean awarded total $1.52$, $r = 0.25$), just under the $90\%$ refusal cut. Q$1$-zero rates fall at $0$--$5\%$ or $24$--$100\%$ with nothing between, and among the latter the selective fraction is $0$--$15\%$ or $51$--$77\%$.

\begin{table}[h]
\centering
\caption{Self-contradiction between the structured Q$1$ score and the prose breakdown, on the two strict-persona runs whose rationales we annotated by hand. A row is self-contradicting when the structured score is $0$ but the rationale itemises positive partial credit. Both runs sit in the selective-field-collapse group of Table~\ref{tab:mechanism} and match its Q$1 = 0$ counts; the second column differs because this one requires reading the prose, which we did not do at scale.}
\label{tab:contradict}
\small
\resizebox{\linewidth}{!}{%
\begin{tabular}{lrrr}
\toprule
Run & Q$1$ score = 0 & ...with prose $> 0$ & mean contradicted prose-credit \\
\midrule
L-C01 (Qwen2.5-Coder-32B + strict) & 437 / 570 (77\%) & 86 (20\%) & 2.60 pts (max 10.0) \\
L-M01 (Qwen3-Coder-30B-A3B + strict) & 357 / 570 (63\%) & 152 (43\%) & 2.75 pts (max 8.5) \\
\bottomrule
\end{tabular}}
\end{table}

\paragraph{Representative example.}
The clearest case in our sample is L-C$01$ student~$\#2$, where both graders gave Q$1 = 12.0 / 12$ and the model emitted Q$1 = 0$. The reasoning blob begins:
\begin{quote}\itshape
The student completed most tasks correctly, including defining the transforms, creating DataLoaders, \ldots\ However, the backbone was not properly frozen, and the bonus task for Test Time Augmentation was not implemented.
\end{quote}
The same blob then itemises a breakdown (``Complete transforms: 1.5'', ``Create DataLoaders: 0.5'', \ldots) summing to $10.0$ points awarded in prose --- the largest contradicted total in Table~\ref{tab:contradict}.

\paragraph{A prediction of the conflict account that fails.}
If the collapse is driven by a conflict between S2 and the rubric breakdown, withdrawing the breakdown should remove one side of the conflict and \emph{relieve} the collapse. Removing the breakdown under \emph{strict} worsens MAE by $+6.42$ on Llama-3.3-70B ($13.52 \to 19.94$), $+5.92$ on GLM-4-32B ($9.66 \to 15.58$), $+3.93$ on DeepSeek-Coder-V2-Lite ($12.09 \to 16.02$), $+1.40$ on Qwen2.5-Coder-32B ($20.29 \to 21.69$) and $+0.15$ on Gemma-3-27B ($11.55 \to 11.69$) --- five of five in the wrong direction. The same manipulation under \emph{neutral} is a clean control, with a largest shift of $+0.85$ and four of five models within $\pm 0.25$, so this is specific to the persona rather than an artifact of a shorter prompt.

The fairest reading is that this weakens the causal story without settling on an alternative; the structural-scaffolding reading of Section~\ref{sec:brittle:mech} fits the L-BD result but is itself untested. Withdrawing the breakdown leaves the rubric table in the prompt, so the conflict is attenuated rather than eliminated; a fully clean test would strip the rubric too, which we have not run. We flag this as open rather than resolved (Section~\ref{sec:limitations}).

\clearpage
\paragraph{Prompt-only baselines.}
\label{app:promptbaselines}
Before fine-tuning, we asked whether the \emph{strict}-persona collapse can be undone by prompting alone. We tried the two repairs the literature suggests on the four CV-exam models whose \emph{strict} runs collapse or refuse (Qwen2.5-Coder-$32$B, GLM-4-$32$B, Mistral-Small-$24$B, Llama-3.1-$8$B), over the first $100$ students (the G-series subset), Q$1$--Q$3$, with the paper's vLLM stack, and compared each repair against the same model's own \emph{strict} and neutral runs restricted to the same students (Table~\ref{tab:promptbaselines}).

The first repair is \emph{decomposition}~\citep{pathak2025rubricneedenhancingllmbased}: instead of grading a whole question in one call, the grader makes one call per rubric task row ($16$/$13$/$11$ rows for Q$1$/Q$2$/Q$3$; penalty and bonus rows handled) and awards only that row's marks; the row awards are summed and clamped to the question maxima. The persona, guidelines, reference solution and submission are exactly those of the paper's prompt. The hope is that a small, concrete criterion leaves the model less room to zero a whole question. It does not. Qwen2.5-Coder-$32$B moves from $21.24$ to $20.00$ MAE, Mistral-Small-$24$B from $27.25$ to $26.27$, and Llama-3.1-$8$B refuses on every row exactly as it refused on every question ($27.26$). Only GLM-4-$32$B re-enters the graded band ($8.90 \to 5.01$), and even then it sits at $2.2\times$ its neutral error. The credit-withholding sentences bind on each criterion just as they bind on the question as a whole.

The second repair is \emph{grade-then-arbitrate}~\citep{chan2023chatevalbetterllmbasedevaluators}: the paper's \emph{strict} prompt and the paper's neutral prompt each grade the question, and a third call under the neutral persona receives both JSON grades together with the rubric and must resolve every disagreement on the rubric's partial-credit scale. The arbiter's grade is the run's grade, at three calls per question instead of one. This does bring every model back into the graded band, but only to where its own neutral prompt already was: GLM-4-$32$B $2.64$ against $2.24$ neutral, Mistral $3.13$ against $3.59$, Llama $7.46$ against $7.73$ (on $99$ students: one arbiter reply, student $87$, Q$2$, was truncated at $4{,}096$ tokens), Qwen $9.29$ against $8.02$. The arbiter defers to the neutral grade, so arbitration repairs the persona by removing it, at three times the cost, and gains nothing beyond the neutral prompt.

Neither repair therefore reaches what fine-tuning reaches. The neutral level the repairs recover is $2.2$--$8.0$ MAE on these four models; the pooled adapters of Section~\ref{sec:finetune} start from that level and end at $1.75$--$2.01$ on the held-out CV students, below the human floor. Scripts: \texttt{prompt\_baselines\_grade.py}, \texttt{analysis/checks/prompt\_baselines\_vs\_paper.py}.

\begin{table}[h]
\centering
\caption{Prompt-only repairs under the \emph{strict} persona, CV exam, first $100$ students, $35$-point base scale; the paper's strict and neutral runs are restricted to the same students. MAE with $95\%$ bootstrap CI ($2000$ resamples); zero = share of students awarded a total of $0$; behaviour classes as in Section~\ref{sec:brittle:numbers}. Human floor on these students: $3.43$.}
\label{tab:promptbaselines}
\small
\resizebox{\linewidth}{!}{%
\begin{tabular}{llrrrl}
\toprule
Model & Run & MAE [CI] & Bias & Zero & Behaviour \\
\midrule
Qwen2.5-Coder-32B & paper neutral (L-A$32$) & 8.02 [7.38, 8.68] & $-7.96$ & 0\% & collapse \\
 & paper strict (L-C$01$) & 21.24 [20.31, 22.14] & $-21.24$ & 18\% & collapse \\
 & decomposed, strict & 20.00 [19.10, 20.93] & $-20.00$ & 3\% & collapse \\
 & arbitrated, strict & 9.29 [8.58, 9.98] & $-9.27$ & 0\% & collapse \\
\midrule
GLM-4-32B & paper neutral (L-X$01$) & 2.24 [1.91, 2.59] & $-0.17$ & 0\% & graded \\
 & paper strict (L-X$02$) & 8.90 [7.81, 10.03] & $-8.81$ & 5\% & collapse \\
 & decomposed, strict & 5.01 [4.50, 5.51] & $-4.95$ & 0\% & graded \\
 & arbitrated, strict & 2.64 [2.23, 3.10] & $-1.53$ & 0\% & graded \\
\midrule
Mistral-Small-24B & paper neutral (L-Z$01$) & 3.59 [3.13, 4.08] & $-2.81$ & 0\% & graded \\
 & paper strict (L-Z$02$) & 27.25 [25.88, 28.64] & $-27.25$ & 99\% & \textbf{refusal} \\
 & decomposed, strict & 26.27 [24.98, 27.55] & $-26.27$ & 56\% & collapse \\
 & arbitrated, strict & 3.13 [2.74, 3.53] & $-2.31$ & 0\% & graded \\
\midrule
Llama-3.1-8B & paper neutral (L-U$01$) & 7.77 [6.88, 8.61] & $-6.24$ & 0\% & graded \\
 & paper strict (L-U$02$) & 27.26 [25.89, 28.65] & $-27.26$ & 100\% & \textbf{refusal} \\
 & decomposed, strict & 27.26 [25.89, 28.65] & $-27.26$ & 100\% & \textbf{refusal} \\
 & arbitrated, strict ($n = 99$) & 7.46 [6.60, 8.28] & $-5.77$ & 0\% & graded \\
\bottomrule
\end{tabular}}
\end{table}

\clearpage
\section{Closed-Model Details}
\label{app:closed}

\begin{table}[t]
\centering
\caption{Closed models from three vendors under \emph{neutral}, \emph{strict} and \emph{lenient} on both exams (Section~\ref{sec:closed:vendors}): MAE against the grader average with $95\%$ bootstrap CI, and bias. Gemini rows are grid runs (D$02$/F$03$/F$02$ and A$01$/C$01$/C$02$ on the CV exam; IG$08$/IG$23$/IG$14$ and IG$01$/IG$05$/IG$06$ on the ML exam, with their $n$ as in Tables~\ref{tab:allruns} and~\ref{tab:iaruns}). OpenAI and Anthropic rows were graded on the full cohorts through the vendors' batch APIs with the identical prompt, reasoning off and temperature $0$ (OpenAI) or thinking disabled (Anthropic, whose API exposes no temperature); every reply parsed. Bold marks a run outside the graded band; no zero-total rate exceeds $1.1\%$. Generated by \texttt{analysis/closed\_vendor\_table.py}.}
\label{tab:closedvendors}
\small
\resizebox{\linewidth}{!}{
\begin{tabular}{llrlrl}
\toprule
Model & Persona & \multicolumn{2}{c}{CV exam (floor $2.61$)} & \multicolumn{2}{c}{ML exam (floor $5.13$)} \\
\cmidrule(lr){3-4}\cmidrule(lr){5-6}
 & & MAE [95\% CI] & bias & MAE [95\% CI] & bias \\
\midrule
claude-opus-5 & \emph{neutral} & $3.54$ $[3.31,\,3.77]$ & $-3.24$ & $4.37$ $[4.13,\,4.63]$ & $-2.64$ \\
 & \emph{strict} & $5.17$ $[4.90,\,5.45]$ & $-5.02$ & --- & --- \\
\midrule
gemini-3.1-pro-preview & \emph{neutral} & $1.86$ $[1.70,\,2.02]$ & $-0.82$ & $3.40$ $[3.16,\,3.65]$ & $+1.18$ \\
 & \emph{strict} & $2.75$ $[2.51,\,2.99]$ & $-2.09$ & $3.67$ $[3.43,\,3.94]$ & $-1.00$ \\
 & \emph{lenient} & $1.79$ $[1.62,\,1.95]$ & $+0.82$ & $4.48$ $[4.22,\,4.75]$ & $+3.46$ \\
\midrule
gemini-flash-lite & \emph{neutral} & $3.34$ $[3.15,\,3.53]$ & $-1.78$ & $7.53$ $[7.20,\,7.87]$ & $+5.98$ \\
 & \emph{strict} & $5.75$ $[5.46,\,6.04]$ & $-5.44$ & $9.02$ $[8.61,\,9.43]$ & $-7.83$ \\
 & \emph{lenient} & $4.49$ $[4.18,\,4.82]$ & $+4.01$ & $\mathbf{17.86}$ $[17.30,\,18.42]$ & $+17.80$ \\
\midrule
gpt-5.4 & \emph{neutral} & $4.69$ $[4.46,\,4.92]$ & $-4.45$ & $4.30$ $[4.07,\,4.55]$ & $-1.95$ \\
 & \emph{strict} & $6.90$ $[6.63,\,7.15]$ & $-6.81$ & $5.11$ $[4.85,\,5.37]$ & $-3.61$ \\
 & \emph{lenient} & $2.69$ $[2.51,\,2.89]$ & $+1.00$ & $8.30$ $[7.96,\,8.64]$ & $+7.94$ \\
\midrule
gpt-5.5 & \emph{neutral} & $2.43$ $[2.28,\,2.59]$ & $-1.65$ & $3.54$ $[3.33,\,3.77]$ & $+1.23$ \\
 & \emph{strict} & $4.43$ $[4.19,\,4.68]$ & $-4.25$ & $3.70$ $[3.48,\,3.93]$ & $-1.38$ \\
 & \emph{lenient} & $2.25$ $[2.08,\,2.44]$ & $+1.13$ & $6.29$ $[6.02,\,6.57]$ & $+5.83$ \\
\bottomrule
\end{tabular}
}
\end{table}

\begin{table}[t]
\centering
\caption{The six full-cohort configurations, and five first-$100$ runs, whose point-estimate MAE meets or undercuts the human inter-grader floor, with $95\%$ bootstrap CIs ($2000$ resamples), plus reference rows. The G-series and M02 rows cover only the first $\sim 100$ students, an easier subset (the unmodified D01 recipe scores $1.23$ there vs.\ $1.64$ on the full cohort; Appendix~\ref{sec:closed:gseries}); they are not comparable to the full-cohort rows, and M02's CI upper bound crosses the floor.}
\label{tab:headline}
\small
\resizebox{\linewidth}{!}{%
\begin{tabular}{llrrrr}
\toprule
Run & Model & $n$ & MAE & $95\%$ CI & Bias \\
\midrule
\multicolumn{6}{l}{\emph{Human inter-grader floor (Section~\ref{sec:floor})}} \\
--- & G${}_1$ vs.\ G${}_2$ & 570 & 2.61 & $[2.37,\ 2.85]$ & --- \\
\midrule
\multicolumn{6}{l}{\emph{Full-cohort configurations}} \\
D01 & gemini-3-flash-preview, neutral & 570 & \textbf{1.64} & $[1.51,\ 1.79]$ & $+0.01$ \\
F02 & gemini-3.1-pro-preview + lenient & 570 & 1.79 & $[1.62,\ 1.95]$ & $+0.82$ \\
D02 & gemini-3.1-pro-preview, neutral & 570 & 1.86 & $[1.70,\ 2.02]$ & $-0.82$ \\
B03 & gemini-flash-lite + thinking & 570 & 2.05 & $[1.89,\ 2.21]$ & $-0.15$ \\
O03 & gpt-5.5 + lenient (batch API) & 570 & 2.25 & $[2.08,\ 2.44]$ & $+1.13$ \\
O01 & gpt-5.5, neutral (batch API) & 570 & 2.43 & $[2.28,\ 2.59]$ & $-1.65$ \\
\midrule
\multicolumn{6}{l}{\emph{First-$100$-students subset (easier subset; see caption)}} \\
D01 (restricted) & gemini-3-flash-preview, baseline recipe on the same $100$ students & 100 & 1.23 & $[1.02,\ 1.49]$ & $-0.18$ \\
G04 & 3-flash-preview, no rubric breakdown & 100 & 1.25 & $[1.01,\ 1.51]$ & $-0.07$ \\
G03 & 3-flash-preview + thinking & 100 & 1.39 & $[1.18,\ 1.61]$ & $-0.12$ \\
G02 & 3-flash-preview, no guidelines & 99 & 1.60 & $[1.30,\ 1.93]$ & $+0.79$ \\
G01 & 3-flash-preview, no reference solution & 98 & 1.63 & $[1.32,\ 2.02]$ & $+0.71$ \\
M02 & 3-flash-preview + lenient & 100 & 2.18 & $[1.75,\ 2.65]$ & $+1.90$ \\
\midrule
\multicolumn{6}{l}{\emph{Reference rows}} \\
A01 & gemini-flash-lite (baseline recipe) & 570 & 3.34 & $[3.15,\ 3.53]$ & $-1.78$ \\
L-R04 & Qwen3-Coder-480B + rigorous (best robust open) & 570 & 3.06 & $[2.88,\ 3.26]$ & $-0.23$ \\
\bottomrule
\end{tabular}}
\end{table}

\begin{table}[t]
\centering
\caption{The paired third-grader test (Eq.~\eqref{eq:paired}) applied to the zero-shot configurations, on the full $570$-student CV cohort: the same statistic and decision rule as Table~\ref{tab:ftpaired}, the standard the fine-tuned models face. \emph{better} means the whole $95\%$ CI of $\overline{d}$ sits below zero, ``$=$ grader'' that it contains zero, \emph{worse} that it lies above. CIs are the normal approximation used in Table~\ref{tab:ftpaired}; a $2000$-resample percentile bootstrap gives the same verdict in every row. With the floor at $2.61$, $\overline{d} = \tfrac{1}{2}(|\mathrm{AI}-\mathrm{G}_1| + |\mathrm{AI}-\mathrm{G}_2|) - 2.61$ exactly. The two single-grader distances differ by up to $0.63$ here, so both are shown. Generated by \texttt{analysis/checks/paired\_test\_zero\_shot\_cv.py}.}
\label{tab:zeroshotpaired}
\small
\setlength{\tabcolsep}{4.5pt}
\resizebox{\linewidth}{!}{%
\begin{tabular}{llrrrrl}
\toprule
Run & Configuration & $|\mathrm{AI}{-}\mathrm{G}_{\mathrm{avg}}|$ & $|\mathrm{AI}{-}\mathrm{G}_1|$ & $|\mathrm{AI}{-}\mathrm{G}_2|$ & $\overline{d}$ [$95\%$ CI] & Verdict \\
\midrule
D01 & 3-flash-preview, neutral        & 1.64 & 2.23 & 1.92 & $-0.54$ $[-0.71, -0.36]$ & \textbf{better} \\
F02 & 3.1-pro-preview $+$ lenient     & 1.79 & 2.32 & 2.15 & $-0.37$ $[-0.56, -0.18]$ & \textbf{better} \\
D02 & 3.1-pro-preview, neutral        & 1.86 & 2.53 & 1.96 & $-0.37$ $[-0.55, -0.18]$ & \textbf{better} \\
B03 & flash-lite $+$ thinking         & 2.05 & 2.55 & 2.29 & $-0.19$ $[-0.39, +0.01]$ & $=$ grader \\
O03 & \texttt{gpt-5.5} $+$ lenient (Section~\ref{sec:closed:vendors}) & 2.25 & 2.59 & 2.62 & $-0.01$ $[-0.22, +0.21]$ & $=$ grader \\
O01 & \texttt{gpt-5.5}, neutral (Section~\ref{sec:closed:vendors}) & 2.43 & 3.10 & 2.51 & $+0.19$ $[-0.02, +0.41]$ & $=$ grader \\
\midrule
\multicolumn{7}{l}{\emph{Reference rows (do not clear the floor)}} \\
A01 & flash-lite, baseline recipe     & 3.34 & 3.83 & 3.46 & $+1.03$ $[+0.77, +1.29]$ & worse \\
N01 & \texttt{claude-opus-5}, neutral (Section~\ref{sec:closed:vendors}) & 3.54 & 4.05 & 3.42 & $+1.13$ $[+0.88, +1.37]$ & worse \\
\bottomrule
\end{tabular}}
\end{table}

\begin{figure}[h]
\centering
\includegraphics[width=0.66\linewidth]{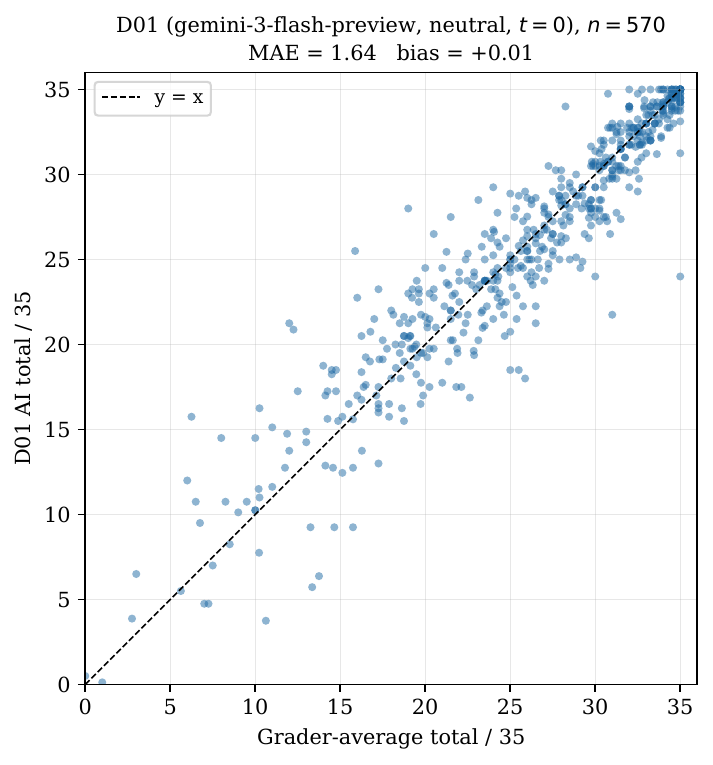}
\caption{D01 (\texttt{gemini-3-flash-preview}, neutral, $t=0$): AI total vs.\ grader-average total over the full $n = 570$ submissions, with $y = x$ as the perfect-agreement reference. Bias is $+0.01$, and residual error concentrates in the under-$15$ band, where graders also disagree more.}
\label{fig:d01scatter}
\end{figure}

\begin{table}[h]
\centering
\caption{Prompt-component ablations on \texttt{gemini-flash-lite}. B03 vs.\ A01 measures dynamic-budget thinking against the baseline's disabled thinking (\texttt{thinking\_budget} $= 0$; Appendix~\ref{app:details}). $n = 570$ except B01 ($567$) and B02/B04 ($569$ each), which exclude students with a permanently failed grading call.}
\label{tab:bseries}
\small
\begin{tabular}{llrr}
\toprule
Run & Change vs.\ A01 & MAE & Bias \\
\midrule
A01 & --- (baseline) & 3.34 & $-1.78$ \\
B01 & drop reference solution & 4.13 & $+3.30$ \\
B02 & drop guidelines & 3.34 & $-1.42$ \\
B03 & thinking on (dynamic) & \textbf{2.05} & $-0.15$ \\
B04 & drop rubric breakdown & 3.22 & $-1.45$ \\
\bottomrule
\end{tabular}
\end{table}

\begin{table}[h]
\centering
\caption{G-series on \texttt{gemini-3-flash-preview}, all on the first-$100$ subset ($n = 100$, except G01/G02 at $98$/$99$ after excluding permanently failed grading calls). D01-restricted is the baseline recipe on the same students. Bootstrap CI on $2000$ resamples.}
\label{tab:gseries}
\small
\begin{tabular}{llrr}
\toprule
Run & Change vs.\ baseline & MAE & $95\%$ CI \\
\midrule
\textbf{D01 (restricted)} & --- (baseline) & \textbf{1.23} & $[1.02,\ 1.49]$ \\
G04 & drop rubric breakdown & 1.25 & $[1.01,\ 1.51]$ \\
G03 & thinking on (dynamic) & 1.39 & $[1.18,\ 1.61]$ \\
G02 & drop guidelines & 1.60 & $[1.30,\ 1.93]$ \\
G01 & drop reference solution & 1.63 & $[1.32,\ 2.02]$ \\
\bottomrule
\end{tabular}
\end{table}

\paragraph{The apparent G-series wins are a sample-size artifact ($n = 100$).}
\label{sec:closed:gseries}

The G-series replays the ablations on \texttt{gemini-3-flash-preview}, whose throughput limited these runs to the first $100$ students. Against D01's full-cohort $1.64$, G04 (no rubric breakdown, $1.25$) appears to win; restricting D01 to the same students gives $1.23$, so the baseline ties G04 within noise and leads the other three (Table~\ref{tab:gseries}). Dropping the reference solution or the guidelines costs about $0.4$ MAE, with CIs still grazing the baseline's.

\paragraph{Flash-Lite at $t = 0$ is deterministic here; sampling adds variance without gain.}
\label{sec:closed:variance}

We measured sampling variance on \texttt{gemini-flash-lite} (E01/E02/E03 at $t = 0.0/0.5/0.7$, five reruns per cell). As everywhere in this paper the statistics cover the three graded questions only (Q$4$ is bonus-only), leaving \num{1687}/\num{1701}/\num{1701} (student, question) cells per configuration. At $t = 0$ the model is fully deterministic: $100\%$ of cells have zero standard deviation across reruns. At $t = 0.5$ the median per-cell standard deviation is $0.65$ points and $22.3\%$ of cells have a standard deviation above one point; at $t = 0.7$ those figures rise to $0.76$ and $31.8\%$. Qwen2.5-Coder-$32$B and Qwen3-Coder-Next at $t = 0.5$ ($\sim 50$-student subset) show $0.76$ and $0.82$, comparable to Gemini at $t = 0.7$. Averaging the reruns does not beat a single $t = 0$ call: mean-of-reruns MAE is $3.38$ at $t = 0$, $3.51$ at $t = 0.5$, and $3.60$ at $t = 0.7$, versus $3.34$ for the single-call A01 baseline (the $0.04$ gap at $t = 0$ comes from E01's smaller cohort, not from nondeterminism). We therefore recommend a single call at $t = 0$ as the production default.

\clearpage
\section{Open-Weights Details}
\label{app:open}

This appendix carries the exhibits behind Section~\ref{sec:open}: neutral-prompt baselines for all $17$ variants and the few-shot comparison. L-A$32$ rationales often penalise valid stylistic divergence from the reference solution --- different variable names or layer orderings --- supporting the over-anchoring account of the $32$B regression informally.

\paragraph{Few-shot prompting moves the open model substantially.}
\label{sec:open:fewshot}
Two worked examples per question (Table~\ref{tab:fewshot}) improve Flash-Lite by $0.31$ MAE and Qwen3-Coder-$30$B-A$3$B by $\approx 1.16$; on the ML exam it nearly halves the same model's error ($11.12 \to 5.84$; Appendix~\ref{app:replication}). Excluding the five demonstration students changes nothing measurable: on the $562$ (Flash-Lite) and $565$ (Qwen3-Coder-$30$B-A$3$B) students left, the uplift is $+0.30$ $[0.05, 0.54]$ and $+1.16$ $[1.04, 1.30]$ against $+0.31$ and $+1.16$ on the full cohort, and the ML exam's arms match ($+5.28$ against $+5.28$ for the $30$B, $+3.56$ against $+3.55$ for Flash-Lite, less its three demonstration students).

\begin{table}[h]
\centering
\caption{Open-weights baselines on the default prompt (neutral persona, $t = 0$), ordered by family then total parameters. MAE in points out of $35$ with $95\%$ bootstrap CI; bias is mean signed error (AI $-$ grader-avg). Six vendors and both dense and mixture-of-experts architectures. All rows $n = 570$.}
\label{tab:openbaselines}
\small
\begin{tabular}{lllrrr}
\toprule
Run & Model & Params & MAE & $95\%$ CI & Bias \\
\midrule
L-S01 & DeepSeek-Coder-V2-Lite (MoE) & 16B & 5.98 & $[5.69,\ 6.29]$ & $-3.44$ \\
\midrule
L-W01 & GLM-4-9B                     & 9B   & 4.31 & $[4.02,\ 4.63]$ & $+1.68$ \\
L-X01 & GLM-4-32B                    & 32B  & \textbf{2.85} & $[2.66,\ 3.05]$ & $+0.50$ \\
L-J01 & GLM-4.5-Air (MoE)            & 106B & 5.66 & $[5.38,\ 5.95]$ & $-5.37$ \\
\midrule
L-F01 & Gemma-3-12B                  & 12B  & 5.20 & $[4.94,\ 5.48]$ & $-2.75$ \\
L-Y01 & Gemma-3-27B                  & 27B  & 4.32 & $[4.12,\ 4.55]$ & $-1.52$ \\
\midrule
L-U01 & Llama-3.1-8B                 & 8B   & 7.31 & $[6.97,\ 7.68]$ & $-5.15$ \\
L-V01 & Llama-3.3-70B                & 70B  & 4.52 & $[4.31,\ 4.73]$ & $-3.60$ \\
\midrule
L-Z01 & Mistral-Small-24B            & 24B  & 3.66 & $[3.46,\ 3.87]$ & $-2.41$ \\
\midrule
L-A07 & Qwen2.5-Coder-7B             & 7B   & 5.68 & $[5.39,\ 5.98]$ & $-4.84$ \\
L-A14 & Qwen2.5-Coder-14B            & 14B  & 3.51 & $[3.30,\ 3.74]$ & $-1.16$ \\
L-A30M & Qwen3-Coder-30B-A3B (MoE)   & 30B  & 4.50 & $[4.29,\ 4.72]$ & $-2.70$ \\
L-A32 & Qwen2.5-Coder-32B            & 32B  & 7.49 & $[7.19,\ 7.78]$ & $-7.35$ \\
L-N01 & Qwen2.5-72B (dense)          & 72B  & 3.14 & $[2.95,\ 3.35]$ & $+0.19$ \\
L-ANxt & Qwen3-Coder-Next 80B (MoE)  & 80B  & 5.17 & $[4.92,\ 5.42]$ & $-4.74$ \\
L-Q01 & Qwen3-235B-A22B (MoE)        & 235B & 4.09 & $[3.87,\ 4.32]$ & $-3.56$ \\
L-R01 & Qwen3-Coder-480B FP8 (MoE)   & 480B & 3.14 & $[2.95,\ 3.35]$ & $+0.51$ \\
\bottomrule
\end{tabular}
\end{table}

\begin{table}[h]
\centering
\caption{Two worked examples per question, drawn from the highest-agreement grader pair and labelled with D$01$'s scores and rationales (in-context distillation, not human exemplars); the five demonstration students remain in the evaluated cohort (Appendix~\ref{app:limitations}), though excluding them moves the uplift by at most $0.01$ (\texttt{analysis/checks/fewshot\_demo\_leakage.py}). Few-shot moves Gemini Flash-Lite by $\approx 0.3$ MAE and Qwen3-Coder-$30$B-A$3$B by $\approx 1.2$ MAE points.}
\label{tab:fewshot}
\small
\begin{tabular}{llrr}
\toprule
Run & Model & MAE & Bias \\
\midrule
A01 & gemini-flash-lite (no few-shot) & 3.34 & $-1.78$ \\
K01 & gemini-flash-lite + few-shot & 3.03 & $+1.09$ \\
L-A30M & Qwen3-Coder-30B-A3B (no few-shot) & 4.50 & $-2.70$ \\
L-K01 & Qwen3-Coder-30B-A3B + few-shot & 3.34 & $-1.77$ \\
\bottomrule
\end{tabular}
\end{table}

\clearpage
\section{Fine-Tuning Details}
\label{app:ft}

All cells use the pinned held-out students ($114$ CV / $208$ ML exam). Evaluation follows each exam's ground truth: the CV exam records score and bonus separately, so agreement is score-only on both sides; the ML exam folds bonus into the grade, so agreement compares AI score $+$ bonus to the grader total.

\begin{table}[t]
\centering
\caption{Worst MAE drift over the three harsh personas (\emph{strict}, \emph{rigorous}, \emph{exacting}): the largest signed difference persona $-$ neutral; when all three are negative it is the least negative, which is why Qwen3-Coder-$30$B-A$3$B's ML entry of $-1.26$ is its \emph{rigorous} drift while \emph{strict} sits at $-2.26$ (marks recipe; full grid in Table~\ref{tab:ftpersonafull}). $\dagger$: negative drift: the persona accidentally \emph{improved} the over-grading base (Section~\ref{sec:finetune:personas}). \emph{lenient}, swept after tuning, is in Table~\ref{tab:ftpersonafull}: the pooled marks adapters drift $+0.05$ to $+0.39$ except Qwen3-Coder-$30$B-A$3$B ($+1.02$ CV, $+1.81$ ML); bd $+0.20$ to $+1.64$.}
\label{tab:ftpersona}
\small
\begin{tabular}{lrrrr}
\toprule
& \multicolumn{2}{c}{CV exam} & \multicolumn{2}{c}{ML exam} \\
\cmidrule(lr){2-3}\cmidrule(lr){4-5}
Model & base & pooled & base & pooled \\
\midrule
Qwen2.5-Coder-7B    & $+5.40$  & $+0.02$ & $+0.22$  & $-0.04$ \\
Qwen2.5-Coder-14B   & $+19.75$ & $-0.02$ & $+1.13$  & $+0.32$ \\
Qwen3-Coder-30B-A3B & $+12.51$ & $+0.07$ & $-1.26^\dagger$ & $+0.29$ \\
Gemma-4-E4B         & $+5.81$  & $+0.05$ & $+6.53$  & $+0.30$ \\
Llama-3.1-8B        & $+17.28$ & $-0.05$ & $+18.62$ & $+0.03$ \\
\bottomrule
\end{tabular}
\end{table}

\paragraph{Hyperparameters and evaluation.}
We use LoRA with rank $16$, $\alpha = 32$, dropout $0.05$, and all-linear targets (attention-only for the $30$B MoE), in bf16 for two epochs at learning rate $2\times10^{-4}$ with cosine decay and effective batch size $16$, held fixed across four data-parallel A100s; Gemma-4 trains on a single A100, as its unused parameters per step trip DDP's reduction check. All evaluation is served by vLLM, including Gemma-4's adapter, which reproduces the HuggingFace-generation numbers to within $0.1$ MAE across its four re-run cells.

\begin{table}[t]
\centering
\caption{Held-out MAE vs.\ the grader average, mean over the five models, both recipes (the per-model marks grid is Table~\ref{tab:ftmain}). The pooled adapter is best or tied in all eight columns.}
\label{tab:ftmeans}
\small
\begin{tabular}{lrrrr}
\toprule
Graded on & base & CV-trained & ML-trained & pooled \\
\midrule
CV exam, marks   & 5.54 & 1.90 & 3.57 & \textbf{1.87} \\
CV exam, bd      & 4.73 & 2.13 & 3.52 & \textbf{2.01} \\
ML exam, marks & 7.95 & 5.36 & 3.54 & \textbf{3.41} \\
ML exam, bd    & 8.76 & 6.16 & 4.24 & \textbf{4.15} \\
\bottomrule
\end{tabular}
\end{table}

\begin{table}[t]
\centering
\caption{The full persona grid (MAE, marks recipe; per-persona bias and the bd grid are in the released workbooks). $\dagger$: the sign-inconsistency cell of Section~\ref{sec:finetune:personas} --- \emph{strict} improves the over-grading base. \emph{len.} is the inflating direction, swept after tuning: the pooled marks adapters drift $+0.05$ to $+0.39$ except Qwen3-$30$B ($+1.02$ CV, $+1.81$ ML, the latter above the $5.19$ floor); every pooled lenient drift is positive (bias $+1.0$ to $+4.4$). bd worst drifts: base up to $+18.5$ on either exam; pooled $\le +0.62$ under the three harsh wordings, with Llama-bd-\emph{strict} at $2.67$ (down from $4.08$ under the CV-exam-only adapter); under \emph{lenient} the bd adapters drift $+0.20$ to $+1.64$ (Qwen3-$30$B ML $4.07 \to 5.71$, Gemma-$4$ ML $4.22 \to 5.51$).}
\label{tab:ftpersonafull}
\small
\setlength{\tabcolsep}{3pt}
\begin{tabular}{llrrrrrrrrrr}
\toprule
& & \multicolumn{5}{c}{CV exam} & \multicolumn{5}{c}{ML exam} \\
\cmidrule(lr){3-7}\cmidrule(lr){8-12}
Model & Arm & neut. & strict & rigor. & exact. & len. & neut. & strict & rigor. & exact. & len. \\
\midrule
Qwen2.5-7B  & base   & 6.60 & 12.00 & 9.07 & 6.07 & 4.81 & 6.48 & 6.70 & 6.04 & 6.15 & 7.50 \\
            & pooled & 1.84 & 1.84 & 1.85 & 1.81 & 2.00 & 3.40 & 3.36 & 3.30 & 3.31 & 3.52 \\
Qwen2.5-14B & base   & 4.06 & \textbf{23.81} & 3.65 & 7.31 & 3.79 & 7.70 & 8.83 & 7.71 & 6.33 & 11.18 \\
            & pooled & 1.89 & 1.85 & 1.87 & 1.81 & 2.00 & 3.29 & 3.61 & 3.23 & 3.28 & 3.50 \\
Qwen3-30B   & base   & 4.76 & 17.28 & 4.96 & 5.76 & 4.49 & 8.60 & 6.34$^\dagger$ & 7.34 & 6.30 & 13.00 \\
            & pooled & 1.75 & 1.80 & 1.79 & 1.82 & 2.78 & 3.45 & 3.73 & 3.58 & 3.72 & 5.26 \\
Gemma-4     & base   & 4.29 & 10.10 & 4.69 & 6.03 & 3.89 & 4.66 & 11.19 & 5.11 & 6.22 & 7.39 \\
            & pooled & 1.88 & 1.93 & 1.89 & 1.82 & 2.26 & 3.53 & 3.83 & 3.54 & 3.57 & 3.87 \\
Llama-8B    & base   & 7.99 & \textbf{25.28} & 12.35 & 11.95 & 6.86 & 12.30 & \textbf{30.92} & 7.16 & 7.97 & 29.23 \\
            & pooled & 2.01 & 1.96 & 1.93 & 1.91 & 2.05 & 3.37 & 3.39 & 3.26 & 3.26 & 3.46 \\
\bottomrule
\end{tabular}
\end{table}

\begin{table}[t]
\centering
\caption{Paired third-grader test (Eq.~\eqref{eq:paired}) for the pooled adapters. Floors: $2.83$ (CV), $5.19$ (ML). \emph{better}: the whole $95\%$ CI of $\overline{d}$ below zero; ``$=$ grader'': it contains zero. Both single-grader distances are shown: $|\mathrm{AI}-\mathrm{G}_1|$ is the larger in all ten ML rows and nine of ten CV rows, by up to $0.22$ (CV) and $0.57$ (ML). No paired-bootstrap difference excludes zero, and one ML student whose G$_1$ slot reads $0.0$ against a real G$_2$ mark (Appendix~\ref{app:replication}) alone accounts for $0.21$--$0.25$ of that gap, so we read the asymmetry as the grader-slot artifact, not a systematic preference. Three CV upper bounds fall inside $\pm 0.05$ and are printed to three decimals; all stay below zero under both the normal CI used here and a $2000$-resample percentile bootstrap, though 7B bd, CV is marginal (Section~\ref{sec:limitations}). Recomputed by \texttt{analysis/checks/ftpaired\_marginal\_verdicts.py} and \texttt{ftpaired\_symmetry\_guard.py}.}
\label{tab:ftpaired}
\small
\setlength{\tabcolsep}{4.5pt}
\begin{tabular}{llrrrrl}
\toprule
Cell & $n$ & $|\mathrm{AI}{-}\mathrm{G}_{\mathrm{avg}}|$ & $|\mathrm{AI}{-}\mathrm{G}_1|$ & $|\mathrm{AI}{-}\mathrm{G}_2|$ & $\overline{d}$ [$95\%$ CI] & Verdict \\
\midrule
7B marks, CV     & 114 & 1.84 & 2.35 & 2.35 & $-0.48$ $[-0.90, -0.05]$ & \textbf{better} \\
14B marks, CV    & 114 & 1.89 & 2.38 & 2.35 & $-0.46$ $[-0.89, -0.044]$ & \textbf{better} \\
30B marks, CV    & 114 & 1.75 & 2.38 & 2.17 & $-0.56$ $[-0.98, -0.13]$ & \textbf{better} \\
Gemma marks, CV  & 114 & 1.88 & 2.41 & 2.34 & $-0.45$ $[-0.89, -0.020]$ & \textbf{better} \\
Llama marks, CV  & 114 & 2.01 & 2.54 & 2.34 & $-0.39$ $[-0.85, +0.07]$ & $=$ grader \\
7B marks, ML     & 208 & 3.40 & 4.19 & 3.96 & $-1.11$ $[-1.73, -0.49]$ & \textbf{better} \\
14B marks, ML    & 208 & 3.29 & 4.11 & 4.02 & $-1.13$ $[-1.75, -0.50]$ & \textbf{better} \\
30B marks, ML    & 208 & 3.45 & 4.34 & 3.95 & $-1.04$ $[-1.65, -0.43]$ & \textbf{better} \\
Gemma marks, ML  & 208 & 3.53 & 4.45 & 4.05 & $-0.94$ $[-1.57, -0.30]$ & \textbf{better} \\
Llama marks, ML  & 208 & 3.37 & 4.33 & 4.03 & $-1.01$ $[-1.65, -0.36]$ & \textbf{better} \\
\midrule
7B bd, CV        & 114 & 1.94 & 2.41 & 2.35 & $-0.45$ $[-0.90, -0.001]$ & \textbf{better} \\
14B bd, CV       & 114 & 1.84 & 2.34 & 2.29 & $-0.51$ $[-0.93, -0.10]$ & \textbf{better} \\
30B bd, CV       & 114 & 1.99 & 2.57 & 2.35 & $-0.37$ $[-0.81, +0.08]$ & $=$ grader \\
Gemma bd, CV     & 114 & 2.22 & 2.53 & 2.58 & $-0.27$ $[-0.71, +0.17]$ & $=$ grader \\
Llama bd, CV     & 114 & 2.04 & 2.54 & 2.38 & $-0.37$ $[-0.80, +0.06]$ & $=$ grader \\
7B bd, ML        & 208 & 4.19 & 5.09 & 4.54 & $-0.37$ $[-1.06, +0.31]$ & $=$ grader \\
14B bd, ML       & 208 & 4.02 & 4.87 & 4.49 & $-0.51$ $[-1.16, +0.14]$ & $=$ grader \\
30B bd, ML       & 208 & 4.07 & 5.07 & 4.50 & $-0.40$ $[-1.08, +0.28]$ & $=$ grader \\
Gemma bd, ML     & 208 & 4.22 & 5.09 & 4.54 & $-0.37$ $[-1.05, +0.31]$ & $=$ grader \\
Llama bd, ML     & 208 & 4.25 & 5.05 & 4.71 & $-0.31$ $[-0.98, +0.37]$ & $=$ grader \\
\bottomrule
\end{tabular}
\end{table}

\paragraph{Q$3$ of the CV exam retains a residual ceiling.}
Fine-tuning lowers MAE on every graded CV-exam question, but a residual ceiling remains on Q$3$, semantic segmentation, the most open-ended: post-tuning MAE $1.1$--$1.3$ across models under the marks recipe, against $0.6$--$1.0$ on Q$1$/Q$2$.

\paragraph{The catastrophic tail disappears, on both exams.}
Under a neutral prompt the base models mis-grade up to $43$ of $114$ CV-exam students (Llama) and $66$ of $208$ ML-exam students by more than $10$ marks (${>}16$ on the ML exam's larger scale); pooled, every model is at $0$--$1$ (CV) and $2$--$4$ (ML) under marks ($0$--$1$ and $4$--$6$ under bd).

\paragraph{Errors become human-like, on both exams.}
We correlate each grader's per-student absolute error with the exam's intrinsic ambiguity, $|\mathrm{G}_1 - \mathrm{G}_2|$. On the CV exam the bases show no consistent relationship ($-0.31$ to $+0.37$) and four of five blunder on submissions the two graders \emph{agreed} on; after pooled fine-tuning the correlation is positive for every model on both exams ($+0.35$ to $+0.48$), the residual errors concentrating on the genuinely ambiguous submissions. Part of that correlation is arithmetic: the grader-average target carries the graders' own noise, so any low-bias grader's residual co-varies with $|\mathrm{G}_1 - \mathrm{G}_2|$ (the coupling Appendix~\ref{app:gt} notes for pairs) and we do not subtract it; the base-to-tuned change establishes the loss of blunders on submissions the graders agreed on, not human-like judgement.

\clearpage
\section{Per-Question Error Decomposition}
\label{app:perq}

Table~\ref{tab:perqbias} decomposes the mean signed error per question for the headline runs of both exams; both sides are sums of per-question marks, so the decomposition is exact, reproducing each run's published bias (Tables~\ref{tab:headline},~\ref{tab:allruns} and~\ref{tab:iaruns}) to within $4 \times 10^{-15}$.

Across the $44$ runs decomposed --- the ten in the table plus the $17$ matched neutral/strict open-weights pairs of Table~\ref{tab:brittle} --- the largest cancellation is L-A$14$ (Qwen2.5-Coder-$14$B, neutral): per-question biases of $+0.89$, $-1.30$ and $-0.75$ sum to $-1.16$, hiding $1.78$ of $2.94$ points of directional error ($61\%$); the largest in the table is A$01$, at $1.25$ of $3.03$. Cancellation is confined to the near-unbiased runs: all $17$ matched \emph{strict} runs and all four ML-exam Gemini runs have $C = 0$, so no persona-collapse claim rests on a cancelled total.

\begin{table}[t]
\centering
\caption{Per-question mean signed error for the headline runs of both exams, over exactly the students each run's published metric uses. $b_q$ is the mean of $\mathrm{AI}_q - \bar{\mathrm{G}}_q$ on question $q$ (CV exam: $35$-point base scale, Q$4$ bonus excluded; ML exam: score plus bonus, $\approx 65$-point scale). \emph{Total} is the run's published bias. $C = \sum_q |b_q| - |\sum_q b_q|$ is the cancellation the total hides, zero exactly when every question is pushed the same way. At two decimals a row can miss $\sum_q b_q = {}$\emph{Total} by $0.01$ through rounding. Generated by \texttt{analysis/checks/per\_question\_signed\_error.py}.}
\label{tab:perqbias}
\small
\setlength{\tabcolsep}{4.5pt}
\begin{tabular}{llrrrrrrr}
\toprule
Run & Configuration & $n$ & $b_1$ & $b_2$ & $b_3$ & Total & $\sum_q |b_q|$ & $C$ \\
\midrule
\multicolumn{9}{l}{\emph{CV exam --- $35$-point base scale (Q$1$ $12$, Q$2$ $11$, Q$3$ $12$)}} \\
D01 & 3-flash-preview, neutral        & 570 & $-0.15$ & $+0.08$ & $+0.08$ & $+0.01$ & 0.30 & 0.29 \\
D02 & 3.1-pro-preview, neutral        & 570 & $-0.44$ & $-0.07$ & $-0.31$ & $-0.82$ & 0.82 & 0.00 \\
F02 & 3.1-pro-preview $+$ lenient     & 570 & $-0.01$ & $+0.28$ & $+0.55$ & $+0.82$ & 0.84 & 0.02 \\
B03 & flash-lite $+$ thinking         & 570 & $-0.50$ & $+0.04$ & $+0.30$ & $-0.15$ & 0.85 & 0.69 \\
A01 & flash-lite, neutral (baseline)  & 570 & $-0.28$ & $-2.12$ & $+0.63$ & $-1.78$ & 3.03 & 1.25 \\
C01 & flash-lite $+$ strict           & 570 & $-1.80$ & $-3.08$ & $-0.55$ & $-5.44$ & 5.44 & 0.00 \\
\midrule
\multicolumn{9}{l}{\emph{ML exam --- $\approx 65$-point scale (Q$1$ $26$, Q$2$ $17$, Q$3$ $22$)}} \\
IG08 & 3.1-pro-preview, neutral       & 1038 & $+0.43$ & $+0.20$ & $+0.55$ & $+1.18$ & 1.18 & 0.00 \\
IG07 & 3-flash-preview, neutral       & 1038 & $+1.55$ & $+0.10$ & $+0.59$ & $+2.24$ & 2.24 & 0.00 \\
IG01 & flash-lite, neutral            & 1013 & $+4.00$ & $+0.79$ & $+1.19$ & $+5.98$ & 5.98 & 0.00 \\
IG05 & flash-lite $+$ strict          & 1008 & $-1.97$ & $-2.82$ & $-3.04$ & $-7.83$ & 7.83 & 0.00 \\
\bottomrule
\end{tabular}
\end{table}

\clearpage
\section{Ground-Truth Structure}
\label{sec:groundtruth}
\label{app:gt}

Two checks probe the grader-average target, with D$01$ as the AI side and, on the ML exam, its best full-cohort run (IG$08$, \texttt{gemini-3.1-pro-preview}).

\begin{figure}[t]
\centering
\includegraphics[width=0.78\linewidth]{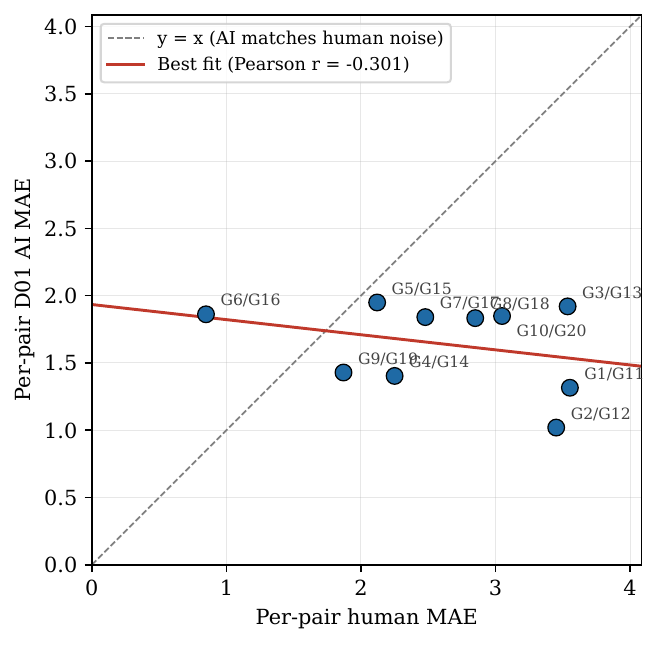}
\caption{Per-pair human disagreement (x-axis) against D$01$'s per-pair MAE (y-axis), one point per grader pair. The two are uncorrelated (Pearson $r = -0.30$, $p = 0.40$): the AI grader sits roughly equidistant from all pairs rather than inheriting pair-level noise.}
\label{fig:pairscatter}
\end{figure}

\paragraph{Whether the AI inherits per-pair human noise is exam-specific.}
\label{sec:groundtruth:perpair}
If the AI inherited its target's noise structure, its per-pair error should track the human MAE, which spans $0.85$ to $3.55$. It does not ($r = -0.301$, $p = 0.40$; Table~\ref{tab:perpair}, Figure~\ref{fig:pairscatter}). Agreement with any one grader's judgements is a different metric, not measurable here.

\emph{And it does not generalise.} On the ML exam's $18$ stable pairs the test inverts --- $r = +0.571$ (permutation $p = 0.014$), a difference itself significant (Fisher $z = 2.10$, $p = 0.036$), and the sign holds for every AI side we tried ($+0.57$ to $+0.72$). Some coupling must be arithmetic, since a noisier pair injects its variance into the target, but the correction is not clean enough to rest on; we report the divergence, not a mechanism; the ML exam's contiguous-block grading assignment is one candidate (Appendix~\ref{app:replication}).

\paragraph{Per-grader harshness is real, modest, and best measured within pairs.}
\label{sec:groundtruth:perta}
The $20$ graders span a $6.55$-point range in mean awarded total. A one-way ANOVA over the $1{,}140$ grades puts that at $\eta^2 = 0.051$ ($p = 5\times10^{-6}$), but those grades are not independent: each submission is marked by both members of one fixed pair, and each pair marks its own batch of $55$--$60$ students, confounding between-pair harshness with batch ability (Appendix~\ref{app:replication}). The $\eta^2$ splits orthogonally into $0.034$ between pairs and $0.017$ within pairs, the only part identified from the \emph{same} papers. A linear mixed model $\text{total} \sim \text{grader}$ with a random intercept per student (student variance $48.69$, residual $5.73$, $\mathrm{ICC} = 0.89$) gives the ten within-pair contrasts standard errors of $0.44$--$0.46$ against $1.36$--$1.40$ for the between-pair ones, and the likelihood-ratio tests separate them: collapsing graders to pair means costs little ($\chi^2(9) = 21.1$, $p = 0.012$), dropping the within-pair contrasts costs a great deal ($\chi^2(10) = 168.9$, $p = 5\times10^{-31}$; blocked-ANOVA partial $\eta^2 = 0.26$ of the paired-difference variance). Within pairs the mean signed difference runs $-2.75$ to $+3.11$ points, six of ten pairs are non-zero after Holm correction, and a sign-flip omnibus over the $570$ paired differences gives $p < 10^{-4}$. A student's expected score does depend on which pair graded them, and the AI's MAE incorporates that noise in its target; but only the within-pair third of the $5\%$ is measurable as harshness rather than as batch ability. That $\eta^2$ is design-dependent and does not transfer; the quantity that does --- signed bias between two graders on the \emph{same} papers --- agrees across exams at $0.61\times$ the floor (Appendix~\ref{app:replication}).

Together the two results say the best closed-model grader acts as a $\text{MAE} \approx 1.6$ stabiliser around a noisy human consensus on the CV exam, sitting below all but the tightest pair's disagreement bar.

The largest single-paper disagreement is $24.1$ points. The one-way tests use all $1{,}140$ grades ($570$ students $\times$ two graders), and the non-parametric Kruskal--Wallis test agrees with the ANOVA ($H = 53.80$, $p = 3.5\times10^{-5}$); both share the independence assumption that the paired analysis above replaces. Neither that analysis nor the one-way tests turn on the one negative total (a $\mathrm{TA}_1$ total of $-1.8$): removing it gives $H = 54.34$, a within-pair $\eta^2$ share of $0.017$, and $\chi^2(10) = 171.6$.

\begin{table}[h]
\centering
\caption{Per-pair human disagreement vs.\ D$01$'s per-pair AI MAE, sorted by human MAE.}
\label{tab:perpair}
\small
\begin{tabular}{lrrrr}
\toprule
Pair & $n$ & Human MAE & D$01$ AI MAE & D$01$ AI bias \\
\midrule
G${}_{6}$ / G${}_{16}$  & 56 & 0.85 & 1.86 & $+1.18$ \\
G${}_{9}$ / G${}_{19}$  & 55 & 1.87 & 1.43 & $-0.46$ \\
G${}_{5}$ / G${}_{15}$  & 57 & 2.12 & 1.95 & $-0.90$ \\
G${}_{4}$ / G${}_{14}$  & 56 & 2.25 & 1.40 & $-0.88$ \\
G${}_{7}$ / G${}_{17}$  & 57 & 2.48 & 1.84 & $+1.11$ \\
G${}_{8}$ / G${}_{18}$  & 58 & 2.85 & 1.83 & $-0.34$ \\
G${}_{10}$ / G${}_{20}$ & 60 & 3.05 & 1.85 & $-0.91$ \\
G${}_{2}$ / G${}_{12}$  & 58 & 3.45 & 1.02 & $+0.12$ \\
G${}_{3}$ / G${}_{13}$  & 56 & 3.54 & 1.92 & $+1.63$ \\
G${}_{1}$ / G${}_{11}$  & 57 & 3.55 & 1.32 & $-0.36$ \\
\bottomrule
\end{tabular}
\end{table}

\clearpage
\section{ML-Exam Replication Details}
\label{app:replication}

The ML exam of Section~\ref{sec:brittle:replication} is an introductory AI practical from another course at the same institution: $1{,}038$ students after excluding $24$ whose uploads contained no gradable notebook, dual-graded by $49$ graders, three questions (regression, PyTorch, classification) worth $23{+}3$, $14{+}3$ and $19{+}3$ marks including bonuses ($56$ base points plus up to $9$ bonus points, $65$ in all). Question maxima come from the marking scheme's task tables, not its header totals, which contradict them in all three questions; two instructor corrections issued by email mid-grading are folded in, both affecting Q$3$. Per-question grader marks fold bonus into the score, so the comparable AI total is score plus bonus. A student who did not submit a question is scored $0$ by graders and models alike.

The grading instrument is identical to Section~\ref{sec:setup} (same persona strings, JSON schema, output rules and grading scale, reference solution, rubric breakdown, no chain-of-thought, temperature $0$, one sample per question), except that this exam has no student-facing guidelines document, so the \texttt{gd} component is fixed at $0$ throughout.

\paragraph{The grid.}
The replication comprises $162$ configurations --- $30$ closed-model ($23$ Gemini IG-, $6$ OpenAI IO-, $1$ Anthropic IN-) and $132$ open-weights (IA-) --- replaying the CV exam's programme: neutral and matched \emph{strict} runs for all $17$ open-weights models, five-persona sweeps on $13$, the mechanism probes of Section~\ref{sec:brittle:attrib}, component removals, few-shot demonstrations on one closed and one open model, temperature-variance probes, and a closed arm covering all four Gemini models, \texttt{gpt-5.5}, \texttt{gpt-5.4} and \texttt{claude-opus-5}. Table~\ref{tab:iaruns} (Appendix~\ref{app:iaruns}) lists every run.

\begin{table}[h]
\centering
\caption{ML exam, $1{,}038$ dual-graded students: matched neutral/strict pairs for all $17$ open-weights models, sorted by damage ratio. MAE against the grader average on a $\approx 65$-point scale, $95\%$ bootstrap CI on the strict run, strict MAE as a multiple of the ML exam's human floor ($5.13$, defined as in Section~\ref{sec:floor}). Behaviour classes are those of Section~\ref{sec:brittle:numbers} and describe the strict run, at this exam's floor-matched band $\text{MAE} \ge 15.7$ ($3.07\times$ its $5.13$ floor, the multiple the CV exam's $8$ fixes); four strict runs and one neutral baseline cross it. A fixed $\text{MAE} \ge 8$ would be only $1.56\times$ this floor and would label ten strict runs and eleven neutral baselines collapse. A refusal's MAE is a ceiling artifact, not a severity measurement. Ratios below $1$ are real improvements, examined below. Generated by \texttt{analysis/introduction\_to\_ai\_make\_paper\_tables.py}.}
\label{tab:replication}
\small
\resizebox{\linewidth}{!}{%
\begin{tabular}{lrrlrrrrl}
\toprule
model & neutral & strict & strict CI & ratio & $\times$floor & zero-rate & bias & behaviour \\
\midrule
GLM-4.5-Air & 5.60 & 18.49 & [17.96, 19.02] & $\times 3.30$ & 3.6 & 37.8\% & $-18.31$ & collapse \\
Qwen2.5-Coder-32B & 4.57 & 13.24 & [12.79, 13.69] & $\times 2.90$ & 2.6 & 10.5\% & $-13.07$ & graded \\
Llama-3.1-8B & 13.16 & 29.58 & [28.70, 30.45] & $\times 2.25$ & 5.8 & 100.0\% & $-29.58$ & \textbf{refusal} \\
Llama-3.3-70B & 4.70 & 10.17 & [9.77, 10.58] & $\times 2.16$ & 2.0 & 9.0\% & $-9.63$ & graded \\
Mistral-Small-24B & 8.96 & 17.32 & [16.77, 17.87] & $\times 1.93$ & 3.4 & 32.2\% & $-17.20$ & collapse \\
Qwen3-Coder-Next 80B & 4.47 & 8.63 & [8.26, 9.00] & $\times 1.93$ & 1.7 & 2.3\% & $-8.12$ & graded \\
GLM-4-32B & 8.03 & 12.18 & [11.68, 12.69] & $\times 1.52$ & 2.4 & 14.1\% & $-10.68$ & graded \\
Qwen2.5-Coder-14B & 8.61 & 11.06 & [10.61, 11.50] & $\times 1.28$ & 2.2 & 16.6\% & $-10.46$ & graded \\
GLM-4-9B & 13.15 & 16.68 & [16.15, 17.20] & $\times 1.27$ & 3.2 & 35.7\% & $-16.52$ & collapse \\
Qwen2.5-Coder-7B & 6.88 & 7.45 & [7.11, 7.80] & $\times 1.08$ & 1.5 & 7.4\% & $-5.67$ & graded \\
Qwen3-235B-A22B & 5.40 & 5.18 & [4.89, 5.47] & $\times 0.96$ & 1.0 & 1.7\% & $-3.39$ & graded \\
Gemma-3-27B & 8.33 & 7.61 & [7.23, 7.99] & $\times 0.91$ & 1.5 & 0.6\% & $-6.25$ & graded \\
Gemma-3-12B & 10.36 & 6.79 & [6.49, 7.10] & $\times 0.66$ & 1.3 & 0.5\% & $+0.09$ & graded \\
Qwen3-Coder-480B & 10.10 & 6.04 & [5.75, 6.34] & $\times 0.60$ & 1.2 & 1.0\% & $+3.25$ & graded \\
DeepSeek-Coder-V2-Lite & 16.41 & 9.22 & [8.83, 9.62] & $\times 0.56$ & 1.8 & 0.0\% & $+2.42$ & graded \\
Qwen2.5-72B & 10.64 & 5.42 & [5.15, 5.70] & $\times 0.51$ & 1.1 & 2.3\% & $-1.67$ & graded \\
Qwen3-Coder-30B-A3B & 11.12 & 5.58 & [5.31, 5.87] & $\times 0.50$ & 1.1 & 2.2\% & $-1.55$ & graded \\
\bottomrule
\end{tabular}}
\end{table}

\begin{figure}[t]
\centering
\includegraphics[width=\linewidth]{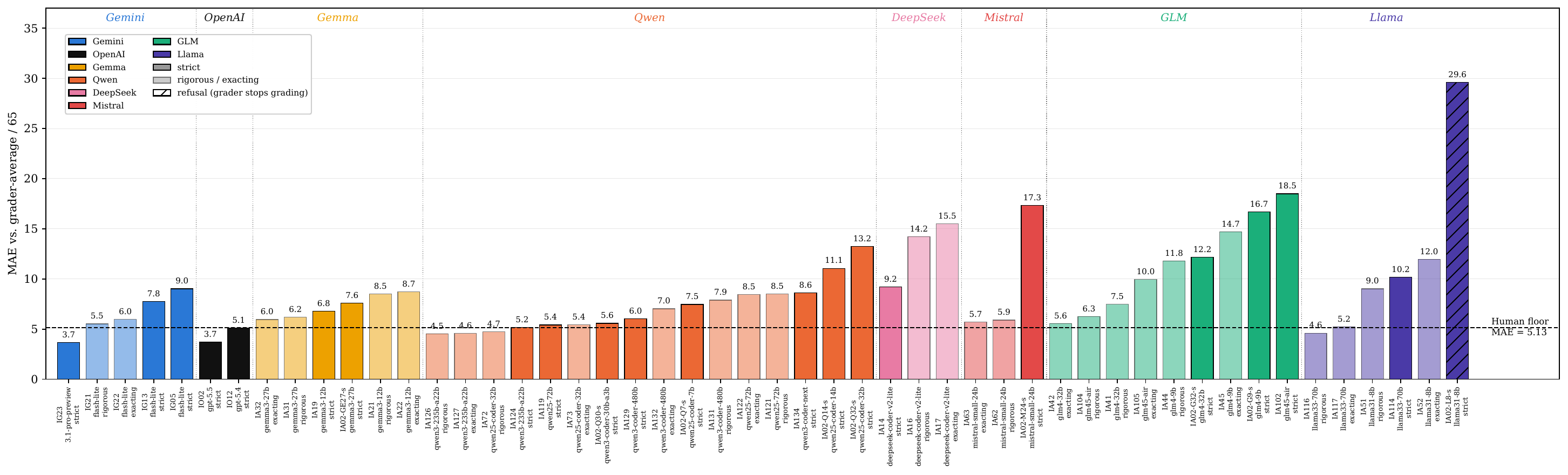}
\caption{Every strict-flavoured persona run on the ML exam's full cohort ($50$ runs: \emph{strict}/\emph{rigorous}/\emph{exacting} at the default prompt configuration, including IG$13$, which additionally enables thinking), coloured by family and by wording (\emph{strict} darker; \emph{rigorous}/\emph{exacting} lighter). The dashed line is the ML exam's human inter-grader floor ($\text{MAE} = 5.13$). The refusal (Llama-3.1-8B $+$ \emph{strict}) is hatched: its bar height is the distance to the grader mean, not a severity measurement. The ML exam's counterpart to Figure~\ref{fig:brittle}.}
\label{fig:iabrittle}
\end{figure}

\paragraph{When severity helps, it is calibration, not robustness.}
Fifteen of the $17$ neutral baselines over-mark, with biases from $+1.51$ to $+14.89$ (the exceptions are Qwen2.5-Coder-$32$B at $-0.21$ and GLM-4.5-Air at $-1.34$), and across the $16$ non-refusal pairs a model's neutral bias predicts the strict effect at $r = -0.73$ (Spearman $-0.73$, permutation $p = 0.002$; $r = -0.50$ with the refusal's ceiling-artifact MAE included), whereas the CV exam's neutral biases sit at or below zero for all but two models and zero of $17$ improve. The seven improvements in Table~\ref{tab:replication} are heavy over-markers pushed down --- Qwen3-Coder-$30$B-A$3$B (bias $+10.44$), Qwen2.5-$72$B ($+10.31$), Qwen3-Coder-$480$B ($+9.52$) --- two miscalibrations cancelling: the $480$B is one such case ($\text{MAE} = 10.10$ at neutral, twice its floor), whereas Qwen3-$235$B-A$22$B is genuinely indifferent to the sentence on both exams (here bias $+3.73$, strict effect $-0.22$).

\paragraph{Mechanism.}
The probe of Section~\ref{sec:brittle:mech} separates the same two shapes on the ML exam (Table~\ref{tab:repmech}). It is read only for runs that left the graded band and zeroed at least $5\%$ of Q$1$, the volume gate the CV exam's classifier applies at $10\%$; two runs fall below it, DeepSeek-Coder-V2-Lite ($6$ of $1038$, $0.6\%$) and Qwen3-Coder-Next $80$B ($39$ of $1001$, $3.9\%$).

\begin{table}[h]
\centering
\caption{ML exam, \emph{strict} runs: blanket versus selective zeroing, all $17$ models. Q$1 = 0$ is the count of students awarded zero on Q$1$ out of those the run graded; the next column is how many of \emph{those} still received a non-zero Q$2$. A run is read only if it left the graded band \emph{and} zeroed Q$1$ on at least $5\%$ of the students it graded (the CV exam applies the same volume gate at $10\%$, \texttt{MECH\_ZERO\_RATE}; on these runs any cut between $4.8\%$ and $13.3\%$ gives the same partition): below that, the zeros are ordinary non-attempts and the share is noise on a handful of students. Generated by \texttt{analysis/introduction\_to\_ai\_make\_paper\_tables.py}.}
\label{tab:repmech}
\small
\begin{tabular}{lrrrl}
\toprule
model (strict) & Q$1 = 0$ & of those, Q$2 > 0$ & share & reading \\
\midrule
Llama-3.1-8B & 1038 / 1038 (100\%) & 0 & 0\% & blanket \\
GLM-4-9B & 669 / 1038 (64\%) & 255 & 38\% & selective \\
GLM-4.5-Air & 521 / 1038 (50\%) & 110 & 21\% & blanket \\
GLM-4-32B & 477 / 1038 (46\%) & 295 & 62\% & n/a (in band) \\
Mistral-Small-24B & 375 / 1038 (36\%) & 21 & 6\% & blanket \\
Qwen2.5-Coder-14B & 196 / 1038 (19\%) & 13 & 7\% & n/a (in band) \\
Llama-3.3-70B & 152 / 1038 (15\%) & 40 & 26\% & n/a (in band) \\
Qwen2.5-Coder-32B & 144 / 1038 (14\%) & 19 & 13\% & n/a (in band) \\
Qwen2.5-Coder-7B & 138 / 1038 (13\%) & 59 & 43\% & n/a (in band) \\
Qwen3-Coder-30B-A3B & 50 / 1038 (5\%) & 24 & 48\% & n/a (in band) \\
Qwen2.5-72B & 48 / 1038 (5\%) & 21 & 44\% & n/a (in band) \\
Qwen3-Coder-Next 80B & 39 / 1001 (4\%) & 13 & 33\% & n/a (in band) \\
Qwen3-235B-A22B & 30 / 1038 (3\%) & 10 & 33\% & n/a (in band) \\
Gemma-3-12B & 21 / 1038 (2\%) & 14 & 67\% & n/a (in band) \\
Qwen3-Coder-480B & 19 / 1038 (2\%) & 7 & 37\% & n/a (in band) \\
Gemma-3-27B & 15 / 1038 (1\%) & 7 & 47\% & n/a (in band) \\
DeepSeek-Coder-V2-Lite & 6 / 1038 (1\%) & 4 & 67\% & n/a (in band) \\
\bottomrule
\end{tabular}
\end{table}

\paragraph{The attribution series replicates.}
The minimal pairs of Section~\ref{sec:brittle:attrib} reproduce both findings. S2 remains the only single component that stops graders grading: alone it takes Llama-3.1-$8$B to $29.39$ ($98\%$ of students zeroed) and Mistral-Small-$24$B to $26.03$ ($76\%$ zeroed), while the HARSH frame alone moves each of the three models probed by at most $2.4$ points from neutral, twice in the \emph{improving} direction. Nor does the adjective carry the damage: with both policy sentences held verbatim, swapping STRICT for RIGOROUS or FAIR moves MAE by $0.55$--$3.86$ across the five models with no consistent direction --- a FAIR teaching assistant carrying the harsh policy still more than doubles Qwen2.5-Coder-$32$B's error ($10.40$ against $4.57$) --- while the \emph{nharsh} control (a harsh adjective with no policy) lands within $1.2$ points of neutral on all five. Absolute damage is exam-dependent --- S2 alone leaves Qwen2.5-Coder-$32$B in the graded band here ($6.68$, against $12.92$ on the CV exam) --- but which sentence dominates a model is not: S1 outweighs S2 for the $32$B and S2 for the other two, on both exams.

\begin{figure}[t]
\centering
\includegraphics[width=0.66\linewidth]{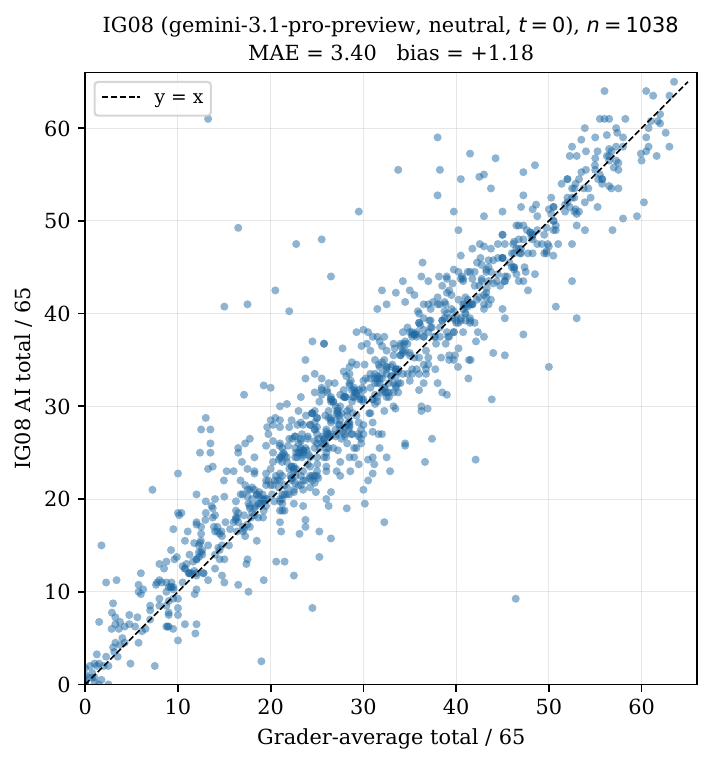}
\caption{IG$08$ (\texttt{gemini-3.1-pro-preview}, neutral, $t=0$): AI total vs.\ grader-average total over the ML exam's full $n = 1{,}038$ cohort, with $y = x$ as the perfect-agreement reference. The best full-cohort configuration on the ML exam ($0.66\times$ its human floor), the counterpart to Figure~\ref{fig:d01scatter}.}
\label{fig:iascatter}
\end{figure}

\paragraph{The closed-model arm.}
The ML exam also carries all four Gemini models under the same instrument. The parity result replicates on the closed side --- all three full-cohort neutral configurations sit wholly below the floor's CI: \texttt{gemini-3.1-pro-preview} at $3.40$ $[3.16, 3.65]$ (IG$08$; Figure~\ref{fig:iascatter}), \texttt{gemini-2.5-pro} at $3.47$, and \texttt{gemini-3-flash-preview} at $3.92$, against a floor of $5.13$ $[4.80, 5.48]$; the $3.1$-pro under \emph{lenient} ($4.48$ $[4.22, 4.75]$) and under \emph{strict} ($3.67$ $[3.43, 3.94]$, bias $-1.00$ against $+1.18$ at neutral; IG$23$) join them. The cross-vendor runs of Section~\ref{sec:closed:vendors} agree: no other closed run leaves this exam's band (Table~\ref{tab:closedvendors}). What does \emph{not} replicate is Gemini's persona immunity at the cheaper tier. Flash-Lite, at $7.53$ under neutral (bias $+5.98$), moves in \emph{both} directions: \emph{strict} takes it to $9.02$ (bias $-7.83$) and \emph{lenient} to $17.86$ (bias $+17.80$), the one closed run past this exam's band, which only four open-weights strict runs cross --- while \emph{rigorous} and \emph{exacting} improve it ($5.52$, $5.96$), the same calibration correction the open side shows. The stronger \texttt{gemini-3-flash-preview} follows the same law on its first-$100$ subset: against a matched-subset baseline of $3.87$ (bias $+3.04$), \emph{strict} lands at $2.87$ (bias $+0.09$) and \emph{lenient} at $6.99$ (bias $+6.60$). Component ablations echo the CV exam with larger magnitudes: dropping the reference solution costs Flash-Lite $4.93$ points ($7.53 \to 12.46$) against $0.79$ there, and thinking helps by $2.16$ ($7.53 \to 5.37$) against $1.29$.

\paragraph{Few-shot demonstrations transfer, with one measurement caveat.}
Two worked examples per question (from the two highest-agreement grader pairs, labelled by IG$08$; the three students remain in the evaluated cohort, $3/1{,}038$) nearly halve the open model's error: Qwen3-Coder-$30$B-A$3$B goes from $11.12$ to $5.84$, bias $+10.44$ to $+4.56$, on the full cohort (IA$201$). On Flash-Lite (IG$18$) they lengthen the prompt enough that $334$ students' calls fail permanently, with biased dropout: the dropped students average $32.78$ points against the survivors' $28.02$ (Welch $p = 3\times10^{-7}$). On the matched $679$-student subsample the effect holds ($7.87 \to 4.32$ MAE, bias $+6.52 \to +0.37$), but its headline numbers are not comparable to full-cohort rows and Table~\ref{tab:iaruns} flags its $n$.

\paragraph{Determinism at $t = 0$ does not transfer.}
On the CV exam Flash-Lite reproduced $100\%$ of (student, question) cells exactly across five $t = 0$ reruns (Appendix~\ref{sec:closed:variance}). Here the same probe (IG$10$) reproduces only $59.3\%$; the median per-cell standard deviation is still $0.00$ but a $40.7\%$ tail varies, and at $t = 0.5$/$0.7$ the medians rise to $0.98$/$1.14$ (IG$11$/IG$12$) against $0.65$/$0.76$ there. Two open-weights probes on $41$ students at $t = 0.5$ (IA$92$, Qwen2.5-Coder-$32$B; IA$136$, Qwen3-Coder-Next) show medians of $0.65$ and $0.84$.

\paragraph{Component removals do not transfer either.}
The CV exam's over-anchoring result ($7.49 \to 3.12$ on Qwen2.5-Coder-$32$B with the reference solution removed) does not reproduce: the same removal moves it $4.57 \to 4.73$ here. The rubric-breakdown test of Appendix~\ref{app:brittle} softens: under \emph{strict}, removing the breakdown worsens the collapse on GLM-4-$32$B ($+6.26$) and Llama-3.3-$70$B ($+3.96$) but is flat on Qwen2.5-Coder-$32$B ($+0.19$), Gemma-3-$27$B ($-0.26$) and DeepSeek-Coder-V2-Lite ($-0.10$), neutral controls within $\pm 0.52$ --- against five-of-five worsening there.

\paragraph{Ground-truth structure of the ML exam.}
Its floor is $5.13 / 65$ ($95\%$ bootstrap CI $[4.80, 5.48]$ at \num{50000} resamples --- the lower bound is not stable to two decimals at \num{2000}; Pearson $r = 0.871$). Nine of the $1{,}038$ rows record $0.0$ for exactly one grader while the partner awarded real marks: ungraded slots stored as zeros. They move the floor by $0.05$ (inside the CI, so the headline retains them) but dominate the extremes, the maximum single-paper disagreement being $41.0$ without them and $51.0$ with. Excluding them everywhere changes no conclusion: the floor falls to $5.09$ $[4.76, 5.42]$, the best full-cohort configurations move by at most $0.06$ MAE (IG$08$ $3.40 \to 3.35$, IG$09$ $3.47 \to 3.42$, IG$07$ $3.92 \to 3.86$, Qwen3-Coder-Next neutral $4.47 \to 4.44$), no run of Table~\ref{tab:replication} or Table~\ref{tab:iaruns} changes behaviour class or damage-ratio order (largest move $0.21$ MAE, the Llama-3.1-$8$B refusal), and on the $208$ held-out students (three of the nine; the other six sit in the $830$-student fine-tuning split) the pooled adapters grade $0.08$--$0.11$ MAE better, every Table~\ref{tab:ftpaired} verdict intact (\texttt{analysis/checks/ml\_unrecorded\_grader\_rows.py}). A further nine rows carry $0.0$ from both graders and are genuine non-submissions, retained.

The pairing is hybrid, unlike the CV exam's $10$ fixed pairs: $49$ assistants work in $48$ grading slots (one slot is a pair who marked jointly) forming $51$ pairings, of which $18$ are stable pairs of $38$--$51$ students covering $816$ of $1{,}038$ ($79\%$); the rest are ad hoc combinations around floating graders, and two of the $51$ on the roster returned no marks. Every per-pair figure is scoped to that backbone, across which human MAE runs $1.90$ to $10.36$ --- a $5.4\times$ spread, against $4.2\times$ on the CV exam --- with eight of $18$ pairs above the pooled floor, against five of ten. Mean within-pair bias magnitude is $3.14$ ($0.61\times$ floor) against $1.58$ ($0.61\times$ floor): systematic harshness is the same fraction of the noise floor on both exams, whereas a one-way $\eta^2$ over grader identity is not comparable here, because each pair marks a contiguous block of students of widely varying ability.

\begin{figure}[t]
\centering
\includegraphics[width=0.78\linewidth]{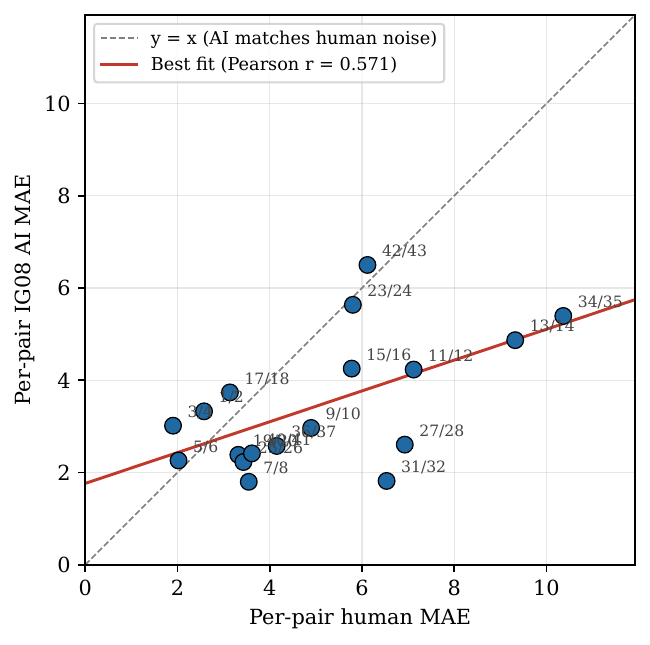}
\caption{ML exam: per-pair human disagreement (x-axis) against IG$08$'s per-pair MAE (y-axis) over the $18$-pair stable backbone. Unlike the CV exam (Figure~\ref{fig:pairscatter}), the two correlate ($r = +0.571$): here the AI's error is largest exactly where the humans disagree most.}
\label{fig:iapairscatter}
\end{figure}

The per-pair test of Appendix~\ref{sec:groundtruth:perpair} inverts here: per-pair human MAE against IG$08$'s per-pair MAE gives Pearson $r = +0.571$ (permutation $p = 0.014$, \num{20000} shuffles; Spearman $+0.447$; Figure~\ref{fig:iapairscatter}) against $-0.301$ on the CV exam, a difference significant at Fisher $z = 2.10$, $p = 0.036$, and the sign survives changing the AI side (Qwen3-Coder-Next neutral $+0.565$, Qwen2.5-Coder-$32$B $+0.723$). Some of the coupling is arithmetic, a noisier pair injecting its variance into the target. One candidate for the rest is the assignment itself: ML pairs mark contiguous blocks of students whose ability varies widely, so block difficulty raises both human disagreement and model error within a pair, whereas the CV exam's fixed pairs each drew a comparable slice. The AI is still the more uniform grader, its per-pair MAE spanning $3.6\times$.

\paragraph{What the replication does and does not establish.}
\emph{Strict} worsens ten of $17$ models and takes four above this exam's floor-matched band ($\text{MAE} \ge 15.7$), none of them there at neutral. What transfers is the floor-relative attainability of parity (IG$08$ at $0.66\times$ this exam's floor; the best open-weights baseline, Qwen3-Coder-Next, $4.47$ $[4.24, 4.72]$, also wholly below its CI), the S2 mechanism, the blanket/selective split and the danger of an untested persona sentence; model choice, persona direction, component effects, failure shape (Qwen2.5-Coder-$32$B inverts from selective field collapse to blanket zeroing) and $t = 0$ determinism are exam-level properties.

\section{All Grader Configurations}
\label{app:runs}

Table~\ref{tab:allruns} lists every one of the $171$ grader configurations with its headline metrics, one row per run. The machine-readable version, with per-question breakdowns and the full column set, ships as \texttt{analysis/computer\_vision\_master\_comparison.csv} \artifactsin.

\begingroup
\scriptsize
\setlength{\tabcolsep}{2.5pt}
\setlength{\LTcapwidth}{\linewidth}
\begin{longtable}{llllcrrcrl}
\caption{Every grader configuration in the study, one row per run: the $33$ closed-model configurations ($25$ Gemini via Vertex, $6$ OpenAI and $2$ Anthropic via their batch APIs; Section~\ref{sec:closed:vendors}) followed by the $138$ open-weights (L-) configurations, each sorted by run id. \emph{Prompt} lists the components present (S~reference solution, G~grading guidelines, B~rubric breakdown, R~thinking enabled, F~few-shot demonstrations); personas prefixed \texttt{mp} are the mechanism probes of Section~\ref{sec:brittle:attrib}: \texttt{mpstrict}, \texttt{mprigorous} and \texttt{mpfair} keep the \emph{strict} preset's two policy sentences and swap only its headword; \texttt{mpframe} is the HARSH frame alone, \texttt{mpnoclause} the frame with S1, \texttt{mps2only} the frame with S2, \texttt{mpharsh} the frame with both (the \emph{strict} text, rerun on the current stack); \texttt{mpnharsh} is the neutral frame with ``harsh'' in place of ``strict''. L-MP$04$ and L-PS$03$ are one physical run (\texttt{qwen2.5-coder-32b}, \emph{mpnoclause}), listed under both the headword series and the policy $2\times2$. $n$ is the number of students scored against the grader average; MAE and bias are on the $35$-point base scale, the $95\%$ bootstrap CI on MAE; behaviour classes are defined in Section~\ref{sec:brittle:numbers}. A refusal's MAE is a ceiling artifact, not a severity measurement. D$03$ ran with thinking active regardless of its flag (\texttt{gemini-2.5-pro} cannot disable it; Appendix~\ref{app:details}). \texttt{claude-opus-5} exposes no sampling temperature (thinking disabled; shown as ---). The final block lists the $8$ prompt-only baselines of Appendix~\ref{app:promptbaselines} (first $100$ students; the persona column names the strategy), which are not configurations of the paper's grader and are excluded from every count; their CIs use $2000$ resamples. Model keys match the released spreadsheets. Generated by \texttt{analysis/computer\_vision\_make\_appendix\_table.py}.}\label{tab:allruns}\\
\toprule
Run & Model & Persona & Prompt & $t$ & $n$ & MAE & $95\%$ CI & Bias & Behaviour \\
\midrule
\endfirsthead
\multicolumn{10}{l}{\emph{Table~\ref{tab:allruns} continued}}\\
\toprule
Run & Model & Persona & Prompt & $t$ & $n$ & MAE & $95\%$ CI & Bias & Behaviour \\
\midrule
\endhead
\bottomrule
\endfoot
\multicolumn{10}{l}{\emph{Closed-model configurations (Gemini: A--P series; OpenAI: O; Anthropic: N)}}\\*
A01 & \texttt{flash-lite} & neutral & SGB & $0$ & $570$ & $3.34$ & $[3.15,\ 3.53]$ & $-1.78$ & graded \\
B01 & \texttt{flash-lite} & neutral & GB & $0$ & $567$ & $4.13$ & $[3.83,\ 4.42]$ & $+3.30$ & graded \\
B02 & \texttt{flash-lite} & neutral & SB & $0$ & $569$ & $3.34$ & $[3.16,\ 3.52]$ & $-1.42$ & graded \\
B03 & \texttt{flash-lite} & neutral & SGBR & $0$ & $570$ & $2.05$ & $[1.89,\ 2.21]$ & $-0.15$ & graded \\
B04 & \texttt{flash-lite} & neutral & SG & $0$ & $569$ & $3.22$ & $[3.03,\ 3.43]$ & $-1.45$ & graded \\
C01 & \texttt{flash-lite} & strict & SGB & $0$ & $570$ & $5.75$ & $[5.46,\ 6.04]$ & $-5.44$ & graded \\
C02 & \texttt{flash-lite} & lenient & SGB & $0$ & $569$ & $4.49$ & $[4.18,\ 4.82]$ & $+4.01$ & graded \\
D01 & \texttt{3-flash-preview} & neutral & SGB & $0$ & $570$ & $1.64$ & $[1.51,\ 1.79]$ & $+0.01$ & graded \\
D02 & \texttt{3.1-pro-preview} & neutral & SGB & $0$ & $570$ & $1.86$ & $[1.70,\ 2.02]$ & $-0.82$ & graded \\
D03 & \texttt{2.5-pro} & neutral & SGB & $0$ & $570$ & $4.10$ & $[3.83,\ 4.38]$ & $-3.85$ & graded \\
E01 & \texttt{flash-lite} & neutral & SGB & $0$ & $549$ & $3.38$ & $[3.19,\ 3.59]$ & $-1.85$ & graded \\
E02 & \texttt{flash-lite} & neutral & SGB & $0.5$ & $563$ & $3.51$ & $[3.31,\ 3.71]$ & $-2.20$ & graded \\
E03 & \texttt{flash-lite} & neutral & SGB & $0.7$ & $563$ & $3.60$ & $[3.40,\ 3.79]$ & $-2.32$ & graded \\
F01 & \texttt{flash-lite} & strict & SGBR & $0$ & $570$ & $3.73$ & $[3.50,\ 3.98]$ & $-3.28$ & graded \\
F02 & \texttt{3.1-pro-preview} & lenient & SGB & $0$ & $570$ & $1.79$ & $[1.62,\ 1.95]$ & $+0.82$ & graded \\
F03 & \texttt{3.1-pro-preview} & strict & SGB & $0$ & $570$ & $2.75$ & $[2.51,\ 2.99]$ & $-2.09$ & graded \\
G01 & \texttt{3-flash-preview} & neutral & GB & $0$ & $98$ & $1.63$ & $[1.32,\ 2.02]$ & $+0.71$ & graded \\
G02 & \texttt{3-flash-preview} & neutral & SB & $0$ & $99$ & $1.60$ & $[1.30,\ 1.93]$ & $+0.79$ & graded \\
G03 & \texttt{3-flash-preview} & neutral & SGBR & $0$ & $100$ & $1.39$ & $[1.18,\ 1.61]$ & $-0.12$ & graded \\
G04 & \texttt{3-flash-preview} & neutral & SG & $0$ & $100$ & $1.25$ & $[1.01,\ 1.51]$ & $-0.07$ & graded \\
K01 & \texttt{flash-lite} & neutral & SGBF & $0$ & $567$ & $3.03$ & $[2.83,\ 3.24]$ & $+1.09$ & graded \\
M01 & \texttt{3-flash-preview} & strict & SGB & $0$ & $99$ & $4.23$ & $[3.60,\ 4.88]$ & $-4.16$ & graded \\
M02 & \texttt{3-flash-preview} & lenient & SGB & $0$ & $100$ & $2.18$ & $[1.75,\ 2.65]$ & $+1.90$ & graded \\
N01 & \texttt{claude-opus-5} & neutral & SGB & --- & $570$ & $3.54$ & $[3.31,\ 3.77]$ & $-3.24$ & graded \\
N02 & \texttt{claude-opus-5} & strict & SGB & --- & $570$ & $5.17$ & $[4.90,\ 5.45]$ & $-5.02$ & graded \\
O01 & \texttt{gpt-5.5} & neutral & SGB & $0$ & $570$ & $2.43$ & $[2.28,\ 2.59]$ & $-1.65$ & graded \\
O02 & \texttt{gpt-5.5} & strict & SGB & $0$ & $570$ & $4.43$ & $[4.19,\ 4.68]$ & $-4.25$ & graded \\
O03 & \texttt{gpt-5.5} & lenient & SGB & $0$ & $570$ & $2.25$ & $[2.08,\ 2.44]$ & $+1.13$ & graded \\
O11 & \texttt{gpt-5.4} & neutral & SGB & $0$ & $570$ & $4.69$ & $[4.46,\ 4.92]$ & $-4.45$ & graded \\
O12 & \texttt{gpt-5.4} & strict & SGB & $0$ & $570$ & $6.90$ & $[6.63,\ 7.15]$ & $-6.81$ & graded \\
O13 & \texttt{gpt-5.4} & lenient & SGB & $0$ & $570$ & $2.69$ & $[2.51,\ 2.89]$ & $+1.00$ & graded \\
P01 & \texttt{flash-lite} & rigorous & SGB & $0$ & $570$ & $3.92$ & $[3.70,\ 4.16]$ & $-3.01$ & graded \\
P02 & \texttt{flash-lite} & exacting & SGB & $0$ & $557$ & $4.58$ & $[4.35,\ 4.85]$ & $-4.05$ & graded \\
\midrule
\multicolumn{10}{l}{\emph{Open-weights (L-) configurations}}\\*
L-A07 & \texttt{qwen2.5-coder-7b} & neutral & SGB & $0$ & $570$ & $5.68$ & $[5.39,\ 5.98]$ & $-4.84$ & graded \\
L-A14 & \texttt{qwen2.5-coder-14b} & neutral & SGB & $0$ & $570$ & $3.51$ & $[3.30,\ 3.74]$ & $-1.16$ & graded \\
L-A30M & \texttt{qwen3-coder-30b-a3b} & neutral & SGB & $0$ & $570$ & $4.50$ & $[4.29,\ 4.72]$ & $-2.70$ & graded \\
L-A32 & \texttt{qwen2.5-coder-32b} & neutral & SGB & $0$ & $570$ & $7.49$ & $[7.19,\ 7.78]$ & $-7.35$ & graded \\
L-ANxt & \texttt{qwen3-coder-next} & neutral & SGB & $0$ & $570$ & $5.17$ & $[4.92,\ 5.42]$ & $-4.74$ & graded \\
L-B01 & \texttt{qwen2.5-coder-32b} & neutral & GB & $0$ & $570$ & $3.12$ & $[2.92,\ 3.31]$ & $-1.56$ & graded \\
L-B02 & \texttt{qwen2.5-coder-32b} & neutral & SB & $0$ & $570$ & $5.86$ & $[5.60,\ 6.11]$ & $-5.46$ & graded \\
L-B03 & \texttt{qwen2.5-coder-32b} & neutral & SGBR & $0$ & $570$ & $7.51$ & $[7.22,\ 7.80]$ & $-7.36$ & graded \\
L-B04 & \texttt{qwen2.5-coder-32b} & neutral & SG & $0$ & $570$ & $7.27$ & $[6.98,\ 7.55]$ & $-7.11$ & graded \\
L-BD01 & \texttt{qwen2.5-coder-32b} & neutral & SG & $0$ & $570$ & $7.36$ & $[7.06,\ 7.64]$ & $-7.23$ & graded \\
L-BD02 & \texttt{qwen2.5-coder-32b} & strict & SG & $0$ & $570$ & $21.69$ & $[21.16,\ 22.23]$ & $-21.68$ & collapse \\
L-BD03 & \texttt{llama-3.3-70b} & neutral & SG & $0$ & $570$ & $4.69$ & $[4.46,\ 4.92]$ & $-3.96$ & graded \\
L-BD04 & \texttt{llama-3.3-70b} & strict & SG & $0$ & $570$ & $19.94$ & $[19.43,\ 20.51]$ & $-19.94$ & collapse \\
L-BD05 & \texttt{deepseek-coder-v2-lite} & neutral & SG & $0$ & $570$ & $6.83$ & $[6.51,\ 7.17]$ & $-5.03$ & graded \\
L-BD06 & \texttt{deepseek-coder-v2-lite} & strict & SG & $0$ & $570$ & $16.02$ & $[15.55,\ 16.49]$ & $-15.99$ & collapse \\
L-BD07 & \texttt{glm-4-32b} & neutral & SG & $0$ & $570$ & $2.68$ & $[2.49,\ 2.88]$ & $+0.70$ & graded \\
L-BD08 & \texttt{glm-4-32b} & strict & SG & $0$ & $570$ & $15.58$ & $[14.99,\ 16.20]$ & $-15.46$ & collapse \\
L-BD09 & \texttt{gemma-3-27b} & neutral & SG & $0$ & $570$ & $4.32$ & $[4.11,\ 4.56]$ & $-1.64$ & graded \\
L-BD10 & \texttt{gemma-3-27b} & strict & SG & $0$ & $569$ & $11.69$ & $[11.31,\ 12.11]$ & $-11.64$ & collapse \\
L-C01 & \texttt{qwen2.5-coder-32b} & strict & SGB & $0$ & $570$ & $20.29$ & $[19.83,\ 20.74]$ & $-20.28$ & collapse \\
L-C02 & \texttt{qwen2.5-coder-32b} & lenient & SGB & $0$ & $570$ & $3.95$ & $[3.75,\ 4.17]$ & $-2.15$ & graded \\
L-D30M & \texttt{qwen3-coder-30b-a3b} & neutral & SGBR & $0$ & $570$ & $4.49$ & $[4.28,\ 4.71]$ & $-2.70$ & graded \\
L-DNxt & \texttt{qwen3-coder-next} & neutral & SGBR & $0$ & $570$ & $5.18$ & $[4.93,\ 5.43]$ & $-4.75$ & graded \\
L-E01 & \texttt{qwen2.5-coder-32b} & neutral & SGB & $0.5$ & $50$ & $7.15$ & $[6.26,\ 8.04]$ & $-7.02$ & graded \\
L-E02 & \texttt{qwen3-coder-next} & neutral & SGB & $0.5$ & $48$ & $5.38$ & $[4.64,\ 6.16]$ & $-5.20$ & graded \\
L-F01 & \texttt{gemma-3-12b} & neutral & SGB & $0$ & $570$ & $5.20$ & $[4.94,\ 5.48]$ & $-2.75$ & graded \\
L-F02 & \texttt{gemma-3-12b} & strict & SGB & $0$ & $570$ & $8.77$ & $[8.42,\ 9.15]$ & $-8.43$ & collapse \\
L-F03 & \texttt{gemma-3-12b} & lenient & SGB & $0$ & $570$ & $4.79$ & $[4.54,\ 5.06]$ & $-0.82$ & graded \\
L-F04 & \texttt{gemma-3-12b} & rigorous & SGB & $0$ & $570$ & $5.45$ & $[5.19,\ 5.73]$ & $-3.19$ & graded \\
L-F05 & \texttt{gemma-3-12b} & exacting & SGB & $0$ & $570$ & $5.10$ & $[4.85,\ 5.38]$ & $-2.76$ & graded \\
L-G01 & \texttt{qwen3-coder-30b-a3b} & neutral & GB & $0$ & $570$ & $3.60$ & $[3.37,\ 3.86]$ & $+1.29$ & graded \\
L-G02 & \texttt{qwen3-coder-30b-a3b} & neutral & SB & $0$ & $570$ & $3.63$ & $[3.42,\ 3.85]$ & $-0.18$ & graded \\
L-G04 & \texttt{qwen3-coder-30b-a3b} & neutral & SG & $0$ & $570$ & $4.80$ & $[4.59,\ 5.03]$ & $-3.49$ & graded \\
L-H01 & \texttt{qwen2.5-coder-7b} & neutral & GB & $0$ & $570$ & $4.29$ & $[4.00,\ 4.62]$ & $+2.16$ & graded \\
L-H02 & \texttt{qwen2.5-coder-7b} & neutral & SB & $0$ & $570$ & $5.25$ & $[4.99,\ 5.53]$ & $-4.02$ & graded \\
L-H04 & \texttt{qwen2.5-coder-7b} & neutral & SG & $0$ & $570$ & $5.83$ & $[5.56,\ 6.14]$ & $-4.92$ & graded \\
L-J01 & \texttt{glm-4.5-air} & neutral & SGB & $0$ & $570$ & $5.66$ & $[5.38,\ 5.95]$ & $-5.37$ & graded \\
L-J02 & \texttt{glm-4.5-air} & strict & SGB & $0$ & $570$ & $21.02$ & $[20.49,\ 21.55]$ & $-21.02$ & collapse \\
L-J03 & \texttt{glm-4.5-air} & lenient & SGB & $0$ & $570$ & $4.68$ & $[4.39,\ 4.99]$ & $+4.51$ & graded \\
L-J04 & \texttt{glm-4.5-air} & rigorous & SGB & $0$ & $568$ & $7.29$ & $[6.99,\ 7.64]$ & $-7.21$ & graded \\
L-J05 & \texttt{glm-4.5-air} & exacting & SGB & $0$ & $570$ & $11.29$ & $[10.92,\ 11.65]$ & $-11.27$ & collapse \\
L-K01 & \texttt{qwen3-coder-30b-a3b} & neutral & SGBF & $0$ & $570$ & $3.34$ & $[3.16,\ 3.53]$ & $-1.77$ & graded \\
L-M01 & \texttt{qwen3-coder-30b-a3b} & strict & SGB & $0$ & $570$ & $14.91$ & $[14.52,\ 15.30]$ & $-14.91$ & collapse \\
L-M02 & \texttt{qwen3-coder-30b-a3b} & lenient & SGB & $0$ & $570$ & $4.84$ & $[4.50,\ 5.17]$ & $+2.66$ & graded \\
L-M03 & \texttt{qwen3-coder-next} & strict & SGB & $0$ & $570$ & $11.70$ & $[11.35,\ 12.04]$ & $-11.70$ & collapse \\
L-M04 & \texttt{qwen3-coder-next} & lenient & SGB & $0$ & $570$ & $2.81$ & $[2.64,\ 2.98]$ & $-0.41$ & graded \\
L-M05 & \texttt{qwen2.5-coder-7b} & strict & SGB & $0$ & $570$ & $7.92$ & $[7.54,\ 8.30]$ & $-7.41$ & graded \\
L-M06 & \texttt{qwen2.5-coder-7b} & lenient & SGB & $0$ & $570$ & $3.88$ & $[3.65,\ 4.11]$ & $-0.86$ & graded \\
L-M07 & \texttt{qwen2.5-coder-14b} & strict & SGB & $0$ & $570$ & $24.52$ & $[23.94,\ 25.13]$ & $-24.52$ & collapse \\
L-MP01 & \texttt{qwen2.5-coder-32b} & mpstrict & SGB & $0$ & $570$ & $16.41$ & $[16.01,\ 16.80]$ & $-16.40$ & collapse \\
L-MP02 & \texttt{qwen2.5-coder-32b} & mprigorous & SGB & $0$ & $570$ & $17.74$ & $[17.29,\ 18.16]$ & $-17.73$ & collapse \\
L-MP03 & \texttt{qwen2.5-coder-32b} & mpfair & SGB & $0$ & $570$ & $18.55$ & $[18.11,\ 19.00]$ & $-18.54$ & collapse \\
L-MP04 & \texttt{qwen2.5-coder-32b} & mpnoclause & SGB & $0$ & $570$ & $15.55$ & $[15.16,\ 15.95]$ & $-15.55$ & collapse \\
L-MP05 & \texttt{qwen2.5-coder-32b} & mpnharsh & SGB & $0$ & $570$ & $8.25$ & $[7.93,\ 8.55]$ & $-8.13$ & collapse \\
L-MP06 & \texttt{mistral-small-24b} & mpstrict & SGB & $0$ & $570$ & $25.99$ & $[25.42,\ 26.57]$ & $-25.99$ & near-refusal \\
L-MP07 & \texttt{mistral-small-24b} & mprigorous & SGB & $0$ & $570$ & $25.87$ & $[25.29,\ 26.45]$ & $-25.87$ & near-refusal \\
L-MP08 & \texttt{mistral-small-24b} & mpfair & SGB & $0$ & $570$ & $25.05$ & $[24.48,\ 25.64]$ & $-25.05$ & near-refusal \\
L-MP09 & \texttt{mistral-small-24b} & mpnoclause & SGB & $0$ & $570$ & $10.40$ & $[10.06,\ 10.73]$ & $-10.34$ & collapse \\
L-MP10 & \texttt{mistral-small-24b} & mpnharsh & SGB & $0$ & $570$ & $4.23$ & $[4.02,\ 4.46]$ & $-3.31$ & graded \\
L-MP11 & \texttt{llama-3.3-70b} & mpstrict & SGB & $0$ & $570$ & $12.67$ & $[12.28,\ 13.07]$ & $-12.66$ & collapse \\
L-MP12 & \texttt{llama-3.3-70b} & mprigorous & SGB & $0$ & $570$ & $11.37$ & $[11.00,\ 11.74]$ & $-11.35$ & collapse \\
L-MP13 & \texttt{llama-3.3-70b} & mpfair & SGB & $0$ & $570$ & $11.41$ & $[11.04,\ 11.78]$ & $-11.39$ & collapse \\
L-MP14 & \texttt{llama-3.3-70b} & mpnoclause & SGB & $0$ & $570$ & $8.14$ & $[7.83,\ 8.43]$ & $-8.03$ & collapse \\
L-MP15 & \texttt{llama-3.3-70b} & mpnharsh & SGB & $0$ & $570$ & $4.60$ & $[4.38,\ 4.81]$ & $-3.71$ & graded \\
L-MP16 & \texttt{glm-4-32b} & mpstrict & SGB & $0$ & $570$ & $7.02$ & $[6.58,\ 7.49]$ & $-6.61$ & graded \\
L-MP17 & \texttt{glm-4-32b} & mprigorous & SGB & $0$ & $570$ & $7.43$ & $[7.00,\ 7.86]$ & $-7.12$ & graded \\
L-MP18 & \texttt{glm-4-32b} & mpfair & SGB & $0$ & $570$ & $6.41$ & $[6.00,\ 6.82]$ & $-5.85$ & graded \\
L-MP19 & \texttt{glm-4-32b} & mpnoclause & SGB & $0$ & $570$ & $5.51$ & $[5.22,\ 5.83]$ & $-5.10$ & graded \\
L-MP20 & \texttt{glm-4-32b} & mpnharsh & SGB & $0$ & $570$ & $2.88$ & $[2.71,\ 3.08]$ & $-0.49$ & graded \\
L-MP21 & \texttt{gemma-3-27b} & mpstrict & SGB & $0$ & $570$ & $10.49$ & $[10.12,\ 10.85]$ & $-10.42$ & collapse \\
L-MP22 & \texttt{gemma-3-27b} & mprigorous & SGB & $0$ & $570$ & $9.23$ & $[8.87,\ 9.57]$ & $-9.09$ & collapse \\
L-MP23 & \texttt{gemma-3-27b} & mpfair & SGB & $0$ & $570$ & $8.60$ & $[8.27,\ 8.93]$ & $-8.42$ & collapse \\
L-MP24 & \texttt{gemma-3-27b} & mpnoclause & SGB & $0$ & $570$ & $9.90$ & $[9.52,\ 10.26]$ & $-9.73$ & collapse \\
L-MP25 & \texttt{gemma-3-27b} & mpnharsh & SGB & $0$ & $570$ & $4.56$ & $[4.34,\ 4.79]$ & $-2.36$ & graded \\
L-N01 & \texttt{qwen2.5-72b} & neutral & SGB & $0$ & $570$ & $3.14$ & $[2.95,\ 3.35]$ & $+0.19$ & graded \\
L-N02 & \texttt{qwen2.5-72b} & strict & SGB & $0$ & $570$ & $9.82$ & $[9.46,\ 10.18]$ & $-9.78$ & collapse \\
L-N03 & \texttt{qwen2.5-72b} & lenient & SGB & $0$ & $570$ & $4.47$ & $[4.17,\ 4.81]$ & $+3.69$ & graded \\
L-N04 & \texttt{qwen2.5-72b} & rigorous & SGB & $0$ & $570$ & $3.09$ & $[2.91,\ 3.28]$ & $-0.29$ & graded \\
L-N05 & \texttt{qwen2.5-72b} & exacting & SGB & $0$ & $570$ & $3.43$ & $[3.25,\ 3.62]$ & $-1.57$ & graded \\
L-P01 & \texttt{qwen2.5-coder-32b} & rigorous & SGB & $0$ & $570$ & $8.70$ & $[8.38,\ 9.01]$ & $-8.61$ & collapse \\
L-P02 & \texttt{qwen2.5-coder-32b} & exacting & SGB & $0$ & $570$ & $10.23$ & $[9.91,\ 10.56]$ & $-10.20$ & collapse \\
L-PS01 & \texttt{qwen2.5-coder-32b} & mpframe & SGB & $0$ & $570$ & $9.37$ & $[9.04,\ 9.68]$ & $-9.31$ & collapse \\
L-PS02 & \texttt{qwen2.5-coder-32b} & mps2only & SGB & $0$ & $570$ & $12.92$ & $[12.56,\ 13.27]$ & $-12.91$ & collapse \\
L-PS03 & \texttt{qwen2.5-coder-32b} & mpnoclause & SGB & $0$ & $570$ & $15.55$ & $[15.16,\ 15.95]$ & $-15.55$ & collapse \\
L-PS04 & \texttt{qwen2.5-coder-32b} & mpharsh & SGB & $0$ & $570$ & $20.26$ & $[19.79,\ 20.73]$ & $-20.26$ & collapse \\
L-PS05 & \texttt{mistral-small-24b} & mpframe & SGB & $0$ & $570$ & $5.62$ & $[5.36,\ 5.89]$ & $-5.33$ & graded \\
L-PS06 & \texttt{mistral-small-24b} & mps2only & SGB & $0$ & $570$ & $26.04$ & $[25.45,\ 26.61]$ & $-26.04$ & refusal \\
L-PS07 & \texttt{mistral-small-24b} & mpnoclause & SGB & $0$ & $570$ & $10.33$ & $[9.99,\ 10.66]$ & $-10.28$ & collapse \\
L-PS08 & \texttt{mistral-small-24b} & mpharsh & SGB & $0$ & $570$ & $26.03$ & $[25.44,\ 26.61]$ & $-26.03$ & refusal \\
L-PS09 & \texttt{llama-3.1-8b} & mpframe & SGB & $0$ & $570$ & $8.46$ & $[8.10,\ 8.86]$ & $-6.94$ & collapse \\
L-PS10 & \texttt{llama-3.1-8b} & mps2only & SGB & $0$ & $570$ & $26.04$ & $[25.45,\ 26.61]$ & $-26.04$ & refusal \\
L-PS11 & \texttt{llama-3.1-8b} & mpnoclause & SGB & $0$ & $570$ & $23.45$ & $[22.89,\ 23.99]$ & $-23.45$ & collapse \\
L-PS12 & \texttt{llama-3.1-8b} & mpharsh & SGB & $0$ & $570$ & $26.04$ & $[25.45,\ 26.61]$ & $-26.04$ & refusal \\
L-Q01 & \texttt{qwen3-235b-a22b} & neutral & SGB & $0$ & $570$ & $4.09$ & $[3.87,\ 4.32]$ & $-3.56$ & graded \\
L-Q02 & \texttt{qwen3-235b-a22b} & strict & SGB & $0$ & $570$ & $7.32$ & $[7.01,\ 7.61]$ & $-7.24$ & graded \\
L-Q03 & \texttt{qwen3-235b-a22b} & lenient & SGB & $0$ & $570$ & $6.01$ & $[5.65,\ 6.37]$ & $+5.94$ & graded \\
L-Q04 & \texttt{qwen3-235b-a22b} & rigorous & SGB & $0$ & $570$ & $5.18$ & $[4.92,\ 5.45]$ & $-4.90$ & graded \\
L-Q05 & \texttt{qwen3-235b-a22b} & exacting & SGB & $0$ & $570$ & $6.85$ & $[6.56,\ 7.13]$ & $-6.75$ & graded \\
L-R01 & \texttt{qwen3-coder-480b} & neutral & SGB & $0$ & $570$ & $3.14$ & $[2.95,\ 3.35]$ & $+0.51$ & graded \\
L-R02 & \texttt{qwen3-coder-480b} & strict & SGB & $0$ & $570$ & $3.78$ & $[3.58,\ 3.98]$ & $-2.52$ & graded \\
L-R03 & \texttt{qwen3-coder-480b} & lenient & SGB & $0$ & $570$ & $4.13$ & $[3.83,\ 4.44]$ & $+3.38$ & graded \\
L-R04 & \texttt{qwen3-coder-480b} & rigorous & SGB & $0$ & $570$ & $3.06$ & $[2.88,\ 3.26]$ & $-0.23$ & graded \\
L-R05 & \texttt{qwen3-coder-480b} & exacting & SGB & $0$ & $570$ & $3.33$ & $[3.14,\ 3.52]$ & $-1.56$ & graded \\
L-S01 & \texttt{deepseek-coder-v2-lite} & neutral & SGB & $0$ & $570$ & $5.98$ & $[5.69,\ 6.29]$ & $-3.44$ & graded \\
L-S02 & \texttt{deepseek-coder-v2-lite} & strict & SGB & $0$ & $570$ & $12.09$ & $[11.66,\ 12.52]$ & $-11.87$ & collapse \\
L-S03 & \texttt{deepseek-coder-v2-lite} & lenient & SGB & $0$ & $570$ & $5.92$ & $[5.62,\ 6.22]$ & $-3.31$ & graded \\
L-S04 & \texttt{deepseek-coder-v2-lite} & rigorous & SGB & $0$ & $570$ & $7.01$ & $[6.70,\ 7.34]$ & $-5.31$ & graded \\
L-S05 & \texttt{deepseek-coder-v2-lite} & exacting & SGB & $0$ & $570$ & $6.69$ & $[6.39,\ 7.01]$ & $-4.83$ & graded \\
L-U01 & \texttt{llama-3.1-8b} & neutral & SGB & $0$ & $570$ & $7.31$ & $[6.97,\ 7.68]$ & $-5.15$ & graded \\
L-U02 & \texttt{llama-3.1-8b} & strict & SGB & $0$ & $570$ & $26.04$ & $[25.45,\ 26.61]$ & $-26.04$ & refusal \\
L-U03 & \texttt{llama-3.1-8b} & lenient & SGB & $0$ & $570$ & $6.49$ & $[6.06,\ 6.96]$ & $+5.36$ & graded \\
L-U04 & \texttt{llama-3.1-8b} & rigorous & SGB & $0$ & $570$ & $10.31$ & $[9.88,\ 10.74]$ & $-9.43$ & collapse \\
L-U05 & \texttt{llama-3.1-8b} & exacting & SGB & $0$ & $570$ & $6.89$ & $[6.57,\ 7.23]$ & $-4.50$ & graded \\
L-V01 & \texttt{llama-3.3-70b} & neutral & SGB & $0$ & $570$ & $4.52$ & $[4.31,\ 4.73]$ & $-3.60$ & graded \\
L-V02 & \texttt{llama-3.3-70b} & strict & SGB & $0$ & $570$ & $13.52$ & $[13.11,\ 13.93]$ & $-13.52$ & collapse \\
L-V03 & \texttt{llama-3.3-70b} & lenient & SGB & $0$ & $570$ & $3.99$ & $[3.72,\ 4.28]$ & $+1.82$ & graded \\
L-V04 & \texttt{llama-3.3-70b} & rigorous & SGB & $0$ & $570$ & $4.69$ & $[4.47,\ 4.91]$ & $-3.89$ & graded \\
L-V05 & \texttt{llama-3.3-70b} & exacting & SGB & $0$ & $570$ & $6.74$ & $[6.46,\ 7.01]$ & $-6.56$ & graded \\
L-W01 & \texttt{glm-4-9b} & neutral & SGB & $0$ & $570$ & $4.31$ & $[4.02,\ 4.63]$ & $+1.68$ & graded \\
L-W02 & \texttt{glm-4-9b} & strict & SGB & $0$ & $570$ & $25.48$ & $[24.88,\ 26.04]$ & $-25.48$ & near-refusal \\
L-W03 & \texttt{glm-4-9b} & lenient & SGB & $0$ & $570$ & $6.94$ & $[6.43,\ 7.46]$ & $+6.40$ & graded \\
L-W04 & \texttt{glm-4-9b} & rigorous & SGB & $0$ & $570$ & $4.53$ & $[4.21,\ 4.87]$ & $+1.70$ & graded \\
L-W05 & \texttt{glm-4-9b} & exacting & SGB & $0$ & $570$ & $5.18$ & $[4.80,\ 5.57]$ & $+3.80$ & graded \\
L-X01 & \texttt{glm-4-32b} & neutral & SGB & $0$ & $570$ & $2.85$ & $[2.66,\ 3.05]$ & $+0.50$ & graded \\
L-X02 & \texttt{glm-4-32b} & strict & SGB & $0$ & $570$ & $9.66$ & $[9.15,\ 10.19]$ & $-9.43$ & collapse \\
L-X03 & \texttt{glm-4-32b} & lenient & SGB & $0$ & $570$ & $5.33$ & $[4.99,\ 5.69]$ & $+5.00$ & graded \\
L-X04 & \texttt{glm-4-32b} & rigorous & SGB & $0$ & $570$ & $2.74$ & $[2.56,\ 2.93]$ & $+0.03$ & graded \\
L-X05 & \texttt{glm-4-32b} & exacting & SGB & $0$ & $570$ & $3.72$ & $[3.49,\ 3.96]$ & $-2.34$ & graded \\
L-Y01 & \texttt{gemma-3-27b} & neutral & SGB & $0$ & $570$ & $4.32$ & $[4.12,\ 4.55]$ & $-1.52$ & graded \\
L-Y02 & \texttt{gemma-3-27b} & strict & SGB & $0$ & $570$ & $11.55$ & $[11.16,\ 11.92]$ & $-11.50$ & collapse \\
L-Y03 & \texttt{gemma-3-27b} & lenient & SGB & $0$ & $570$ & $4.55$ & $[4.25,\ 4.86]$ & $+2.29$ & graded \\
L-Y04 & \texttt{gemma-3-27b} & rigorous & SGB & $0$ & $570$ & $4.88$ & $[4.65,\ 5.13]$ & $-3.49$ & graded \\
L-Y05 & \texttt{gemma-3-27b} & exacting & SGB & $0$ & $570$ & $5.47$ & $[5.21,\ 5.73]$ & $-4.63$ & graded \\
L-Z01 & \texttt{mistral-small-24b} & neutral & SGB & $0$ & $570$ & $3.66$ & $[3.46,\ 3.87]$ & $-2.41$ & graded \\
L-Z02 & \texttt{mistral-small-24b} & strict & SGB & $0$ & $570$ & $26.03$ & $[25.44,\ 26.61]$ & $-26.03$ & refusal \\
L-Z03 & \texttt{mistral-small-24b} & lenient & SGB & $0$ & $570$ & $4.76$ & $[4.44,\ 5.12]$ & $+3.57$ & graded \\
L-Z04 & \texttt{mistral-small-24b} & rigorous & SGB & $0$ & $570$ & $6.15$ & $[5.87,\ 6.41]$ & $-5.89$ & graded \\
L-Z05 & \texttt{mistral-small-24b} & exacting & SGB & $0$ & $569$ & $7.04$ & $[6.75,\ 7.33]$ & $-6.93$ & graded \\
\midrule
\multicolumn{10}{l}{\emph{Prompt-only baselines (Appendix~\ref{app:promptbaselines}; not grader configurations)}}\\*
PB-A01 & \texttt{qwen2.5-coder-32b} & strict (arbitrated) & SGB & $0$ & $100$ & $9.29$ & $[8.58,\ 9.98]$ & $-9.27$ & collapse \\
PB-A02 & \texttt{glm-4-32b} & strict (arbitrated) & SGB & $0$ & $100$ & $2.64$ & $[2.23,\ 3.10]$ & $-1.53$ & graded \\
PB-A03 & \texttt{mistral-small-24b} & strict (arbitrated) & SGB & $0$ & $100$ & $3.13$ & $[2.74,\ 3.53]$ & $-2.31$ & graded \\
PB-A04 & \texttt{llama-3.1-8b} & strict (arbitrated) & SGB & $0$ & $99$ & $7.46$ & $[6.60,\ 8.28]$ & $-5.77$ & graded \\
PB-D01 & \texttt{qwen2.5-coder-32b} & strict (decomposed) & SGB & $0$ & $100$ & $20.00$ & $[19.10,\ 20.93]$ & $-20.00$ & collapse \\
PB-D02 & \texttt{glm-4-32b} & strict (decomposed) & SGB & $0$ & $100$ & $5.01$ & $[4.50,\ 5.51]$ & $-4.95$ & graded \\
PB-D03 & \texttt{mistral-small-24b} & strict (decomposed) & SGB & $0$ & $100$ & $26.27$ & $[24.98,\ 27.55]$ & $-26.27$ & collapse \\
PB-D04 & \texttt{llama-3.1-8b} & strict (decomposed) & SGB & $0$ & $100$ & $27.26$ & $[25.89,\ 28.65]$ & $-27.26$ & refusal \\
\end{longtable}
\endgroup

\section{Full Limitations Inventory}
\label{app:limitations}

\paragraph{Persona coverage.}
Qwen2.5-Coder-$14$B received no persona in the full-cohort grid and Qwen3-Coder-Next only \emph{strict}; the held-out sweep adds all three harsh wordings and \emph{lenient} on the $7$B, $14$B and $30$B (base and pooled), leaving the $80$B evaluated on only one wording. On the ML exam the $14$B received \emph{neutral} and \emph{strict} only, and Qwen3-Coder-Next no \emph{rigorous} or \emph{exacting}. The headword series covers five models and the policy $2\times2$ three, with only Qwen2.5-Coder-$32$B and Mistral-Small-$24$B in both, and we did not cross them, so an adjective-only interaction is untested. On the closed side only Gemini received all five wordings; Section~\ref{sec:limitations} itemises the cross-vendor coverage, whose decoding differs where the APIs force it (Anthropic's current models expose no temperature; \texttt{gpt-5.5} is not bit-reproducible at temperature $0$).

\paragraph{Model and serving confounds.}
\texttt{gemini-2.5-pro} rejects a zero thinking budget, so D$03$ against D$01$/D$02$ compares an older thinking-on model with newer thinking-off ones, if anything understating the newer models' advantage. The Qwen3-Coder \emph{Instruct} variants implement no thinking mode, so their thinking-on configurations are inert replicates.

\paragraph{Fine-tuning.}
Held-out MAE CIs are roughly $\pm 0.45$, so the paired test carries the verdicts. The bd arm trains on targets produced by D$01$ (CV exam) and IG$07$ (ML exam), so it measures closed-model distillation; only the marks recipe trains on human labels alone. The $30$B is not a clean size point (MoE, attention-only LoRA); base rows use the fine-tuning harness, not identical reruns of Section~\ref{sec:brittle}. vLLM batching reorders reductions ($\sim 0.01$ MAE); all evaluation, Gemma-4's adapter included, is vLLM-served, with Gemma-4's path cross-validated against HuggingFace generation to within $\sim 0.1$ MAE (Appendix~\ref{app:ft}).

\section{ML-Exam Run Table}
\label{app:iaruns}

Table~\ref{tab:iaruns} lists every one of the $162$ ML-exam configurations individually --- $30$ closed-model and $132$ open-weights runs --- with each run's own confidence interval, bias, and behaviour class. Table~\ref{tab:replication} pairs the matched neutral/strict runs for the headline comparison; this one is the flat record, and is the ML exam's counterpart to Table~\ref{tab:allruns}. The machine-readable version, with per-question breakdowns and the full column set, ships as \texttt{analysis/introduction\_to\_ai\_master\_comparison.csv} \artifactsin.

\begingroup
\scriptsize
\setlength{\tabcolsep}{2.5pt}
\setlength{\LTcapwidth}{\linewidth}
\begin{longtable}{llllcrrcrl}
\caption{Every second-exam grader configuration, one row per run: the $30$ closed-model configurations ($23$ Gemini IG-, $6$ OpenAI IO- and $1$ Anthropic IN-; Section~\ref{sec:closed:vendors}) followed by the $132$ open-weights (IA-) configurations, each in run-id order. \emph{Prompt} lists the components present (S~reference solution, B~rubric breakdown, R~thinking enabled, F~few-shot demonstrations); this exam has no student-facing guidelines document, so the G component of Table~\ref{tab:allruns} never appears. Personas prefixed \texttt{mp} are the mechanism probes of Section~\ref{sec:brittle:attrib}, replayed on this exam (keys decoded in Table~\ref{tab:allruns}'s caption). $n$ is the number of students scored against the grader average ($\ast$: IG$18$'s $693$ are the survivors of prompt-length failures, an easier subsample whose MAE is not comparable to full-cohort rows; Appendix~\ref{app:replication}); MAE and bias are on the $\approx 65$-point scale (score plus bonus, matching how its graders record marks), the $95\%$ bootstrap CI (\num{50000} resamples) on MAE. Behaviour classes are those of Section~\ref{sec:brittle:numbers}, at this exam's floor-matched collapse band ($\text{MAE} \ge 15.7$), and describe a run's error level rather than persona damage: DeepSeek-Coder-V2-Lite over-marks badly enough to leave it unprompted (IA$11$, IA$13$). A refusal's MAE is a ceiling artifact, not a severity measurement. IG$15$--IG$20$ ran on the first $100$ students; IG$10$--IG$12$, IA$92$ and IA$136$ are repeated-sampling probes whose scores are the per-student mean across five reruns (Appendix~\ref{app:replication}). \texttt{claude-opus-5} exposes no sampling temperature (thinking disabled; shown as ---). Run ids match the released tracker and spreadsheets. Generated by \texttt{analysis/introduction\_to\_ai\_make\_appendix\_table.py}.}\label{tab:iaruns}\\
\toprule
Run & Model & Persona & Prompt & $t$ & $n$ & MAE & $95\%$ CI & Bias & Behaviour \\
\midrule
\endfirsthead
\multicolumn{10}{l}{\emph{Table~\ref{tab:iaruns} continued}}\\
\toprule
Run & Model & Persona & Prompt & $t$ & $n$ & MAE & $95\%$ CI & Bias & Behaviour \\
\midrule
\endhead
\bottomrule
\endfoot
\multicolumn{10}{l}{\emph{Closed-model configurations (Gemini IG-, OpenAI IO-, Anthropic IN-)}}\\*
IG01 & \texttt{flash-lite} & neutral & SB & $0$ & $1013$ & $7.53$ & $[7.20,\ 7.87]$ & $+5.98$ & graded \\
IG02 & \texttt{flash-lite} & neutral & B & $0$ & $1032$ & $12.46$ & $[12.03,\ 12.89]$ & $+12.16$ & graded \\
IG03 & \texttt{flash-lite} & neutral & SBR & $0$ & $1037$ & $5.37$ & $[5.09,\ 5.66]$ & $-4.04$ & graded \\
IG04 & \texttt{flash-lite} & neutral & S & $0$ & $1016$ & $6.88$ & $[6.57,\ 7.20]$ & $+5.27$ & graded \\
IG05 & \texttt{flash-lite} & strict & SB & $0$ & $1008$ & $9.02$ & $[8.61,\ 9.43]$ & $-7.83$ & graded \\
IG06 & \texttt{flash-lite} & lenient & SB & $0$ & $1020$ & $17.86$ & $[17.30,\ 18.42]$ & $+17.80$ & collapse \\
IG07 & \texttt{3-flash-preview} & neutral & SB & $0$ & $1038$ & $3.92$ & $[3.68,\ 4.16]$ & $+2.24$ & graded \\
IG08 & \texttt{3.1-pro-preview} & neutral & SB & $0$ & $1038$ & $3.40$ & $[3.16,\ 3.65]$ & $+1.18$ & graded \\
IG09 & \texttt{2.5-pro} & neutral & SB & $0$ & $1035$ & $3.47$ & $[3.24,\ 3.72]$ & $-0.83$ & graded \\
IG10 & \texttt{flash-lite} & neutral & SB & $0$ & $955$ & $7.78$ & $[7.44,\ 8.13]$ & $+6.14$ & graded \\
IG11 & \texttt{flash-lite} & neutral & SB & $0.5$ & $963$ & $7.65$ & $[7.32,\ 8.00]$ & $+5.95$ & graded \\
IG12 & \texttt{flash-lite} & neutral & SB & $0.7$ & $963$ & $7.63$ & $[7.29,\ 7.98]$ & $+5.85$ & graded \\
IG13 & \texttt{flash-lite} & strict & SBR & $0$ & $1038$ & $7.76$ & $[7.43,\ 8.09]$ & $-7.06$ & graded \\
IG14 & \texttt{3.1-pro-preview} & lenient & SB & $0$ & $1033$ & $4.48$ & $[4.22,\ 4.75]$ & $+3.46$ & graded \\
IG15 & \texttt{3-flash-preview} & neutral & B & $0$ & $100$ & $4.35$ & $[3.32,\ 5.64]$ & $+3.51$ & graded \\
IG16 & \texttt{3-flash-preview} & neutral & SBR & $0$ & $100$ & $3.76$ & $[2.78,\ 5.01]$ & $+2.78$ & graded \\
IG17 & \texttt{3-flash-preview} & neutral & S & $0$ & $96$ & $3.37$ & $[2.42,\ 4.70]$ & $+2.38$ & graded \\
IG18 & \texttt{flash-lite} & neutral & SBF & $0$ & $693$$^{\ast}$ & $4.30$ & $[4.01,\ 4.60]$ & $+0.36$ & graded \\
IG19 & \texttt{3-flash-preview} & strict & SB & $0$ & $100$ & $2.87$ & $[1.95,\ 4.10]$ & $+0.09$ & graded \\
IG20 & \texttt{3-flash-preview} & lenient & SB & $0$ & $100$ & $6.99$ & $[5.76,\ 8.43]$ & $+6.60$ & graded \\
IG21 & \texttt{flash-lite} & rigorous & SB & $0$ & $1012$ & $5.52$ & $[5.25,\ 5.79]$ & $+2.79$ & graded \\
IG22 & \texttt{flash-lite} & exacting & SB & $0$ & $1021$ & $5.96$ & $[5.68,\ 6.25]$ & $+3.18$ & graded \\
IG23 & \texttt{3.1-pro-preview} & strict & SB & $0$ & $1038$ & $3.67$ & $[3.43,\ 3.94]$ & $-1.00$ & graded \\
IN01 & \texttt{claude-opus-5} & neutral & SB & --- & $1038$ & $4.37$ & $[4.13,\ 4.63]$ & $-2.64$ & graded \\
IO01 & \texttt{gpt-5.5} & neutral & SB & $0$ & $1038$ & $3.54$ & $[3.33,\ 3.77]$ & $+1.23$ & graded \\
IO02 & \texttt{gpt-5.5} & strict & SB & $0$ & $1038$ & $3.70$ & $[3.48,\ 3.93]$ & $-1.38$ & graded \\
IO03 & \texttt{gpt-5.5} & lenient & SB & $0$ & $1038$ & $6.29$ & $[6.02,\ 6.57]$ & $+5.83$ & graded \\
IO11 & \texttt{gpt-5.4} & neutral & SB & $0$ & $1038$ & $4.30$ & $[4.07,\ 4.55]$ & $-1.95$ & graded \\
IO12 & \texttt{gpt-5.4} & strict & SB & $0$ & $1038$ & $5.11$ & $[4.85,\ 5.37]$ & $-3.61$ & graded \\
IO13 & \texttt{gpt-5.4} & lenient & SB & $0$ & $1038$ & $8.30$ & $[7.96,\ 8.64]$ & $+7.94$ & graded \\
\midrule
\multicolumn{10}{l}{\emph{Open-weights (IA-) configurations}}\\*
IA01-G32-n & \texttt{glm4-32b} & neutral & SB & $0$ & $1038$ & $8.03$ & $[7.69,\ 8.37]$ & $+7.29$ & graded \\
IA01-G9-n & \texttt{glm4-9b} & neutral & SB & $0$ & $1038$ & $13.15$ & $[12.60,\ 13.71]$ & $+12.10$ & graded \\
IA01-GE27-n & \texttt{gemma3-27b} & neutral & SB & $0$ & $1038$ & $8.33$ & $[8.00,\ 8.67]$ & $+6.83$ & graded \\
IA01-L8-n & \texttt{llama31-8b} & neutral & SB & $0$ & $1037$ & $13.16$ & $[12.61,\ 13.71]$ & $+10.27$ & graded \\
IA01-M24-n & \texttt{mistral-small-24b} & neutral & SB & $0$ & $1038$ & $8.96$ & $[8.59,\ 9.34]$ & $+8.16$ & graded \\
IA01-Q14-n & \texttt{qwen25-coder-14b} & neutral & SB & $0$ & $1038$ & $8.61$ & $[8.25,\ 8.98]$ & $+7.40$ & graded \\
IA01-Q30-n & \texttt{qwen3-coder-30b-a3b} & neutral & SB & $0$ & $1038$ & $11.12$ & $[10.64,\ 11.59]$ & $+10.44$ & graded \\
IA01-Q32-n & \texttt{qwen25-coder-32b} & neutral & SB & $0$ & $1038$ & $4.57$ & $[4.33,\ 4.83]$ & $-0.21$ & graded \\
IA01-Q7-n & \texttt{qwen25-coder-7b} & neutral & SB & $0$ & $1038$ & $6.88$ & $[6.56,\ 7.20]$ & $+3.77$ & graded \\
IA02-G32-s & \texttt{glm4-32b} & strict & SB & $0$ & $1038$ & $12.18$ & $[11.68,\ 12.69]$ & $-10.68$ & graded \\
IA02-G9-s & \texttt{glm4-9b} & strict & SB & $0$ & $1038$ & $16.68$ & $[16.15,\ 17.20]$ & $-16.52$ & collapse \\
IA02-GE27-s & \texttt{gemma3-27b} & strict & SB & $0$ & $1038$ & $7.61$ & $[7.23,\ 7.99]$ & $-6.25$ & graded \\
IA02-L8-s & \texttt{llama31-8b} & strict & SB & $0$ & $1038$ & $29.58$ & $[28.70,\ 30.45]$ & $-29.58$ & refusal \\
IA02-M24-s & \texttt{mistral-small-24b} & strict & SB & $0$ & $1038$ & $17.32$ & $[16.77,\ 17.87]$ & $-17.20$ & collapse \\
IA02-Q14-s & \texttt{qwen25-coder-14b} & strict & SB & $0$ & $1038$ & $11.06$ & $[10.61,\ 11.50]$ & $-10.46$ & graded \\
IA02-Q30-s & \texttt{qwen3-coder-30b-a3b} & strict & SB & $0$ & $1038$ & $5.58$ & $[5.31,\ 5.87]$ & $-1.55$ & graded \\
IA02-Q32-s & \texttt{qwen25-coder-32b} & strict & SB & $0$ & $1038$ & $13.24$ & $[12.79,\ 13.69]$ & $-13.07$ & graded \\
IA02-Q7-s & \texttt{qwen25-coder-7b} & strict & SB & $0$ & $1038$ & $7.45$ & $[7.11,\ 7.80]$ & $-5.67$ & graded \\
IA11 & \texttt{deepseek-coder-v2-lite} & neutral & S & $0$ & $1038$ & $16.48$ & $[15.82,\ 17.15]$ & $+14.96$ & collapse \\
IA12 & \texttt{deepseek-coder-v2-lite} & strict & S & $0$ & $1038$ & $9.12$ & $[8.70,\ 9.54]$ & $-5.19$ & graded \\
IA13 & \texttt{deepseek-coder-v2-lite} & neutral & SB & $0$ & $1038$ & $16.41$ & $[15.76,\ 17.07]$ & $+14.89$ & collapse \\
IA14 & \texttt{deepseek-coder-v2-lite} & strict & SB & $0$ & $1038$ & $9.22$ & $[8.83,\ 9.62]$ & $+2.42$ & graded \\
IA15 & \texttt{deepseek-coder-v2-lite} & lenient & SB & $0$ & $1038$ & $16.29$ & $[15.64,\ 16.95]$ & $+14.96$ & collapse \\
IA16 & \texttt{deepseek-coder-v2-lite} & rigorous & SB & $0$ & $1038$ & $14.23$ & $[13.64,\ 14.83]$ & $+12.29$ & graded \\
IA17 & \texttt{deepseek-coder-v2-lite} & exacting & SB & $0$ & $1038$ & $15.50$ & $[14.87,\ 16.13]$ & $+13.90$ & graded \\
IA18 & \texttt{gemma3-12b} & neutral & SB & $0$ & $1038$ & $10.36$ & $[9.96,\ 10.76]$ & $+9.13$ & graded \\
IA19 & \texttt{gemma3-12b} & strict & SB & $0$ & $1038$ & $6.79$ & $[6.49,\ 7.10]$ & $+0.09$ & graded \\
IA20 & \texttt{gemma3-12b} & lenient & SB & $0$ & $1038$ & $13.70$ & $[13.22,\ 14.18]$ & $+13.17$ & graded \\
IA21 & \texttt{gemma3-12b} & rigorous & SB & $0$ & $1038$ & $8.51$ & $[8.16,\ 8.86]$ & $+6.02$ & graded \\
IA22 & \texttt{gemma3-12b} & exacting & SB & $0$ & $1038$ & $8.71$ & $[8.36,\ 9.06]$ & $+6.48$ & graded \\
IA23 & \texttt{gemma3-27b} & neutral & S & $0$ & $1038$ & $8.09$ & $[7.77,\ 8.42]$ & $+6.55$ & graded \\
IA24 & \texttt{gemma3-27b} & strict & S & $0$ & $1038$ & $7.35$ & $[6.98,\ 7.72]$ & $-5.88$ & graded \\
IA25 & \texttt{gemma3-27b} & mpstrict & SB & $0$ & $1038$ & $6.77$ & $[6.43,\ 7.12]$ & $-4.90$ & graded \\
IA26 & \texttt{gemma3-27b} & mprigorous & SB & $0$ & $1038$ & $6.31$ & $[5.98,\ 6.64]$ & $-4.05$ & graded \\
IA27 & \texttt{gemma3-27b} & mpfair & SB & $0$ & $1038$ & $5.93$ & $[5.62,\ 6.25]$ & $-3.35$ & graded \\
IA28 & \texttt{gemma3-27b} & mpnoclause & SB & $0$ & $1038$ & $6.48$ & $[6.16,\ 6.79]$ & $-2.35$ & graded \\
IA29 & \texttt{gemma3-27b} & mpnharsh & SB & $0$ & $1038$ & $7.78$ & $[7.47,\ 8.10]$ & $+5.74$ & graded \\
IA30 & \texttt{gemma3-27b} & lenient & SB & $0$ & $1038$ & $18.54$ & $[18.02,\ 19.06]$ & $+18.47$ & collapse \\
IA31 & \texttt{gemma3-27b} & rigorous & SB & $0$ & $1038$ & $6.18$ & $[5.91,\ 6.46]$ & $+3.18$ & graded \\
IA32 & \texttt{gemma3-27b} & exacting & SB & $0$ & $1038$ & $5.95$ & $[5.69,\ 6.23]$ & $+1.83$ & graded \\
IA33 & \texttt{glm4-32b} & neutral & S & $0$ & $1038$ & $7.51$ & $[7.19,\ 7.84]$ & $+6.75$ & graded \\
IA34 & \texttt{glm4-32b} & strict & S & $0$ & $1038$ & $18.44$ & $[17.86,\ 19.03]$ & $-17.63$ & collapse \\
IA35 & \texttt{glm4-32b} & mpstrict & SB & $0$ & $1038$ & $7.29$ & $[6.92,\ 7.67]$ & $-4.76$ & graded \\
IA36 & \texttt{glm4-32b} & mprigorous & SB & $0$ & $1038$ & $7.11$ & $[6.74,\ 7.50]$ & $-4.74$ & graded \\
IA37 & \texttt{glm4-32b} & mpfair & SB & $0$ & $1038$ & $6.74$ & $[6.38,\ 7.11]$ & $-3.85$ & graded \\
IA38 & \texttt{glm4-32b} & mpnoclause & SB & $0$ & $1038$ & $5.60$ & $[5.29,\ 5.92]$ & $-3.40$ & graded \\
IA39 & \texttt{glm4-32b} & mpnharsh & SB & $0$ & $1038$ & $6.83$ & $[6.52,\ 7.15]$ & $+5.82$ & graded \\
IA40 & \texttt{glm4-32b} & lenient & SB & $0$ & $1038$ & $15.35$ & $[14.90,\ 15.80]$ & $+15.23$ & graded \\
IA41 & \texttt{glm4-32b} & rigorous & SB & $0$ & $1038$ & $7.47$ & $[7.15,\ 7.80]$ & $+6.64$ & graded \\
IA42 & \texttt{glm4-32b} & exacting & SB & $0$ & $1038$ & $5.57$ & $[5.28,\ 5.86]$ & $+2.97$ & graded \\
IA43 & \texttt{glm4-9b} & lenient & SB & $0$ & $1038$ & $25.98$ & $[25.22,\ 26.74]$ & $+25.96$ & collapse \\
IA44 & \texttt{glm4-9b} & rigorous & SB & $0$ & $1038$ & $11.80$ & $[11.29,\ 12.31]$ & $+10.49$ & graded \\
IA45 & \texttt{glm4-9b} & exacting & SB & $0$ & $1038$ & $14.71$ & $[14.11,\ 15.33]$ & $+14.05$ & graded \\
IA46 & \texttt{llama31-8b} & mpframe & SB & $0$ & $1038$ & $11.54$ & $[11.07,\ 12.02]$ & $+7.78$ & graded \\
IA47 & \texttt{llama31-8b} & mps2only & SB & $0$ & $1038$ & $29.39$ & $[28.52,\ 30.25]$ & $-29.37$ & near-refusal \\
IA48 & \texttt{llama31-8b} & mpnoclause & SB & $0$ & $1038$ & $25.87$ & $[25.08,\ 26.65]$ & $-25.84$ & collapse \\
IA49 & \texttt{llama31-8b} & mpharsh & SB & $0$ & $1038$ & $29.58$ & $[28.70,\ 30.45]$ & $-29.58$ & refusal \\
IA50 & \texttt{llama31-8b} & lenient & SB & $0$ & $1037$ & $27.73$ & $[26.92,\ 28.54]$ & $+27.69$ & collapse \\
IA51 & \texttt{llama31-8b} & rigorous & SB & $0$ & $1038$ & $9.01$ & $[8.61,\ 9.42]$ & $+3.07$ & graded \\
IA52 & \texttt{llama31-8b} & exacting & SB & $0$ & $1038$ & $11.96$ & $[11.45,\ 12.48]$ & $+9.15$ & graded \\
IA53 & \texttt{mistral-small-24b} & mpstrict & SB & $0$ & $1038$ & $12.49$ & $[12.03,\ 12.96]$ & $-12.14$ & graded \\
IA54 & \texttt{mistral-small-24b} & mprigorous & SB & $0$ & $1038$ & $8.81$ & $[8.43,\ 9.21]$ & $-7.67$ & graded \\
IA55 & \texttt{mistral-small-24b} & mpfair & SB & $0$ & $1038$ & $8.63$ & $[8.24,\ 9.02]$ & $-7.41$ & graded \\
IA56 & \texttt{mistral-small-24b} & mpnoclause & SB & $0$ & $1038$ & $6.37$ & $[6.08,\ 6.67]$ & $-0.67$ & graded \\
IA57 & \texttt{mistral-small-24b} & mpnharsh & SB & $0$ & $1038$ & $8.09$ & $[7.74,\ 8.45]$ & $+6.94$ & graded \\
IA58 & \texttt{mistral-small-24b} & mpframe & SB & $0$ & $1038$ & $6.58$ & $[6.27,\ 6.90]$ & $+4.10$ & graded \\
IA59 & \texttt{mistral-small-24b} & mps2only & SB & $0$ & $1038$ & $26.03$ & $[25.29,\ 26.77]$ & $-26.02$ & collapse \\
IA60 & \texttt{mistral-small-24b} & mpharsh & SB & $0$ & $1037$ & $17.29$ & $[16.73,\ 17.84]$ & $-17.17$ & collapse \\
IA61 & \texttt{mistral-small-24b} & lenient & SB & $0$ & $1038$ & $15.56$ & $[15.08,\ 16.05]$ & $+15.44$ & graded \\
IA62 & \texttt{mistral-small-24b} & rigorous & SB & $0$ & $1038$ & $5.89$ & $[5.61,\ 6.19]$ & $+2.96$ & graded \\
IA63 & \texttt{mistral-small-24b} & exacting & SB & $0$ & $1038$ & $5.70$ & $[5.43,\ 5.98]$ & $+2.41$ & graded \\
IA64 & \texttt{qwen25-coder-32b} & neutral & S & $0$ & $1038$ & $4.45$ & $[4.21,\ 4.71]$ & $+0.13$ & graded \\
IA65 & \texttt{qwen25-coder-32b} & strict & S & $0$ & $1038$ & $13.43$ & $[12.97,\ 13.89]$ & $-13.25$ & graded \\
IA66 & \texttt{qwen25-coder-32b} & lenient & SB & $0$ & $1038$ & $6.96$ & $[6.66,\ 7.26]$ & $+5.56$ & graded \\
IA67 & \texttt{qwen25-coder-32b} & mpstrict & SB & $0$ & $1038$ & $9.65$ & $[9.26,\ 10.04]$ & $-9.30$ & graded \\
IA68 & \texttt{qwen25-coder-32b} & mprigorous & SB & $0$ & $1038$ & $10.19$ & $[9.80,\ 10.59]$ & $-9.87$ & graded \\
IA69 & \texttt{qwen25-coder-32b} & mpfair & SB & $0$ & $1038$ & $10.40$ & $[10.00,\ 10.81]$ & $-10.09$ & graded \\
IA70 & \texttt{qwen25-coder-32b} & mpnoclause & SB & $0$ & $1038$ & $9.45$ & $[9.05,\ 9.85]$ & $-9.09$ & graded \\
IA71 & \texttt{qwen25-coder-32b} & mpnharsh & SB & $0$ & $1038$ & $4.67$ & $[4.42,\ 4.93]$ & $-0.64$ & graded \\
IA72 & \texttt{qwen25-coder-32b} & rigorous & SB & $0$ & $1038$ & $4.73$ & $[4.47,\ 5.00]$ & $-1.47$ & graded \\
IA73 & \texttt{qwen25-coder-32b} & exacting & SB & $0$ & $1038$ & $5.43$ & $[5.14,\ 5.73]$ & $-3.73$ & graded \\
IA74 & \texttt{qwen25-coder-32b} & mpframe & SB & $0$ & $1038$ & $4.78$ & $[4.52,\ 5.05]$ & $-1.71$ & graded \\
IA75 & \texttt{qwen25-coder-32b} & mps2only & SB & $0$ & $1038$ & $6.68$ & $[6.35,\ 7.02]$ & $-5.67$ & graded \\
IA76 & \texttt{qwen25-coder-32b} & mpharsh & SB & $0$ & $1038$ & $13.24$ & $[12.80,\ 13.70]$ & $-13.07$ & graded \\
IA77 & \texttt{qwen25-coder-7b} & neutral & S & $0$ & $1038$ & $7.55$ & $[7.20,\ 7.89]$ & $+4.92$ & graded \\
IA78 & \texttt{qwen25-coder-7b} & lenient & SB & $0$ & $1038$ & $8.14$ & $[7.79,\ 8.50]$ & $+6.50$ & graded \\
IA79 & \texttt{qwen3-coder-30b-a3b} & neutral & S & $0$ & $1038$ & $9.13$ & $[8.73,\ 9.54]$ & $+7.80$ & graded \\
IA80 & \texttt{qwen3-coder-30b-a3b} & lenient & SB & $0$ & $1038$ & $17.16$ & $[16.58,\ 17.74]$ & $+17.02$ & collapse \\
IA91 & \texttt{qwen25-coder-32b} & neutral & B & $0$ & $1033$ & $4.73$ & $[4.49,\ 4.98]$ & $+1.20$ & graded \\
IA92 & \texttt{qwen25-coder-32b} & neutral & SB & $0.5$ & $41$ & $4.65$ & $[2.78,\ 7.37]$ & $+4.51$ & graded \\
IA93 & \texttt{qwen25-coder-32b} & neutral & SBR & $0$ & $1033$ & $4.59$ & $[4.34,\ 4.85]$ & $-0.24$ & graded \\
IA94 & \texttt{qwen25-coder-7b} & neutral & B & $0$ & $1038$ & $7.51$ & $[7.16,\ 7.88]$ & $+5.57$ & graded \\
IA95 & \texttt{qwen3-coder-30b-a3b} & neutral & B & $0$ & $1013$ & $11.96$ & $[11.50,\ 12.40]$ & $+11.53$ & graded \\
IA96 & \texttt{qwen3-coder-30b-a3b} & neutral & SBR & $0$ & $1017$ & $11.09$ & $[10.61,\ 11.57]$ & $+10.40$ & graded \\
IA101 & \texttt{glm45-air} & neutral & SB & $0$ & $1037$ & $5.60$ & $[5.31,\ 5.89]$ & $-1.34$ & graded \\
IA102 & \texttt{glm45-air} & strict & SB & $0$ & $1038$ & $18.49$ & $[17.96,\ 19.02]$ & $-18.31$ & collapse \\
IA103 & \texttt{glm45-air} & lenient & SB & $0$ & $1038$ & $15.30$ & $[14.74,\ 15.87]$ & $+15.17$ & graded \\
IA104 & \texttt{glm45-air} & rigorous & SB & $0$ & $1037$ & $6.26$ & $[5.95,\ 6.58]$ & $-3.97$ & graded \\
IA105 & \texttt{glm45-air} & exacting & SB & $0$ & $1022$ & $9.95$ & $[9.55,\ 10.36]$ & $-9.15$ & graded \\
IA106 & \texttt{llama33-70b} & neutral & S & $0$ & $1016$ & $4.29$ & $[4.05,\ 4.55]$ & $+0.01$ & graded \\
IA107 & \texttt{llama33-70b} & strict & S & $0$ & $1013$ & $14.13$ & $[13.62,\ 14.64]$ & $-13.64$ & graded \\
IA108 & \texttt{llama33-70b} & mpstrict & SB & $0$ & $1038$ & $9.56$ & $[9.16,\ 9.96]$ & $-9.05$ & graded \\
IA109 & \texttt{llama33-70b} & mprigorous & SB & $0$ & $1038$ & $8.55$ & $[8.18,\ 8.94]$ & $-7.86$ & graded \\
IA110 & \texttt{llama33-70b} & mpfair & SB & $0$ & $1038$ & $8.13$ & $[7.76,\ 8.50]$ & $-7.36$ & graded \\
IA111 & \texttt{llama33-70b} & mpnoclause & SB & $0$ & $1038$ & $6.12$ & $[5.82,\ 6.43]$ & $-4.87$ & graded \\
IA112 & \texttt{llama33-70b} & mpnharsh & SB & $0$ & $1038$ & $4.60$ & $[4.36,\ 4.85]$ & $+1.61$ & graded \\
IA113 & \texttt{llama33-70b} & neutral & SB & $0$ & $1038$ & $4.70$ & $[4.45,\ 4.96]$ & $+1.90$ & graded \\
IA114 & \texttt{llama33-70b} & strict & SB & $0$ & $1038$ & $10.17$ & $[9.77,\ 10.58]$ & $-9.63$ & graded \\
IA115 & \texttt{llama33-70b} & lenient & SB & $0$ & $1038$ & $11.74$ & $[11.34,\ 12.15]$ & $+11.55$ & graded \\
IA116 & \texttt{llama33-70b} & rigorous & SB & $0$ & $1038$ & $4.60$ & $[4.36,\ 4.85]$ & $+1.71$ & graded \\
IA117 & \texttt{llama33-70b} & exacting & SB & $0$ & $1038$ & $5.23$ & $[4.95,\ 5.51]$ & $-3.31$ & graded \\
IA118 & \texttt{qwen25-72b} & neutral & SB & $0$ & $1038$ & $10.64$ & $[10.25,\ 11.03]$ & $+10.31$ & graded \\
IA119 & \texttt{qwen25-72b} & strict & SB & $0$ & $1038$ & $5.42$ & $[5.15,\ 5.70]$ & $-1.67$ & graded \\
IA120 & \texttt{qwen25-72b} & lenient & SB & $0$ & $1038$ & $16.93$ & $[16.44,\ 17.42]$ & $+16.89$ & collapse \\
IA121 & \texttt{qwen25-72b} & rigorous & SB & $0$ & $1038$ & $8.50$ & $[8.16,\ 8.85]$ & $+7.94$ & graded \\
IA122 & \texttt{qwen25-72b} & exacting & SB & $0$ & $1038$ & $8.45$ & $[8.09,\ 8.81]$ & $+7.80$ & graded \\
IA123 & \texttt{qwen3-235b-a22b} & neutral & SB & $0$ & $1037$ & $5.40$ & $[5.13,\ 5.67]$ & $+3.73$ & graded \\
IA124 & \texttt{qwen3-235b-a22b} & strict & SB & $0$ & $1038$ & $5.18$ & $[4.89,\ 5.47]$ & $-3.39$ & graded \\
IA125 & \texttt{qwen3-235b-a22b} & lenient & SB & $0$ & $1038$ & $17.46$ & $[16.99,\ 17.93]$ & $+17.41$ & collapse \\
IA126 & \texttt{qwen3-235b-a22b} & rigorous & SB & $0$ & $1038$ & $4.53$ & $[4.29,\ 4.78]$ & $+2.11$ & graded \\
IA127 & \texttt{qwen3-235b-a22b} & exacting & SB & $0$ & $1038$ & $4.58$ & $[4.32,\ 4.84]$ & $-1.58$ & graded \\
IA128 & \texttt{qwen3-coder-480b} & neutral & SB & $0$ & $1038$ & $10.10$ & $[9.66,\ 10.54]$ & $+9.52$ & graded \\
IA129 & \texttt{qwen3-coder-480b} & strict & SB & $0$ & $1038$ & $6.04$ & $[5.75,\ 6.34]$ & $+3.25$ & graded \\
IA130 & \texttt{qwen3-coder-480b} & lenient & SB & $0$ & $1038$ & $15.37$ & $[14.87,\ 15.88]$ & $+15.29$ & graded \\
IA131 & \texttt{qwen3-coder-480b} & rigorous & SB & $0$ & $1038$ & $7.89$ & $[7.54,\ 8.25]$ & $+6.99$ & graded \\
IA132 & \texttt{qwen3-coder-480b} & exacting & SB & $0$ & $1038$ & $7.03$ & $[6.69,\ 7.37]$ & $+5.04$ & graded \\
IA133 & \texttt{qwen3-coder-next} & neutral & SB & $0$ & $1021$ & $4.47$ & $[4.24,\ 4.72]$ & $+1.51$ & graded \\
IA134 & \texttt{qwen3-coder-next} & strict & SB & $0$ & $1001$ & $8.63$ & $[8.26,\ 9.00]$ & $-8.12$ & graded \\
IA135 & \texttt{qwen3-coder-next} & lenient & SB & $0$ & $1021$ & $7.64$ & $[7.32,\ 7.96]$ & $+7.10$ & graded \\
IA136 & \texttt{qwen3-coder-next} & neutral & SB & $0.5$ & $41$ & $5.04$ & $[3.26,\ 7.76]$ & $+5.04$ & graded \\
IA137 & \texttt{qwen3-coder-next} & neutral & SBR & $0$ & $1021$ & $4.47$ & $[4.24,\ 4.72]$ & $+1.51$ & graded \\
IA201 & \texttt{qwen3-coder-30b-a3b} & neutral & SBF & $0$ & $1038$ & $5.84$ & $[5.56,\ 6.13]$ & $+4.56$ & graded \\
\end{longtable}
\endgroup

\end{document}